%% file: iclr2027_conference.tex
\documentclass{article} 
\usepackage{iclr2027_conference,times}
\iclrfinalcopy
\input{math_commands.tex}

\usepackage{titletoc}
\usepackage{caption}
\usepackage{wrapfig}
\usepackage{enumitem}
\usepackage{graphicx}      
\usepackage{subcaption}    
\usepackage[most]{tcolorbox}
\usepackage{xcolor}
\usepackage{tabularx}

\definecolor{promptblue}{RGB}{245,248,252}
\definecolor{promptborder}{RGB}{150,165,185}

\tcbset{
    promptbox/.style={
        colback=promptblue,
        colframe=promptborder,
        colbacktitle=blue!8,
        coltitle=black,
        boxrule=0.6pt,
        arc=2pt,
        fonttitle=\bfseries,
        title={#1},
        left=1.5mm,
        right=1.5mm,
        top=1mm,
        bottom=1mm
    }
}
\usepackage[font=small,labelfont=bf,skip=5pt]{caption}

\newcommand{\uifig}[4]{%
  \begin{figure}[t]
    \centering
    \includegraphics[width=#1\linewidth]{figures/appendix/#2}%
    \caption{#3}
    \label{#4}
  \end{figure}%
}
\usepackage{hyperref}
\usepackage{url}
\usepackage{mwe}
\usepackage{array}
\usepackage{makecell}
\usepackage{multirow}
\usepackage{amsmath}
\usepackage{colortbl}
\usepackage{booktabs}
\usepackage{xspace}
\usepackage[table]{xcolor}
\usepackage{pifont}
\usepackage{makecell}
\usepackage{tikz}
\definecolor{RankUp}{RGB}{72, 160, 96}
\definecolor{RankDown}{RGB}{205, 92, 92}

\newcommand{\benchmark}{\textsc{AUV-Bench}\xspace}
\newcommand{\cmark}{\textcolor{green!55!black}{\ding{51}}}
\newcommand{\xmark}{\textcolor{red!80!black}{\ding{55}}}

\newcommand{\modellogo}[1]{%
    \raisebox{-0.18\height}{%
        \includegraphics[height=1.05em,keepaspectratio]{#1}%
    }\hspace{0.35em}%
}

\definecolor{TableHeader}{HTML}{E9E7F2}
\definecolor{TableAltRow}{HTML}{F5F3F9}
\definecolor{TableRule}{HTML}{B7B4BE}

\definecolor{DimTypography}{HTML}{F2B164}
\definecolor{DimLayout}{HTML}{83BDB5}
\definecolor{DimSpacing}{HTML}{98A5CF}
\definecolor{DimStyle}{HTML}{B7662C}
\definecolor{DimColor}{HTML}{C67EA5}

\newcommand{\dimcircle}[2]{%
    \tikz[baseline=(char.base)]{
        \node[
            circle,
            fill=#1,
            minimum size=5.2mm,
            inner sep=0pt
        ] (char) {
            \color{white}
            \bfseries
            \scriptsize
            #2
        };
    }%
}

\newcommand{\dimT}{\dimcircle{DimTypography}{T}}
\newcommand{\dimL}{\dimcircle{DimLayout}{L}}
\newcommand{\dimS}{\dimcircle{DimSpacing}{S}}
\newcommand{\dimV}{\dimcircle{DimStyle}{V}}
\newcommand{\dimC}{\dimcircle{DimColor}{C}}

\title{\benchmark: Aesthetic Understanding and Generation EValuation for User Interfaces}

\author{
\textbf{Zhijie Deng}\textsuperscript{1,3}\thanks{Equal contribution.}, \hspace{0.5mm}
\textbf{Ling Li}\textsuperscript{1}\footnotemark[1], \hspace{0.5mm}
\textbf{Junhao Ji}\textsuperscript{3}, \hspace{0.5mm}
\textbf{Siwei Lyu}\textsuperscript{3},
\hspace{0.5mm}
\textbf{Zhipeng Xu}\textsuperscript{3}, \hspace{0.5mm}
\textbf{Zulong Chen}\textsuperscript{3}\thanks{Corresponding authors:
jiahengwei@hkust-gz.edu.cn and zulong.czl@alibaba-inc.com.}, \\
\textbf{Rongyao Fang}\textsuperscript{3},
\hspace{0.5mm}
\textbf{Shuai Bai}\textsuperscript{3},
\hspace{0.5mm}
\textbf{Xuming Hu}\textsuperscript{1,2},
\hspace{0.5mm}
\textbf{Jiaheng Wei}\textsuperscript{1}\footnotemark[2]
\\
\textsuperscript{1}The Hong Kong University of Science and Technology (Guangzhou) \\
\textsuperscript{2}The Hong Kong University of Science and Technology \quad
\textsuperscript{3}Alibaba Group \\
\texttt{\small zdeng190@connect.hkust-gz.edu.cn,
li297@connect.hkust-gz.edu.cn}
}
\begin{document}

\maketitle

\input{sections/0_abstract.tex}
\input{sections/1_introduction.tex}

\input{sections/2_relatedwork.tex}
\input{sections/3_benchmark.tex}
\input{sections/4_experiments}
\input{sections/5_conclusion}

\section*{AI use statement}

In this work, we used generative AI tools for four purposes.
First, large language models (LLMs) were used to assist with language polishing
and improving the clarity of the manuscript.
Second, LLMs were used to assist with code development and debugging; all
AI-assisted code was reviewed, verified, and tested by multiple authors before
use in our experiments.
Third, generative AI models were used during benchmark construction, including
generating a subset of the webpages and reconstructing natural-language
requirements for Text-to-UI Generation, as described in the paper.
Fourth, GPT-5.4 was used as an automatic aesthetic judge for Text-to-UI
Generation; its scoring behavior was validated and calibrated against
professional-designer annotations from Aesthetic Scoring before being applied
to the generation results. All AI-assisted outputs used in the final work were reviewed by the authors,
and human-reviewed or validated where specified in the corresponding benchmark
construction and evaluation procedures. We take responsibility for the final
content of this work, including text, code, data, experimental results, and
artifacts produced with the aid of generative AI.

\section*{Reproducibility statement}
We will release the source code and dataset upon acceptance of the paper to facilitate reproducibility and future research.


\bibliography{iclr2027_conference}
\bibliographystyle{iclr2027_conference}

\newpage
\appendix
\startcontents[appendix]

\begin{center}
    {\LARGE\bfseries Appendix}
\end{center}

\vspace{3mm}

\section*{Table of Contents}
\hrule
\printcontents[appendix]{}{1}{
    \setcounter{tocdepth}{2}
}
\vspace{3mm}
\hrule


\newpage
\input{sections/appendix_ase_rule}
\input{sections/appendix_human_anno}
\input{sections/appendix_benchmark_details}
\input{sections/appendix_experiment}
\input{sections/appendix_more_expe}
\end{document}

%% file: math_commands.tex
\usepackage{amsmath,amsfonts,bm}

\def\eqref#1{equation~\ref{#1}}

\def\1{\bm{1}}

\DeclareMathAlphabet{\mathsfit}{\encodingdefault}{\sfdefault}{m}{sl}
\SetMathAlphabet{\mathsfit}{bold}{\encodingdefault}{\sfdefault}{bx}{n}



%% file: sections/0_abstract.tex
\begin{abstract}
Multimodal foundation models are increasingly used for evaluating and generating user interfaces (UIs), often producing seemingly reasonable aesthetic judgments and visually plausible pages. However, under professional design scrutiny, their behavior can differ substantially from that of human designers. In professional design practice, designers rely on a systematic set of aesthetic principles that consistently guide judgment, diagnosis, repair, and creation. A coherent aesthetic capability should therefore connect aesthetic judgment with design actions. Existing evaluations, however, typically assess these abilities in isolation, making it difficult to determine whether task-level success reflects a shared aesthetic understanding or merely fragmented task-specific competence. To address this gap, we introduce \benchmark, developed in collaboration with professional UI designers around 1,395 executable web interfaces and four tasks: aesthetic scoring, diagnosis, repair, and text-to-UI generation. The tasks share a pool of UIs and aesthetic principles, with diagnosis and repair further aligned on 660 controlled-degradation instances to enable instance-level analysis of judgment and action. Evaluation of 12 models reveals a capability imbalance: models show moderate agreement with professional designers in holistic aesthetic scoring, yet exact diagnosis-chain success peaks at only 24.7\%. On the aligned diagnosis--repair cases, correct judgments and successful repairs do not consistently coincide, exposing a \textit{Judgment--Action Gap} between identifying aesthetic problems and successfully acting on them. In open-ended generation, even leading models achieve only moderate aesthetic quality under human-calibrated evaluation. Overall, current models exhibit partial aesthetic competence, but still lack the fine-grained understanding and judgment--action coherence required for reliable UI design. Our code and dataset are available at \url{https://github.com/yuu250/AUV_Bench}.
\end{abstract}

%% file: sections/1_introduction.tex
\section{Introduction}

Multimodal foundation models are increasingly taking on the dual roles of
``critics'' and ``designers'' of user interfaces (UIs): they can evaluate the
visual quality of interfaces~\citep{duan2024uicrit,chen2024mllm,chen2025safeeraser} and generate
visually plausible webpages from visual or natural-language
requirements~\citep{si2025design2code,awal2025webmmu}. Yet under professional
design scrutiny, both their aesthetic judgments and generated interfaces still
exhibit substantial gaps from those of experienced human designers
~\citep{huang2024aesbench,an2026aeseval}. More importantly, evaluating these
abilities in isolation cannot reveal whether they are supported by a shared and
transferable understanding of aesthetic principles, or merely reflect
task-specific competence. In professional design practice, experienced designers internalize accumulated
aesthetic knowledge and design experience into stable judgment criteria, and
consistently apply them across interface evaluation, problem diagnosis, design
repair, and open-ended creation
~\citep{cross2004expertise,lawson2004schemata}. Therefore, a model with mature
UI aesthetic capability should not merely perform well on individual tasks; its
aesthetic knowledge should transfer across tasks and consistently connect design
judgment with corresponding design actions. We define this consistency as
\textit{Aesthetic Judgment--Action Coherence}. This raises a central question:
\textit{Have current multimodal models developed a coherent capability that
connects aesthetic judgment with design action?}

Existing evaluations are not yet well suited to systematically answer this question. As shown in Table~\ref{tab:benchmark_comparison}, prior work has primarily followed two directions: one focuses on UI generation, editing, and front-end implementation, evaluating visual fidelity, functional correctness, and executability~\citep{lin2025webuibench,liu2026webcoderbench,lu2025webgenbench,deng2025guard}; the other focuses on aesthetic understanding, evaluating scoring, design feedback, and defect localization~\citep{wang2026uxbench,lin2024designprobe,du2026venus,yuqian,deng2026codeblock}. However, these evaluations typically treat performance on individual tasks as separate endpoints. Successfully generating a visually reasonable interface does not directly indicate that a model can reliably judge its design quality; likewise, accurately identifying an aesthetic issue does not guarantee that the corresponding judgment can guide an effective repair. Studying this connection requires evaluation to be organized around shared aesthetic principles, with judgment and corrective action aligned on the same interface and target defect. Existing evaluations provide limited support for such principle-level and instance-level linkage, making it difficult to determine whether success across different tasks reflects the consistent use of aesthetic design knowledge~\citep{imteyaz2026design,an2026aeseval}.
\begin{table*}[t]
    \centering
    \caption{
    Comparison of existing benchmarks along key dimensions for evaluating UI aesthetic judgment and design action.
    \textbf{Expert Aesthetic GT} indicates whether aesthetic ground truth is provided or validated by professional designers or relevant domain experts.
    \textbf{Principle-Grounded} denotes explicit evaluation against predefined aesthetic design principles, while
    \textbf{Controlled Violations} indicates controlled manipulation of specific aesthetic principles.
    \textbf{Judgment--Action Linkage} indicates whether aesthetic diagnosis and corrective action are directly paired on the same interface and controlled aesthetic violation, enabling instance-level analysis of judgment--action coherence.
    \cmark\ and \xmark\ denote support and non-support, respectively.
    }
    \label{tab:benchmark_comparison}

    \setlength{\tabcolsep}{4.0pt}
    \renewcommand{\arraystretch}{1.08}

    \resizebox{\textwidth}{!}{
    \begin{tabular}{lccccccccccl}
        \toprule

        \multirow{2}{*}{\textbf{Benchmark}}
        &
        \multicolumn{2}{c}{\textbf{Benchmark Setup}}
        &
        \multicolumn{2}{c}{\textbf{Principle Grounding}}
        &
        \multicolumn{4}{c}{\textbf{Core Capabilities}}
        &
        \multirow{2}{*}{
            \makecell{
                \textbf{Judgment--Action}\\
                \textbf{Linkage}
            }
        }
        &
        \multicolumn{2}{c}{\textbf{Statistics}}
        \\

        \cmidrule(lr){2-3}
        \cmidrule(lr){4-5}
        \cmidrule(lr){6-9}
        \cmidrule(lr){11-12}

        &
        \makecell{\textbf{Executable}\\\textbf{UI}}
        &
        \makecell{\textbf{Expert}\\\textbf{Aesthetic GT}}
        &
        \makecell{\textbf{Principle-}\\\textbf{Grounded}}
        &
        \makecell{\textbf{Controlled}\\\textbf{Violations}}
        &
        \makecell{\textbf{Aesthetic}\\\textbf{Scoring}}
        &
        \makecell{\textbf{Aesthetic}\\\textbf{Diagnosis}}
        &
        \makecell{\textbf{Aesthetic}\\\textbf{Repair}}
        &
        \makecell{\textbf{Text-to-UI}\\\textbf{Generation}}
        &
        &
        \textbf{Domain}
        &
        \textbf{Scale}
        \\

        \midrule

        \rowcolor{blue!7}
        \multicolumn{12}{c}{
            \textit{\textbf{Benchmarks for UI Generation}}
        }
        \\

        Design2Code~\citep{si2025design2code}
        & \cmark
        & \xmark
        & \xmark
        & \xmark
        & \xmark
        & \xmark
        & \xmark
        & \xmark
        & \xmark
        & Web UI
        & 484 pages
        \\

        WebUIBench~\citep{lin2025webuibench}
        & \cmark
        & \xmark
        & \xmark
        & \xmark
        & \xmark
        & \xmark
        & \xmark
        & \xmark
        & \xmark
        & Web UI
        & 21K QA / 0.7K+ sites
        \\

        WebMMU~\citep{awal2025webmmu}
        & \cmark
        & \xmark
        & \xmark
        & \xmark
        & \xmark
        & \xmark
        & \xmark
        & \xmark
        & \xmark
        & Web UI
        & 8.1K tasks / 2,059 pages
        \\

        WebGen-Bench~\citep{lu2025webgenbench}
        & \cmark
        & \xmark
        & \xmark
        & \xmark
        & \xmark
        & \xmark
        & \xmark
        & \cmark
        & \xmark
        & Web App
        & 647 tests
        \\

        WebCoderBench~\citep{liu2026webcoderbench}
        & \cmark
        & \xmark
        & \xmark
        & \xmark
        & \xmark
        & \xmark
        & \xmark
        & \cmark
        & \xmark
        & Web App
        & 1,572 requirements
        \\

        UI-Bench~\citep{jung2025ui}
        & \cmark
        & \cmark
        & \xmark
        & \xmark
        & \xmark
        & \xmark
        & \xmark
        & \cmark
        & \xmark
        & Web UI
        & 300 sites / 4K+ judgments
        \\

        \midrule

        \rowcolor{blue!7}
        \multicolumn{12}{c}{
            \textit{\textbf{Benchmarks for Aesthetic Evaluation}}
        }
        \\

        UICrit~\citep{duan2024uicrit}
        & \xmark
        & \cmark
        & \xmark
        & \xmark
        & \cmark
        & \cmark
        & \xmark
        & \xmark
        & \xmark
        & Mobile UI
        & 983 UIs / 3,059 critiques
        \\

        DesignProbe~\citep{lin2024designprobe}
        & \xmark
        & \cmark
        & \xmark
        & \xmark
        & \xmark
        & \xmark
        & \xmark
        & \xmark
        & \xmark
        & Graphic Design
        & $\sim$1.6K questions
        \\

        PhotoBench~\citep{qi2025photographer}
        & \xmark
        & \xmark
        & \xmark
        & \xmark
        & \cmark
        & \xmark
        & \xmark
        & \xmark
        & \xmark
        & Photography
        & 1.5K questions
        \\

        AesEval-Bench~\citep{an2026aeseval}
        & \xmark
        & \xmark
        & \cmark
        & \cmark
        & \cmark
        & \cmark
        & \xmark
        & \xmark
        & \xmark
        & Graphic Design
        & 4.5K QA / $\sim$1.2K designs
        \\

        AesGuide~\citep{du2026venus}
        & \xmark
        & \cmark
        & \xmark
        & \xmark
        & \cmark
        & \cmark
        & \xmark
        & \xmark
        & \xmark
        & Photography
        & 1K eval / 10.7K total
        \\

        \midrule

        \rowcolor{green!7}
        \textbf{\benchmark}
        & \textbf{\cmark}
        & \textbf{\cmark}
        & \textbf{\cmark}
        & \textbf{\cmark}
        & \textbf{\cmark}
        & \textbf{\cmark}
        & \textbf{\cmark}
        & \textbf{\cmark}
        & \textbf{\cmark}
        & \textbf{Web UI}
        & \textbf{1,395 base UIs}
        \\

        \bottomrule
    \end{tabular}
    }

\end{table*}

\begin{figure}[htbp]
\vspace{-5mm}
\centering
\includegraphics[width=1.0\textwidth]{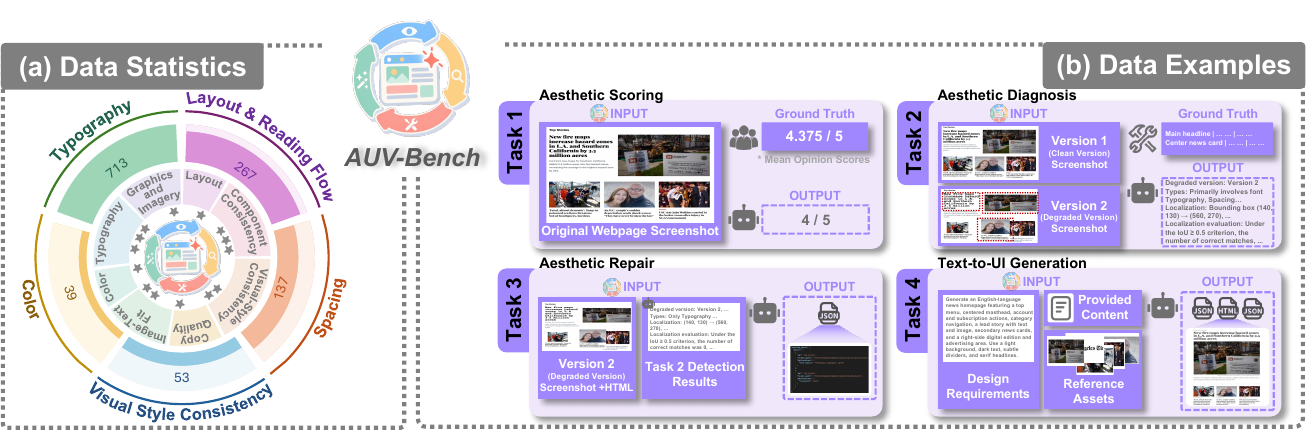}
\caption{Overview of \benchmark. (a) Data Statistics and (b) representative Data Examples of four tasks spanning aesthetic understanding (\emph{Aesthetic Scoring} and \emph{Aesthetic Diagnosis}) and design action (\emph{Aesthetic Repair} and \emph{Text-to-UI Generation}).}
\label{fig:intro}
\end{figure}

To address this gap, we introduce \benchmark, an \textbf{A}esthetic \textbf{U}nderstanding and generation e\textbf{V}aluation for UIs. An overview of \benchmark is shown in Figure~\ref{fig:intro}. We work with experienced UI designers to organize systematic aesthetic principles, recruit 51 professional designers for human annotation, and construct a shared data foundation containing 1,395 executable Web UIs. Based on this foundation, \textit{Aesthetic Scoring} and \textit{Aesthetic Diagnosis} evaluate aesthetic understanding, while \textit{Aesthetic Repair} and \textit{Text-to-UI Generation} evaluate design action.

To further study \textit{Aesthetic Judgment--Action Coherence}, we use controlled degradations to align diagnosis and repair on the same webpage, and introduce \textit{Judgment--Action Association} (JAA) to measure the instance-level association between judgment correctness and repair success. Our systematic evaluation reveals clear imbalances across aesthetic capabilities: frontier models show moderate agreement with expert judgments in holistic aesthetic scoring, while precise defect attribution and localization remain more challenging; correct judgments and successful repairs are generally positively associated, yet they do not form a stable correspondence, revealing a clear \textit{Judgment--Action Gap}. In open-ended generation, even relatively strong models achieve only moderate aesthetic quality under human-calibrated evaluation, further distinguishing relative model advantages from design quality measured on a professional rating scale.
Our main contributions are as follows:
\begin{itemize}[leftmargin=*, itemsep=2pt, topsep=2pt, parsep=0pt, partopsep=0pt]

    \item \textbf{A unified benchmark for UI aesthetic understanding and design action.}
Unlike existing benchmarks that evaluate aesthetic understanding or UI generation in isolation, \benchmark evaluates scoring, diagnosis, repair, and text-to-UI generation over 1,395 executable web interfaces, enabling unified evaluation from judgment to action.

    \item \textbf{A principle-grounded protocol for cross-task aesthetic evaluation.}
    We operationalize professional aesthetic principles into task-specific evaluation criteria and 13 controlled source-level violations with executable verification. Diagnosis and repair are further paired on 660 controlled-degradation instances, enabling direct instance-level measurement of whether aesthetic judgments translate into corrective actions.

    \item \textbf{New empirical insights into the limits of current UI aesthetic capabilities.}
    Models show moderate holistic aesthetic judgment, yet exact fine-grained diagnosis peaks at only 24.7\%. More importantly, successful repair does not consistently coincide with correct explicit judgment, while strong relative generation performance still yields only moderate absolute aesthetic quality, revealing substantial gaps between apparent task competence and coherent aesthetic understanding.

\end{itemize}

%% file: sections/2_relatedwork.tex
\section{Related Work}
\label{sec:related_work}
\textbf{UI Generation and Front-End Development.}
Early works such as Rico, pix2code, and GUI skeleton generation established large-scale UI modeling and visual-to-code generation~\citep{deka2017rico,beltramelli2018pix2code,chen2018ui}. More recent systems, including Design2Code, Sketch2Code, and UI2Code, advance webpage reconstruction, code generation, and iterative refinement~\citep{si2025design2code,li2025sketch2code,yang2025ui2code,wan2025divide}. WebUIBench, FullFront, WebMMU, and DesignBench broaden evaluation to UI understanding and general front-end capabilities~\citep{lin2025webuibench,sun2025fullfront,awal2025webmmu,xiao2025designbench,he2026vision2web}, while FrontendBench, WebGen-Bench, and WebCoderBench emphasize executable and functional correctness~\citep{zhu2025frontendbench,lu2025webgenbench,liu2026webcoderbench}. More recently, UI-Bench evaluates the design quality of generated interfaces, and Design Theater studies whether design reasoning supports downstream implementation~\citep{jung2025ui,imteyaz2026design}. These works increasingly extend UI evaluation from reconstruction and functionality toward design quality and action.

\textbf{Aesthetic and Design Evaluation.}
Computational aesthetics has traditionally focused on holistic preference prediction, as exemplified by AVA and NIMA~\citep{murray2012ava,talebi2018nima}. With multimodal models, AesBench evaluates aesthetic understanding, while DesignProbe and GPT-based evaluation examine reasoning over design properties and principles~\citep{huang2024aesbench,lin2024designprobe,haraguchi2024can,lin2026can}. UICrit introduces structured and localized UI critiques, while DesignPref studies individual visual-design preferences~\citep{duan2024uicrit,peng2025designpref,ko2026criticmate}. Recent benchmarks move toward finer-grained analysis: AesEval-Bench covers aesthetic scoring and defect localization, UXBench evaluates detailed UI and UX reasoning, and UXBench-Actionability measures whether critiques support downstream repair~\citep{an2026aeseval,mao2026reasoning,wang2026uxbench,jeon2026mllms}. However, these works typically evaluate aesthetic scoring, diagnosis, repair, or generation separately, or connect only a subset of them. \benchmark instead places all four capabilities under shared professional aesthetic principles, enabling systematic analysis of how aesthetic judgment translates into design action.

%% file: sections/3_benchmark.tex
\section{\benchmark}
\label{sec:bench_construct}
This section presents the design of \benchmark, including its task formulation, benchmark construction, and evaluation protocol.

\subsection{Overview}

\benchmark evaluates UI aesthetics through four complementary capabilities built on a shared pool of executable reference UIs and professional aesthetic guidelines. Specifically, we introduce four tasks: (1) \textbf{Aesthetic Scoring} takes a UI image as input and outputs eight dimension-level aesthetic scores on a 1--5 scale, measuring holistic aesthetic judgment; (2) \textbf{Aesthetic Diagnosis} takes a pair of clean and degraded UI images and outputs the detected aesthetic violation(s) together with their affected regions, evaluating principle-level diagnosis and localization; (3) \textbf{Aesthetic Repair} takes a degraded UI image, its HTML source code, and the model's own diagnosis, and outputs a source-code patch that corrects the identified aesthetic issues; and (4) \textbf{Text-to-UI Generation} takes a natural-language webpage requirement together with textual content and image assets, and outputs an executable webpage without access to a reference screenshot. \textbf{Aesthetic Diagnosis} and \textbf{Aesthetic Repair} share the same controlled degradation instances, enabling direct instance-level analysis of judgment--action alignment.

\subsection{Dataset Statistics}
\label{sec:dataset_statistics}

\begin{figure*}[t]
    \centering
    \includegraphics[width=\textwidth]{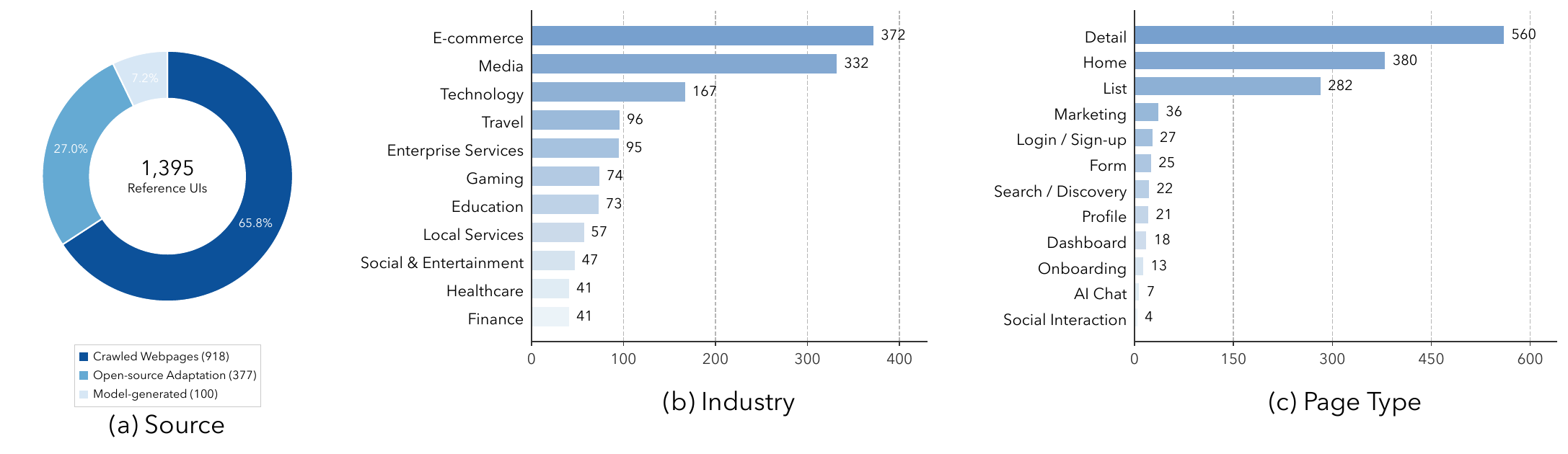}
    \caption{
    Distribution of the 1,395 reference UIs.
    (a) Data sources.
    (b) Industry categories.
    (c) Page types.
    }
    \label{fig:reference_ui_distribution}
\end{figure*}

\benchmark comprises 1,395 executable reference UIs covering 11 industry
categories and 12 page types (Figure~\ref{fig:reference_ui_distribution}).
The crawled subset spans 386 web domains, and textual content covers 29
languages. Aesthetic Scoring and Text-to-UI Generation use all pages,
while Aesthetic Diagnosis and Aesthetic Repair share 660
controlled-degradation cases with 1,209 annotated aesthetic violations
across 13 rule types. Counting each task independently yields 4,110
evaluation instances. Aesthetic Scoring retains ratings from 44
professional UI designers after quality control. Further details are
provided in Appendix~\ref{sec:bench_details}.

\subsection{Task Formulation}
\label{sec:task_formulation}

Let $h$ and $x$ denote a UI's executable HTML source and rendered screenshot.
Superscripts $\mathrm{ref}$ and $\mathrm{deg}$ denote reference and degraded
versions; controlled degradation injects aesthetic violations while preserving
content. Aesthetic Scoring predicts eight-dimensional scores
$\hat{\mathbf{s}}$ from $x^{\mathrm{ref}}$, using designer mean opinion scores
(MOS) $\mathbf{s}$ as targets. Aesthetic Diagnosis and Aesthetic Repair share
instances with violation annotations $v$ specifying affected aesthetic dimensions
and regions. Diagnosis identifies the degraded screenshot and predicts
the corresponding annotations $\hat{v}$; repair uses $(h^{\mathrm{deg}},x^{\mathrm{deg}},\hat{v})$
to correct the violations. For Text-to-UI Generation, $q$ comprises the webpage
requirement, textual content, and image assets; the model generates source
$\hat{h}$, rendered as $\hat{x}$, without access to $x^{\mathrm{ref}}$.

\subsection{Benchmark Construction}

\begin{figure}[h]
\centering
\includegraphics[width=1.0\textwidth]{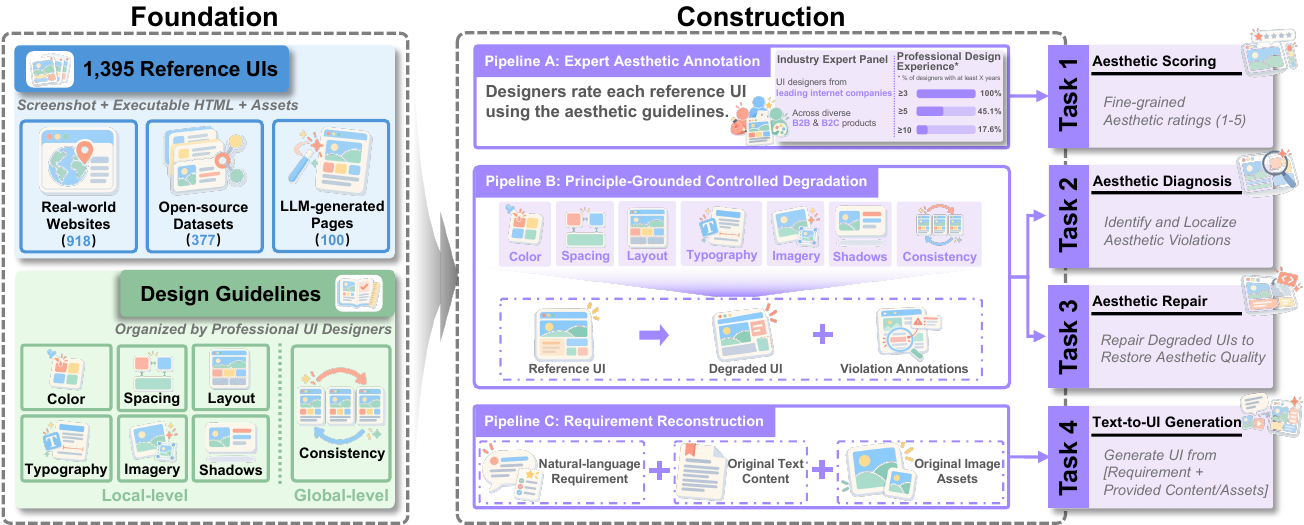}
\caption{
Overview of \benchmark construction. Starting from 1,395 executable UIs and professional aesthetic guidelines, we build four complementary tasks through expert annotation, controlled degradation, and requirement reconstruction. The controlled-degradation pipeline draws inspiration from all seven guideline dimensions and retains five operational categories for source-level editing and verification.
}
\label{fig:benchmark}
\end{figure}

Figure~\ref{fig:benchmark} summarizes our construction pipeline. We build four
tasks from 1,395 executable reference UIs through expert annotation,
controlled aesthetic degradation, and requirement reconstruction.
Professional UI designers co-develop the
guidelines and review reference pages, degradation rules, and reconstructed
requirements.

\textbf{Professional Aesthetic Guidelines.}
Together with three senior UI designers, each with at least five years of industry experience at leading Internet companies, we develop aesthetic guidelines grounded in HCI and visual-design principles across seven dimensions: color, spacing, layout, typography, imagery, shadows, and global visual consistency. These guidelines define a shared design-principle space, which we further adapt to the evaluation objectives of different tasks. For \textit{Aesthetic Scoring}, we reorganize the principles into eight perceptually coherent dimensions that can be reliably judged from rendered UIs: color, typography, graphics and imagery, layout, component consistency, visual-style consistency, copy quality, and image--text fit. Each dimension is rated on a shared 1--5 scale; detailed definitions are provided in Table~\ref{tab:fine_grained_criteria}. For \textit{Aesthetic Diagnosis} and \textit{Aesthetic Repair}, we retain principles that can yield perceptible, localizable, and attributable violations while supporting source-level editing and post-rendering verification. We organize them into five operational dimensions: typography, layout and reading flow, spacing, visual-style consistency, and color, and further instantiate 13 controlled aesthetic violation rules, detailed in Table~\ref{tab:controlled_violation_rules}.

For \textit{Aesthetic Diagnosis} and \textit{Aesthetic Repair}, the evaluation must instead correspond to specific aesthetic problems that can be explicitly identified and corrected. We therefore reorganize principles that admit perceptible, localizable, and attributable violations, while also supporting source-level modification and post-rendering verification, into five operational dimensions: typography, layout and reading flow, spacing, visual-style consistency, and color. Within these dimensions, we further instantiate the relevant design principles as 13 executable aesthetic violation rules, which are used to construct controlled degradations and define corresponding diagnosis and repair targets. The complete rule set is provided in Table~\ref{tab:controlled_violation_rules}.

\textbf{Aesthetic Scoring Annotation.}
Each reference UI is initially rated independently by three professional
designers using the shared eight-dimensional, 1--5 rubric. Annotators follow
common rating criteria and review visual exemplars to calibrate their
interpretation of the scale. After annotator-level quality control, retained
ratings are aggregated into dimension-level MOS, which serve as evaluation
targets.

\textbf{Principle-Grounded Controlled Degradation.}
We first screen reference UIs for rule applicability and existing violations,
applying a degradation only when the required structure is present and the
target region satisfies the corresponding principle. Each intervention
alters a small subset of eligible elements; designers iteratively review
the rules and perturbation parameters. Automatic validation retains samples
only when the intended defects are perceptible and localized, the detected
violation set matches the injected recipe, and no unintended layout changes
or unrelated aesthetic violations occur. We construct single-violation and
compositional cases at mild and severe levels. Each retained case includes
reference and degraded UIs with violation annotations, shared by Diagnosis
and Repair.

\textbf{Requirement Reconstruction.}
For \textit{Text-to-UI Generation}, GPT-5.4 reconstructs an English requirement
from each reference UI, describing its purpose, information structure, major
functions, and overall visual direction while omitting precise visual
parameters. Professional UI designers review each requirement and its
objectively verifiable checklist, preserving functional intent while leaving
visual implementation choices to the evaluated model.

Details of the guidelines, annotation, reference UI collection, controlled
degradation, and requirement reconstruction are provided in
Appendices~\ref{sec:ui-aesthetics}, \ref{app:human_evaluation},
\ref{app:reference_ui}, \ref{app:controlled_violations}, and
\ref{app:requirement_reconstruction}, respectively.

\subsection{Evaluation Protocol}
\label{sec:evaluation_protocol}

We evaluate multimodal foundation models across four complementary tasks,
covering aesthetic scoring, diagnosis, repair, and open-ended UI generation.

\textbf{\textit{Task 1: Aesthetic Scoring.}}
For instance $i$, the model predicts $\hat{\mathbf{s}}_i$ from
$x_i^{\mathrm{ref}}$ on a 1--5 scale. We compare $\hat{\mathbf{s}}_i$ with
$\mathbf{s}_i$ using \textit{Spearman Rank Correlation (SRCC)} and
\textit{Mean Absolute Error (MAE)}, measuring ranking agreement and absolute
error, respectively. Both metrics are computed across instances for each
dimension and averaged equally across all eight dimensions.

\textbf{\textit{Task 2: Aesthetic Diagnosis.}}
We randomly order $x_i^{\mathrm{ref}}$ and $x_i^{\mathrm{deg}}$ as candidates
$V_1$ and $V_2$. The model identifies the degraded candidate and predicts
$\hat{v}_i$, evaluated against $v_i$.
For degradation detection, we report \textit{Balanced Accuracy (BAcc)},
defined as
\(\mathrm{BAcc}=(\mathrm{Recall}_{V_1}+\mathrm{Recall}_{V_2})/2\),
where the two recalls correspond to instances with the first or second
candidate degraded, respectively.
For violation attribution, the labels are the five aesthetic dimensions:
typography, layout and reading flow, spacing, visual style consistency, and
color. We use \textit{detection-gated Macro-$J$}, where
\(J_d=\mathrm{TPR}_d-\mathrm{FPR}_d\) and
\(\mathrm{Macro}\text{-}J=\frac{1}{5}\sum_{d=1}^{5}J_d\).
Here, $d$ indexes the five attribution dimensions, and $\mathrm{TPR}_d$ and
$\mathrm{FPR}_d$ denote their true-positive and false-positive rates, respectively.
An incorrect degradation detection results in an empty attribution prediction.
Localization is evaluated using
\textit{version-aware Micro-F1@0.5}, where regions in $\hat{v}_i$ and $v_i$
are matched only on the correct version and with intersection over union
(IoU) of at least 0.5.

We further report the \textit{Exact Chain Success Rate (ECS)}, which requires
all three stages to be correct:
\begin{equation}
\mathrm{ECS}
=
\frac{1}{N}
\sum_{i=1}^{N}
\mathbf{1}
\left[D_i \land A_i \land L_i\right],
\label{eq:ecs}
\end{equation}
where $D_i$, $A_i$, and $L_i$ are Boolean indicators for correct degradation
detection, exact matching of the predicted and ground-truth dimension sets,
and localization of all ground-truth regions without false positives,
respectively.
Here, $N$ is the full number of evaluation instances in the current task,
$\mathbf{1}[\cdot]$ equals one if its condition holds and zero otherwise,
and $\land$ denotes logical conjunction.

\textbf{\textit{Task 3: Aesthetic Repair.}}
The model receives $(h_i^{\mathrm{deg}},x_i^{\mathrm{deg}},\hat{v}_i)$,
using its own Task~2 diagnosis, and outputs a constrained source-code patch.
The patched page is re-rendered for deterministic rule, DOM, and visual checks.
We report the \textit{Repair Pass Rate}. For instance $i$, the Boolean
indicator $R_i$ denotes repair success:
\begin{equation}
R_i
=
F_i \land C_i \land P_i \land S_i,
\label{eq:repair_pass}
\end{equation}
where $F_i$ indicates that all target violations are repaired, $C_i$ that no
excessive collateral damage is introduced, $P_i$ that the original content is
preserved, and $S_i$ that visual changes remain within the permitted repair
scope.
The final score is
\(\mathrm{RepairPass}=N^{-1}\sum_{i=1}^{N}\mathbf{1}[R_i]\).
An incorrect $\hat{v}_i$ can still yield a successful repair if all four
conditions hold.

\textbf{\textit{Task 4: Text-to-UI Generation.}}
Given $q_i$, the model generates $\hat{h}_i$ without access to
$x_i^{\mathrm{ref}}$. We evaluate the rendered $\hat{x}_i$ along two
complementary axes.

First, GPT-5.4 scores $\hat{x}_i$ using the Task~1 rubric, yielding raw scores
$a_{i,d}$ on a 1--5 scale, where $d$ indexes its eight dimensions. To correct bias relative
to designer ratings~\citep{niculescu2005predicting,deng2025lm}, we apply
dimension-specific monotonic calibration functions $f_d(\cdot)$, learned
exclusively from Task~1 MOS and frozen before Task~4 evaluation.
For a valid generation, the calibrated score is
$\mathrm{AesJudge}_i=\frac{1}{8}\sum_{d=1}^{8}f_d(a_{i,d})\in[1,5]$.
Invalid or failed generations receive zero. For model $m$, we report
$\mathrm{AesJudge}_m=\frac{1}{N}\sum_{i=1}^{N}\mathrm{AesJudge}_i$
over the fixed benchmark denominator $N$, yielding a model-level score in
$[0,5]$.
Judge validation and calibration details are provided in
Appendix~\ref{sec:more_exp}.

Second, an independent multimodal judge performs blind pairwise comparisons
of the rendered screenshots produced by two models for the same $q_i$.
When both webpages render successfully, the judge determines a win, loss,
or tie based on their overall aesthetic quality.
If only one webpage renders successfully, it wins automatically; if both fail,
neither receives credit.
For model $m$, its \textit{Overall Pairwise Win Rate} is
\begin{equation}
\mathrm{PairwiseWR}_m
=
\frac{1}{P-1}
\sum_{j\neq m}
\frac{W_{m,j}+0.5T_{m,j}}{N},
\label{eq:pairwise_wr}
\end{equation}
where $P$ is the number of participating models, $j$ indexes the other
models, and $W_{m,j}$ and $T_{m,j}$ count model $m$'s wins and ties against
model $j$, respectively.
All queries remain in the fixed denominator $N$, including cases where both
webpages fail to render. PairwiseWR lies in $[0,1]$ and is reported as a
percentage, with higher values indicating stronger relative preference
across the benchmark. Losses and rendering failures receive zero credit.

\textbf{Judgment--Action Association.}
To quantify judgment--action association, we pair diagnosis and repair
outcomes for each instance $i$. Correct judgment is indicated by
$J_i = D_i \land A_i$, requiring correct degradation detection and exact
dimension attribution. Repair success is given by $R_i$ in
Eq.~\ref{eq:repair_pass}.

Over the $N=660$ paired instances, the marginal judgment and repair success
rates are $p_J = N^{-1}\sum_i \mathbf{1}[J_i]$ and
$p_R = N^{-1}\sum_i \mathbf{1}[R_i]$, respectively, and the joint success
rate is $p_{JR} = N^{-1}\sum_i \mathbf{1}[J_i \land R_i]$.
We measure their association using the binary $\phi$ coefficient, termed
\textit{Judgment--Action Association} (JAA):
\begin{equation}
\mathrm{JAA}
=
\phi_{J,R}
=
\frac{p_{JR}-p_Jp_R}
{\sqrt{p_J(1-p_J)\,p_R(1-p_R)}}.
\label{eq:jaa}
\end{equation}

JAA measures whether correct diagnosis and successful repair tend to occur on the same instances. Positive values indicate more frequent co-occurrence than expected under independence, values near zero indicate little instance-level association, and negative values indicate less frequent co-occurrence. All instances remain in the fixed benchmark denominator, including invalid or failed repair outputs; JAA is undefined when either $J$ or $R$ has zero variance. Together with the task-specific performance metrics, JAA provides an instance-level characterization of \textit{Judgment--Action Coherence}.

%% file: sections/4_experiments.tex
\section{Experiments}
\label{sec:experiments}

\begin{table*}[t]

    \vspace{-10mm}
    \centering
    \caption{
    Main results across the four evaluation tasks and cross-task analysis.
    $\uparrow$ ($\downarrow$) indicates that higher (lower) is better. 
    -- indicates that JAA is undefined because no instance satisfies $J_i=1$.
    }
    \label{tab:main_results}

    \small
    \setlength{\tabcolsep}{4.0pt}
    \renewcommand{\arraystretch}{1.08}

    \resizebox{\textwidth}{!}{
    \begin{tabular}{@{}lcccccccccc@{}}
\toprule

\multirow{2}{*}{\textbf{Model}}
& \multicolumn{2}{c}{\textbf{Aesthetic Scoring}}
& \multicolumn{4}{c}{\textbf{Aesthetic Diagnosis}}
& \multicolumn{1}{c}{\textbf{Aesthetic Repair}}
& \multicolumn{2}{c}{\textbf{Text-to-UI Generation}}
& \multirow{2}{*}{
    \raisebox{-1em}{
        \makecell{\textbf{Judgment--Action}\\\textbf{Association ($\phi$)$\uparrow$}}
    }
}
\\

\cmidrule(lr){2-3}
\cmidrule(lr){4-7}
\cmidrule(lr){8-8}
\cmidrule(lr){9-10}

& \textbf{SRCC}$\uparrow$
& \textbf{MAE}$\downarrow$
& \textbf{Det. BAcc}$\uparrow$
& \textbf{Attr. Macro-J}$\uparrow$
& \textbf{Loc. F1@0.5}$\uparrow$
& \textbf{ECS}$\uparrow$
& \textbf{Repair Pass}$\uparrow$
& \textbf{Cal. Aes. Judge}$\uparrow$
& \textbf{Pairwise WR}$\uparrow$
&
\\

\midrule

\rowcolor{blue!5}
\multicolumn{11}{c}{
    \textit{\textbf{Frontier Models}}
} \\
\midrule

\modellogo{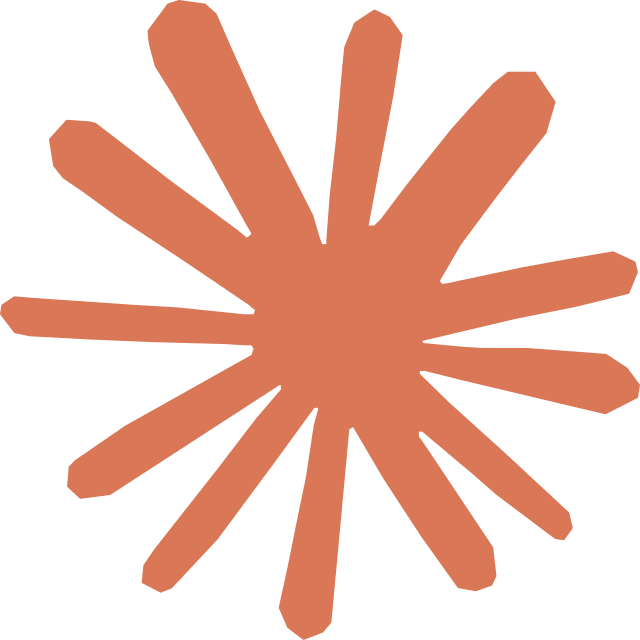}
Claude Opus 5
& 0.56
& \cellcolor{TableHeader}\textbf{0.81}
& \cellcolor{TableHeader}\textbf{91.10}
& \cellcolor{TableHeader}\textbf{58.90}
& \cellcolor{TableHeader}\textbf{36.70}
& \cellcolor{TableHeader}\textbf{24.70}
& \cellcolor{TableHeader}\textbf{75.60}
& 3.13
& 71.50
& 0.32 \\

\modellogo{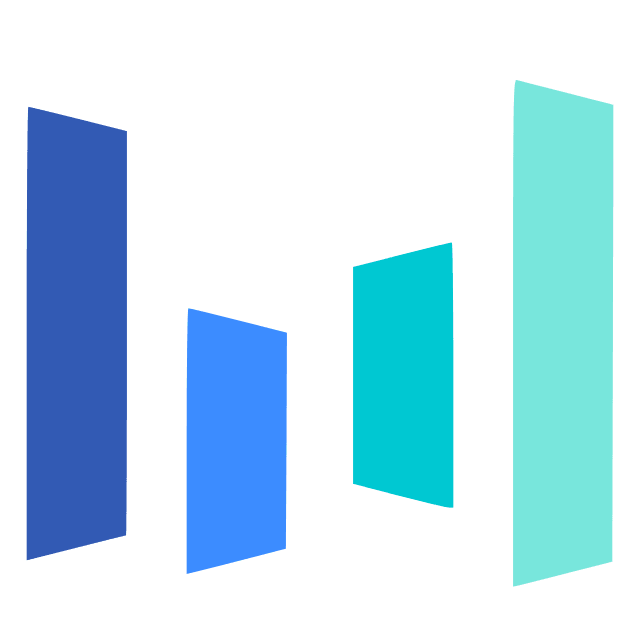}
Doubao-Seed-2.1-Pro
& 0.50
& 1.10
& 79.60
& 36.20
& 7.60
& 4.10
& 45.00
& 2.78
& 62.90
& 0.31 \\

\modellogo{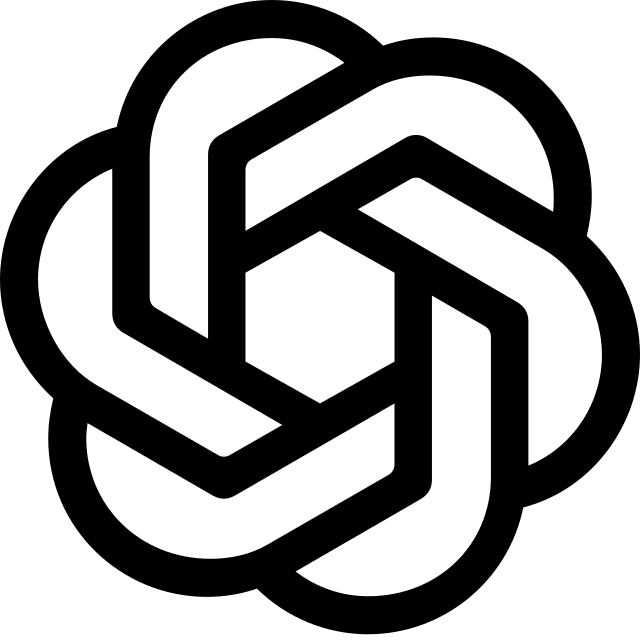}
GPT-5.6 Sol
& 0.55
& 0.90
& 87.60
& 55.40
& 28.20
& 19.40
& 65.50
& 3.17
& 78.00
& 0.38 \\

\modellogo{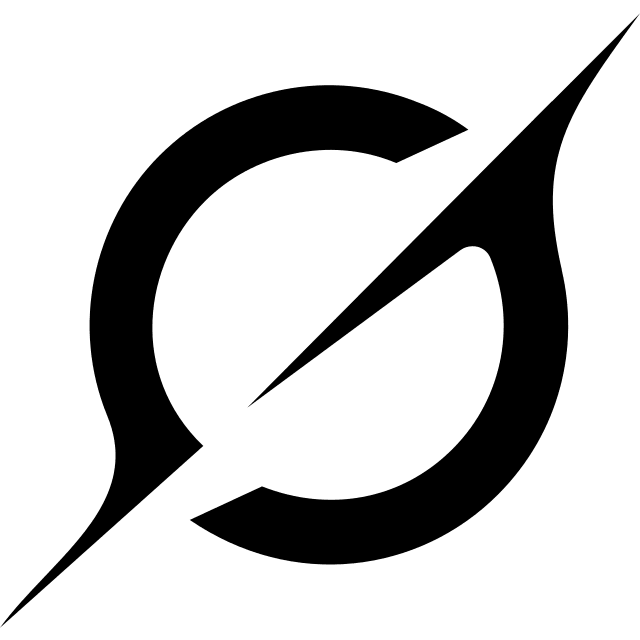}
Grok 4.6
& 0.54
& 0.95
& 81.20
& 42.70
& 6.50
& 2.90
& 70.00
& 3.08
& 62.50
& 0.25 \\

\modellogo{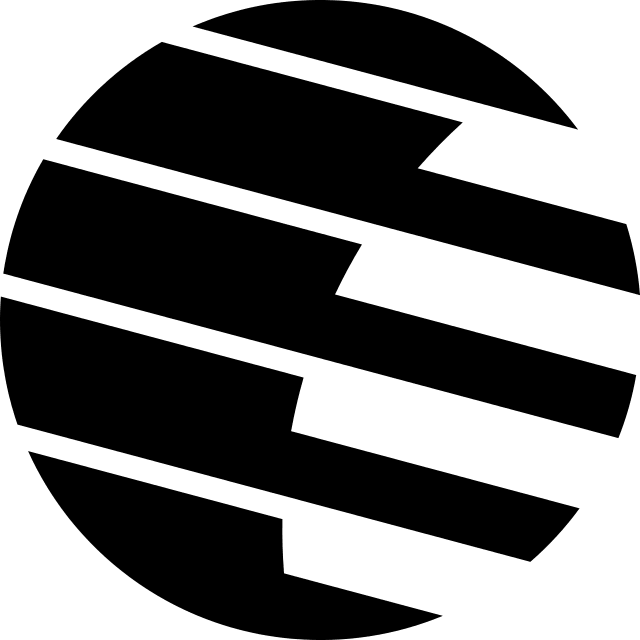}
Kimi-K3
& 0.57
& 1.10
& 83.80
& 43.30
& 26.70
& 15.20
& 57.30
& \cellcolor{TableHeader}\textbf{3.18}
& \cellcolor{TableHeader}\textbf{79.50}
& \cellcolor{TableHeader}\textbf{0.44} \\

\modellogo{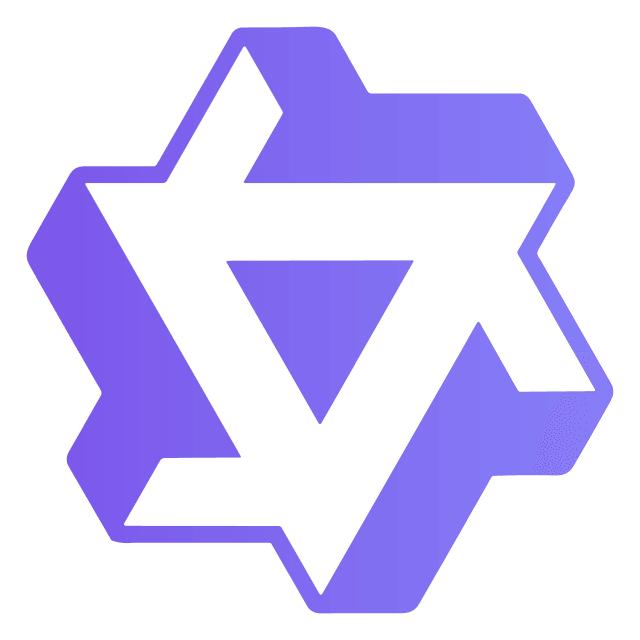}
Qwen3.7-Plus
& \cellcolor{TableHeader}\textbf{0.57}
& 1.25
& 60.80
& 22.00
& 1.60
& 0.90
& 34.20
& 3.09
& 57.80
& 0.07 \\

\modellogo{figures/logos/qwen.png}
Qwen3.8-Max
& 0.54
& 1.15
& 79.50
& 40.10
& 20.50
& 9.40
& 54.70
& 3.04
& 78.20
& 0.30 \\

\midrule

\rowcolor{blue!5}
\multicolumn{11}{c}{
    \textit{\textbf{Open-Weight Models}}
} \\
\midrule

\modellogo{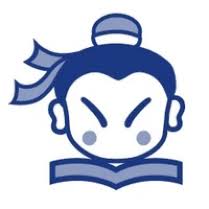}
InternVL3.5-8B-Instruct
& 0.23
& 1.46
& 43.20
& 1.30
& 0.00
& 0.00
& 0.91
& 1.75
& 21.40
& $-0.01$ \\

\modellogo{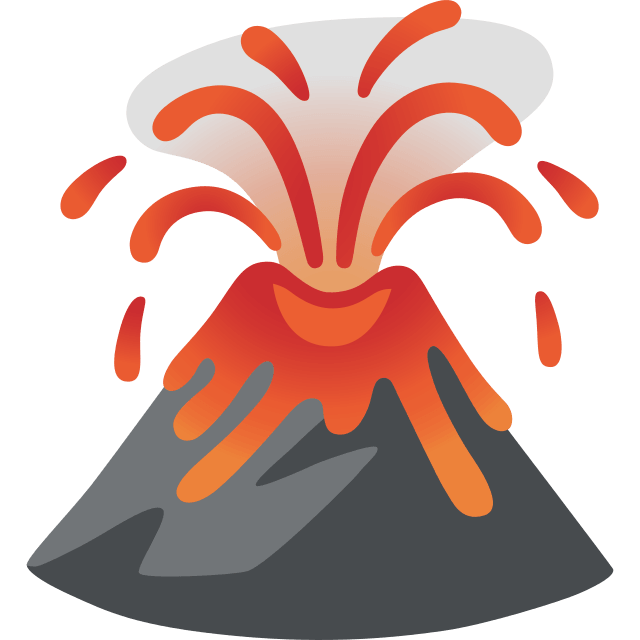}
LLaVA-OneVision2-8B-Instruct
& 0.45
& 0.94
& 40.90
& -2.90
& 0.00
& 0.00
& 0.30
& 1.11
& 12.50
& 0.00 \\

\modellogo{figures/logos/qwen.png}
Qwen3-VL-8B-Instruct
& 0.45
& 1.56
& 58.70
& 8.70
& 1.20
& 0.80
& 0.61
& 1.84
& 28.10
& $-0.01$ \\

\modellogo{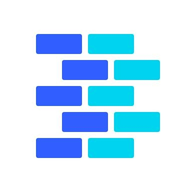}
MiniCPM-V-4.6
& 0.05
& 1.65
& 48.40
& 3.10
& 0.00
& 0.00
& 0.15
& 0.53
& 5.30
& -- \\

\midrule

\rowcolor{blue!5}
\multicolumn{11}{c}{
    \textit{\textbf{Expert Models for Image Aesthetics Assessment}}
} \\
\midrule

\modellogo{figures/logos/llava-color.png}
AesExpert-7B
& 0.13
& 1.12
& 16.70
& -0.40
& 0.00
& 0.00
& 0.00
& 0.16
& 1.40
& -- \\

\bottomrule
    \end{tabular}
    }
\end{table*}

\subsection{Experimental Setup}
\label{sec:exp_setup}

We evaluate a diverse set of multimodal models spanning frontier models, open-weight general-purpose models, and a specialized model for image aesthetics assessment. The frontier models include Claude Opus 5, Doubao-Seed-2.1-Pro, GPT-5.6 Sol, Grok 4.6, Kimi-K3, Qwen3.7-Plus, and Qwen3.8-Max. For open-weight models, we evaluate InternVL3.5-8B-Instruct~\citep{wang2025internvl3}, LLaVA-OneVision2-8B-Instruct~\citep{an2026llava}, Qwen3-VL-8B-Instruct~\citep{bai2025qwen3}, and MiniCPM-V-4.6~\citep{cui2026minicpm}. We additionally include AesExpert-7B~\citep{huang2024aesexpert}, a specialized model for image aesthetics assessment. All models are evaluated under the same task-specific protocol. More details are provided in Appendix~\ref{app:experimental_details}.

\subsection{Main Results}
\label{sec:main_results}
\textbf{\ding{182} Current models capture overall aesthetic quality to some extent, but fine-grained principle-level diagnosis remains a major bottleneck.}
In Aesthetic Scoring, frontier models show moderate agreement with professional designers, suggesting that they can partially capture the overall aesthetic quality of a UI. However, performance drops substantially when evaluation moves from holistic assessment to identifying specific aesthetic problems. Models can often distinguish the degraded interface, yet still struggle to determine \emph{which} aesthetic principle is violated and \emph{where} the problem occurs, with the best exact diagnosis-chain success reaching only 24.7\%. These results indicate that progressing from coarse aesthetic perception to precise, principle-grounded visual diagnosis remains a key limitation of current models.

\textbf{\ding{183}  Aesthetic judgment and repair are positively associated, yet a clear judgment--action gap remains.}
Most frontier models exhibit positive Judgment--Action Association, with JAA values ranging from 0.07 to 0.44, indicating that correct principle-level diagnosis tends to coincide with successful repair on the same instances. However, this alignment remains incomplete: successful repairs can occur without correct explicit judgment, while correct judgments do not always lead to successful repairs. These results reveal a persistent judgment--action gap, suggesting that current models have yet to develop a reliably coherent aesthetic capability that consistently connects principle-level understanding with corrective design action.

\textbf{\ding{184} Open-ended UI generation remains challenging, and relative superiority does not imply high absolute design quality.}
In Text-to-UI Generation, models must determine layout, visual hierarchy, typography, spacing, color, and component organization directly from natural-language requirements, textual content, and visual assets, without access to a reference screenshot or an existing implementation scaffold. Although frontier models show clear differences in relative performance, their human-calibrated absolute aesthetic scores remain only 2.78--3.18 out of 5, while some models achieve Pairwise Win Rates close to 80\%. This discrepancy shows that consistently outperforming competing models does not necessarily correspond to strong absolute design quality. Overall, current models remain limited in their ability to autonomously coordinate multiple aesthetic principles and consistently synthesize high-quality interfaces in open-ended settings.

\begin{figure}
    \vspace{-10mm}
    \centering
    \includegraphics[width=1\linewidth]{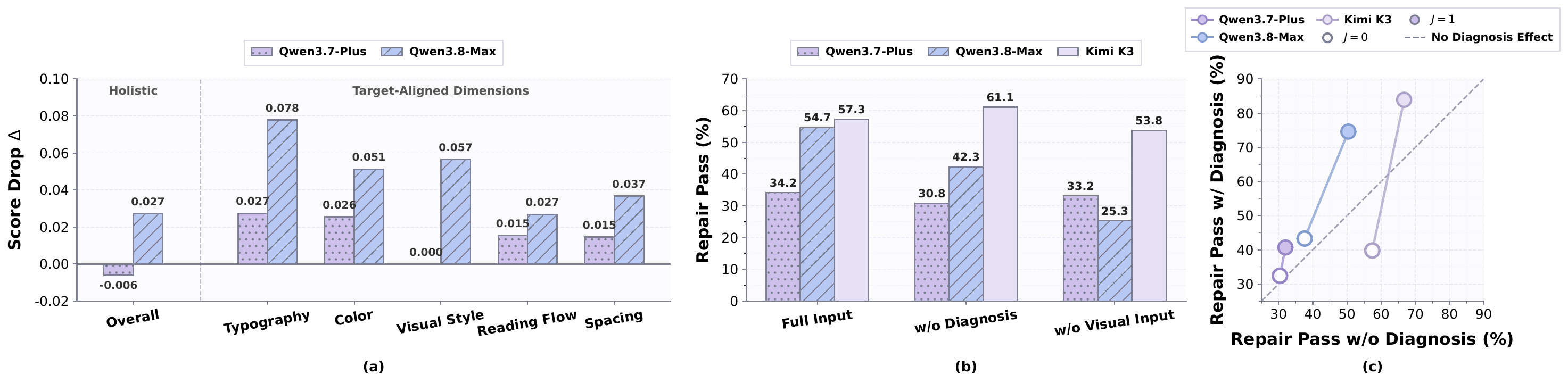}
    \caption{In-depth analyses of aesthetic sensitivity and judgment--action coherence. 
(a) Aesthetic score changes under controlled degradations. 
(b) Repair performance under input ablations. 
(c) Repair with versus without Task 2 diagnosis; points above (below) the diagonal indicate positive (negative) diagnosis effects.}
    \label{fig:task1_task3}
\end{figure}

\subsection{In-depth Analysis}
\label{sec:deep_analysis}

\ding{182} \textbf{Current models’ aesthetic scores often fail to distinguish controlled changes in specific aesthetic attributes.}
We use the controlled degradation samples from Aesthetic Diagnosis to evaluate Qwen3.7-Plus and Qwen3.8-Max on aesthetic scoring. As shown in Figure~\ref{fig:task1_task3} (a), for 83.5\% and 86.4\% of samples, respectively, the overall score remains unchanged after degradation, with average changes of only $-0.006$ and $0.027$. Even on dimensions directly corresponding to the injected defects, most scores remain unchanged. Qwen3.8-Max shows relatively higher local sensitivity: typography degradation reduces the corresponding score by $0.078$ on average, compared with $0.027$ for Qwen3.7-Plus, yet both changes remain small relative to the 1--5 scale. These results indicate that current models can respond to some local aesthetic changes, but their principle-level sensitivity remains weak and rarely propagates to overall aesthetic judgment.

\ding{183} \textbf{Current models rely on HTML-based shortcuts for UI repair instead of fully grounding their decisions in visual input.}
We isolate the contribution of visual input by removing screenshots while keeping the diagnosis, HTML, evaluation, and GT unchanged. As shown in Figure~\ref{fig:task1_task3} (b), repair Pass drops only from 34.2\% to 33.2\% for Qwen3.7-Plus and from 57.3\% to 53.8\% for Kimi-K3, suggesting that many repairs can be completed mainly from the structural, stylistic, and attribute information in the diagnosis and HTML. In contrast, Qwen3.8-Max drops sharply from 54.7\% to 25.3\%, indicating stronger reliance on explicit visual grounding. This model-level variation shows that Task 3 is not driven purely by visual understanding: some models can exploit executable HTML structure to bypass sufficient modeling of the page's visual state, forming an HTML-based shortcut.

\begin{figure}[t]
    \centering
\begin{minipage}[t]{0.53\textwidth}
    \centering
    \vspace{0pt}
    \includegraphics[width=\linewidth]{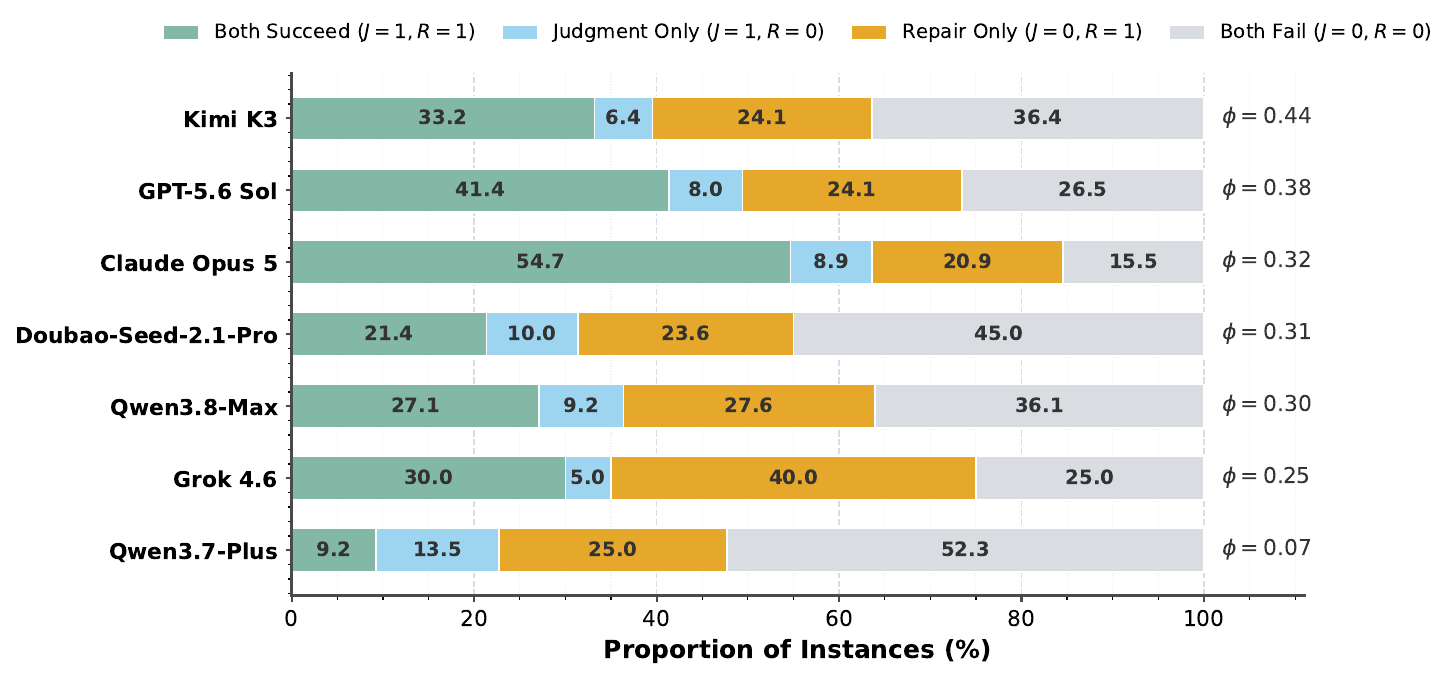}
    \caption{
    Decomposition of judgment--action outcomes over the 660 paired
    Diagnosis--Repair instances. Each bar separates
    $(J{=}1,R{=}1)$, $(J{=}1,R{=}0)$, $(J{=}0,R{=}1)$, and
    $(J{=}0,R{=}0)$ outcomes; JAA ($\phi$) summarizes their
    instance-level association.
    }
    \label{fig:ja_decomposition}
    \label{fig:jaa_outcomes}
\end{minipage}
\hfill
\begin{minipage}[t]{0.44\textwidth}
    \vspace{0pt}
    \centering
    \captionof{table}{
    Text-to-UI Generation with the Claude Code harness.
    $\Delta$Rank compares Pairwise WR rankings with the same seven models in the main evaluation.
    }
    \label{tab:harness_task4}

    \small
    \setlength{\tabcolsep}{3.2pt}
    \renewcommand{\arraystretch}{1.06}

    \resizebox{\linewidth}{!}{
    \begin{tabular}{@{}lccc@{}}
        \toprule
        \textbf{Model}
        & \makecell{\textbf{Cal. Aes.}\\\textbf{Judge}$\uparrow$}
        & \makecell{\textbf{Pairwise}\\\textbf{WR}$\uparrow$}
        & \makecell{\textbf{$\Delta$Rank}\\\textbf{vs. Main}} \\
        \midrule

        \modellogo{figures/logos/kimi.png}
        Kimi-K3
        & \cellcolor{TableHeader}\textbf{3.13}
        & \cellcolor{TableHeader}\textbf{65.86}
        & -- \\

        \modellogo{figures/logos/openai.png}
        GPT-5.6 Sol
        & 3.09
        & 55.91
        & \textcolor{RankUp}{$\uparrow\,1$} \\

        \modellogo{figures/logos/qwen.png}
        Qwen3.7-Plus
        & 3.10
        & 55.38
        & \textcolor{RankUp}{$\uparrow\,4$} \\

        \modellogo{figures/logos/qwen.png}
        Qwen3.8-Max
        & 3.12
        & 51.34
        & \textcolor{RankDown}{$\downarrow\,2$} \\

        \modellogo{figures/logos/claude.png}
        Claude Opus 5
        & 3.09
        & 48.66
        & \textcolor{RankDown}{$\downarrow\,1$} \\

        \modellogo{figures/logos/bytedance.png}
        Doubao-Seed-2.1-Pro
        & 3.06
        & 40.86
        & \textcolor{RankDown}{$\downarrow\,1$} \\

        \modellogo{figures/logos/grok.png}
        Grok 4.6
        & 3.03
        & 31.99
        & \textcolor{RankDown}{$\downarrow\,1$} \\

        \bottomrule
    \end{tabular}
    }
\end{minipage}
\end{figure}

\ding{184} \textbf{Where does the Judgment--Action Gap come from?}
To unpack the scalar JAA score, we decompose each paired Diagnosis--Repair
instance into four outcomes according to judgment correctness $J$ and repair
success $R$. Figure~\ref{fig:ja_decomposition} reveals substantial off-diagonal
mass across models: correct judgments may fail to yield successful repairs,
while many successful repairs occur without correct explicit judgment.
This asymmetry is particularly pronounced for some models, showing that strong
repair performance can arise without correspondingly reliable aesthetic
understanding. The results make the Judgment--Action Gap explicit: judgment and
action are associated, but do not form a stable instance-level correspondence.

\ding{185} \textbf{Correct aesthetic diagnosis substantially improves downstream repair.} We group samples by judgment correctness, where $J=1$ denotes correct degraded-UI identification and exact matching of the affected aesthetic dimensions, and compare Aesthetic Repair with and without the model's own diagnosis. As shown in Figure~\ref{fig:task1_task3} (c), when $J=1$, the diagnosis improves repair by $+8.7$, $+24.2$, and $+17.2$ percentage points for Qwen3.7-Plus, Qwen3.8-Max, and Kimi-K3, respectively. When $J=0$, the benefit is much weaker and model-dependent; for Kimi-K3, repair performance drops by $17.5$ points. These results indicate that accurate aesthetic judgments can provide useful design guidance, whereas unreliable judgments offer substantially less consistent downstream value.

\ding{186} \textbf{Does the harness help? A coding harness does not consistently improve UI generation quality.}
We further evaluate whether equipping frontier models with a Claude Code harness, which enables iterative code generation and editing, improves Text-to-UI Generation. As shown in Table~\ref{tab:harness_task4}, the harness does not consistently improve calibrated aesthetic scores, which remain tightly clustered around 3.0. The harness also changes the relative ordering of models: Qwen3.7-Plus rises from seventh to third among the seven evaluated models, whereas Qwen3.8-Max drops from second to fourth. This reshuffling suggests that models differ substantially in how effectively they exploit agentic coding scaffolds. More importantly, additional opportunities for code manipulation alone do not consistently translate into higher aesthetic quality, indicating that open-ended UI generation remains constrained by the model's underlying aesthetic decision-making capability rather than merely by its coding interface.

%% file: sections/5_conclusion.tex
\section{Conclusion}

We introduce \benchmark, a benchmark for jointly evaluating UI aesthetic judgment and design action across scoring, diagnosis, repair, and generation. Experiments show that current models exhibit partial aesthetic competence but still struggle with fine-grained diagnosis, open-ended generation, and consistent judgment--action alignment. We hope \benchmark supports future progress toward more coherent and reliable aesthetic capabilities for UI design.

%% file: sections/appendix_ase_rule.tex
\section{Aesthetic Design Principles}
\label{sec:ui-aesthetics}

Perceptual similarity to reference designs provides a useful measure of visual
reconstruction quality, but does not directly characterize whether an interface
follows the aesthetic principles used in professional design practice. To
provide a principle-level basis for UI aesthetic evaluation, we collaborated
with professional UI designers to co-develop a structured aesthetic design
guideline. The guideline draws on established HCI and visual-design
principles, industry design conventions, and explicit standards where
applicable, with an emphasis on properties that can be described
quantitatively from a rendered interface or its design source.

We organize the guideline into seven dimensions: color, spacing, layout,
typography, imagery, shadow, and consistency. For each dimension, we describe
the underlying design rationale and the corresponding design rules. When the
co-developed guideline specifies an explicit quantitative or programmatic
implementation, we retain it as part of the operational definition.
Table~\ref{tab:ui-checks} summarizes the rules for which such implementation
procedures are specified. Representative violating and conforming examples are
provided throughout the section.

\begin{table}[hb]
\centering
\footnotesize
\setlength{\tabcolsep}{4pt}
\renewcommand{\arraystretch}{1.2}
\caption{
Summary of aesthetic rules with explicitly specified quantitative or
programmatic checks in the co-developed professional UI design guideline.
}
\label{tab:ui-checks}
\begin{tabular}{@{}
  >{\raggedright\arraybackslash}p{0.13\linewidth}
  >{\raggedright\arraybackslash}p{0.39\linewidth}
  >{\raggedright\arraybackslash}p{0.41\linewidth}@{}}
\toprule
\textbf{Dimension}
& \textbf{Rule}
& \textbf{Implementation} \\
\midrule

color
& Primary : secondary : accent $=6:3:1$
& Measure the pixel proportion of each color role and compute its KL
divergence from the target ratio \\[2pt]

& Normal text/background contrast $\geq 4.5{:}1$;
large text $\geq 3{:}1$;
graphics and UI components $\geq 3{:}1$
& Compute the WCAG contrast ratio from foreground and background
relative luminances using Eq.~(\ref{eq:contrast}) \\

\midrule

Spacing
& Element sizes and spacing use $4$, $12$, or integer multiples of
$8$ pixels
& Compute the residual of element coordinates and dimensions modulo
$8$, while admitting $4$ and $12$ as additional values \\

\midrule

Typography
& At most one CJK and one Latin typeface per page
& Extract typeface information from the design source and compare
typeface usage \\[2pt]

& $h_{\mathrm{line}} = s + 8$ and at most five type sizes
& Extract font-size and line-height parameters from the design source \\[2pt]

& Font weights use $\{400,500,600\}$ and at most three weights per
application
& Extract \texttt{font-weight} values from the design source \\[2pt]

& At most $85$ characters per line
& OCR followed by character counting \\[2pt]

& Text in tables and lists is left-aligned, while Arabic numerals and
other numerical content are right-aligned
& Check the consistency of the left or right boundary coordinates
within a column \\

\midrule

Imagery
& Image aspect ratios belong to
$\{1,\tfrac{4}{3},\tfrac{16}{9},\tfrac{3}{4},\tfrac{9}{16}\}$
& Compute the deviation between the rendered aspect ratio and the
nearest admissible ratio \\[2pt]

& Images remain sufficiently clear to be recognizable
& Apply image-quality assessment to identified image regions \\

\midrule

Shadow
& Component shadow depth follows semantic levels $0$ to $3$
& Recognize UI components, extract their shadow parameters, and compare
them with the corresponding semantic level \\

\midrule

Consistency
& Equivalent containers maintain consistent corner radius, shadow,
and spacing
& Identify container components and compare their corresponding
parameters \\[2pt]

& Components of the same type follow one shared specification
& Identify component classes and compare parameters across instances \\

\bottomrule
\end{tabular}
\end{table}

\subsection{color}
\label{ssec:color}

color plays an important role in visual communication and in conveying
multiple types of information within an interface~\citep{bonnardel2011impact,seckler2015linking}.
A well-designed enterprise application should use color clearly and consistently
so that both functional information and product identity can be communicated
effectively.

We distinguish three color roles. The \emph{theme color} represents the
product identity and is generally associated with the brand color. It is
commonly used for primary buttons and text, important operation states, and
highlighted information. The theme color is typically extended into a tonal
scale within the same color family. \emph{Functional colors} represent
explicit information and states, including success, error, failure, warning,
and links. Their use should follow common user expectations and remain
consistent within the same product system. \emph{Neutral colors} are widely
used for text, backgrounds, borders, and dividers.

\textbf{color proportion.}
To avoid excessive color usage, we adopt a $60{:}30{:}10$
allocation. Approximately $60\%$ of the interface is assigned to the primary
color, which in enterprise applications is commonly the color of large
surface or content regions. Approximately $30\%$ is assigned to secondary
colors, which are typically neutral colors, and the remaining $10\%$ is
reserved for accent colors drawn from the functional and theme palettes.

For quantitative assessment, we estimate the empirical distribution
$\hat{p}$ of the three color roles over interface pixels and compare it with
the target distribution

\[
p^{\star} = (0.6, 0.3, 0.1).
\]

The deviation is measured using

\begin{equation}
D_{\mathrm{KL}}
\left(
\hat{p}\,\|\,p^{\star}
\right).
\label{eq:color-kl}
\end{equation}

Figure~\ref{fig:color-proportion} shows a representative violation in which
the accent color occupies an excessively large region.

\uifig{1.00}{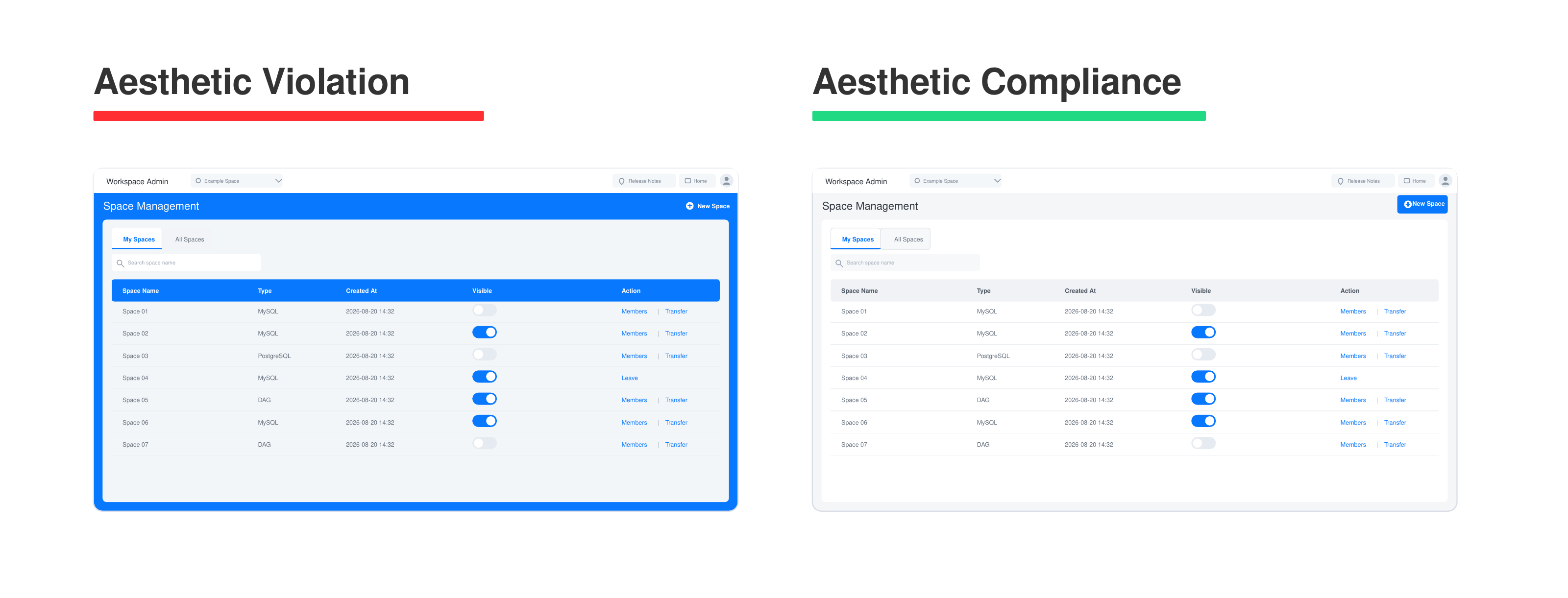}
  {\textbf{color proportion.}
   The violating interface applies the accent color to a large portion of the
   page, while the conforming interface follows a more restrained color
   allocation.}
  {fig:color-proportion}

\textbf{Contrast.}
Color contrast determines whether textual and graphical information can be
reliably perceived, as illustrated by the confirmation messages in
Figure~\ref{fig:color-contrast}. Following the WCAG AA
requirements\footnote{https://www.w3.org/TR/WCAG22/}, we adopt a minimum
contrast ratio of $4.5{:}1$ for normal text against its background. Large text
requires a minimum contrast ratio of $3{:}1$, where large text is defined as
bold text of at least $18.66$\,px or regular text of at least $24$\,px.
Graphics and user-interface components require a minimum contrast ratio of
$3{:}1$.

WCAG is a set of accessibility guidelines developed by the W3C Web
Accessibility Initiative to improve the accessibility of Web content for
users with different perceptual and cognitive abilities. For implementation,
we compute the WCAG contrast ratio as

\begin{equation}
  \mathrm{contrast}
  =
  \frac{\max(Y_f, Y_b) + 0.05}{\min(Y_f, Y_b) + 0.05},
  \label{eq:contrast}
\end{equation}

where $Y_f$ and $Y_b$ denote the foreground and background
relative luminances in $[0,1]$, respectively, computed from
linearized sRGB values following WCAG 2.2.

\uifig{0.95}{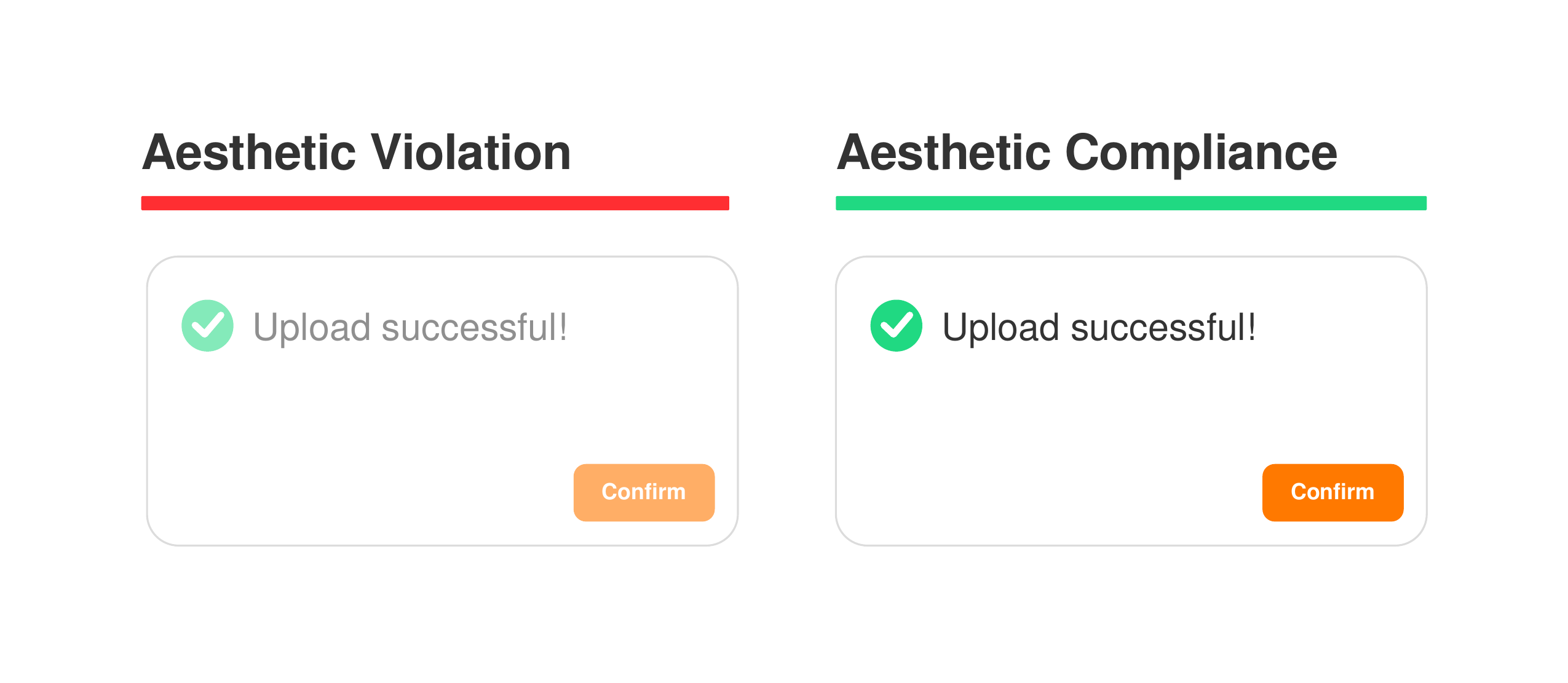}
  {\textbf{color contrast.}
   A confirmation message with insufficient foreground-background contrast,
   compared with a version satisfying the prescribed contrast requirement.}
  {fig:color-contrast}

\textbf{color semantics.}
Functional colors should conform to common semantic expectations, as colors can carry context-dependent semantic and affective associations~\citep{elliot2014color}.
Red commonly communicates urgency or intensity and is therefore suitable for
promotional actions and error states. Blue commonly conveys technology and
professionalism and is frequently used in technology-oriented products.
Green is associated with nature and safety and is commonly used for
environmental contexts or successful states. Maintaining these semantic
associations reduces unnecessary interpretation effort and makes interface
states easier to understand.
Figure~\ref{fig:color-semantics} contrasts red and green indicators for the
same success state.

\uifig{0.95}{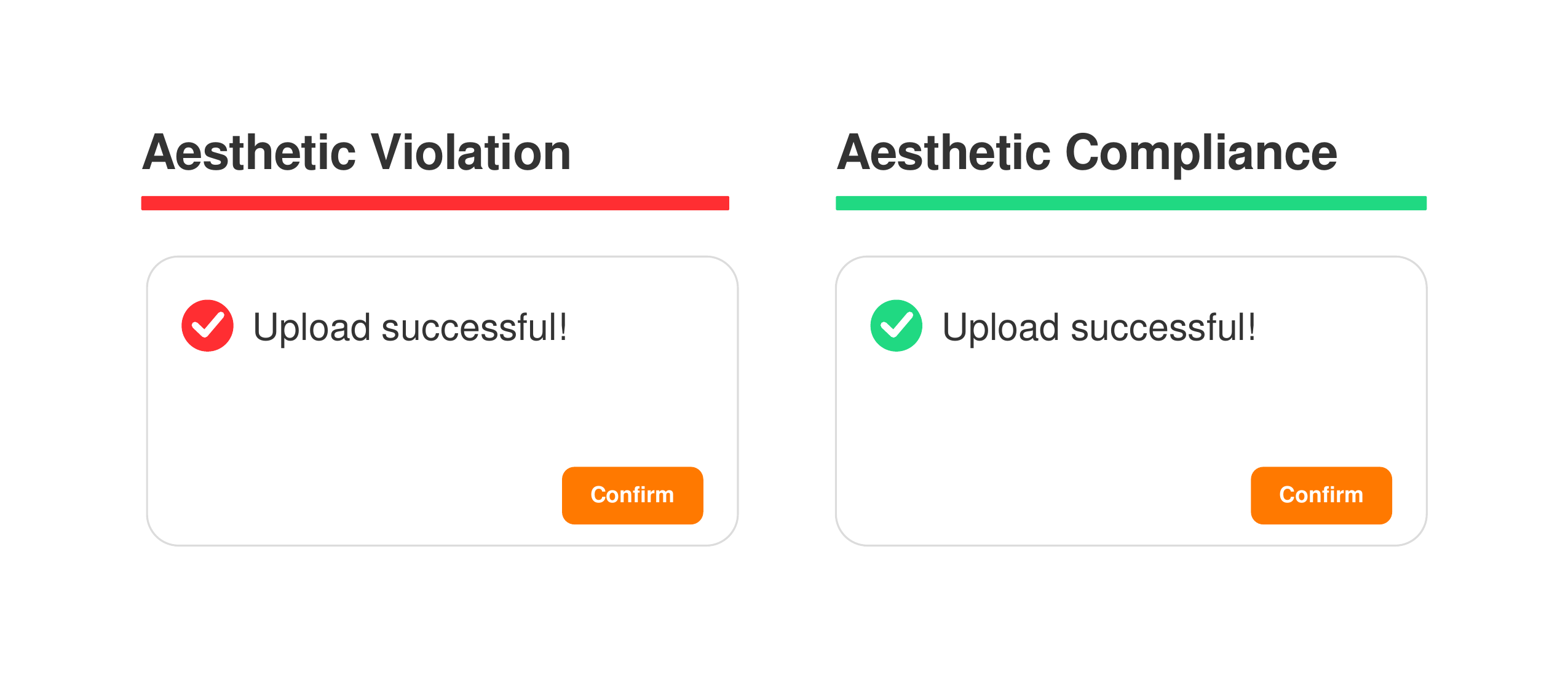}
  {\textbf{Functional color semantics.}
   The violating example uses a color whose conventional semantic meaning
   conflicts with the represented state, while the conforming example follows
   the expected color semantics.}
  {fig:color-semantics}

\subsection{Spacing}
\label{ssec:spacing}

Spacing defines the basic spatial relationships among interface elements.
Enterprise applications commonly use a grid system to establish visual order,
with $8$ pixels serving as a widely adopted base unit. An even-valued base
unit is compatible with common display environments, while a regular grid also
reduces arbitrary layout decisions and provides a shared spatial vocabulary
between design and implementation.

We therefore require interface elements to be arranged using an
$8$-pixel base grid. To support denser layouts and smaller local gaps, $4$ and
$12$ pixels are additionally admitted as valid spacing values. The resulting
spacing scale (Figure~\ref{fig:spacing-scale}) includes values such as
$4$, $8$, $12$, $16$, $24$, $32$,
$40$, $48$, $64$, $96$, and $160$ pixels.

\uifig{1.00}{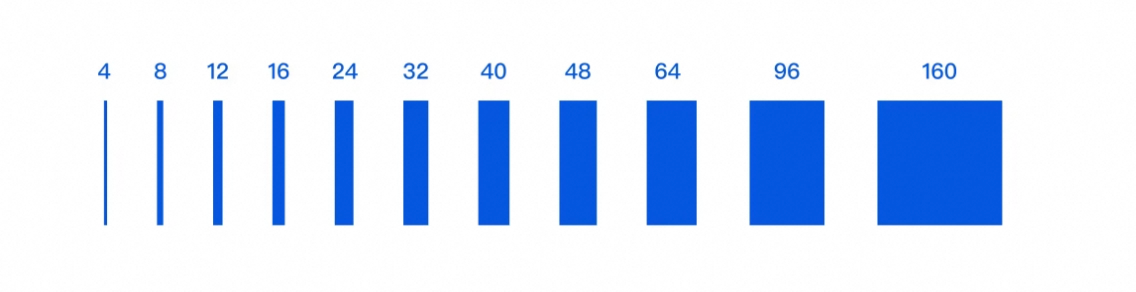}
  {\textbf{Spacing scale.}
   Representative spacing values derived from the $8$-pixel base grid,
   together with the additional $4$- and $12$-pixel increments.}
  {fig:spacing-scale}

For a set of measured element coordinates and dimensions $\{x_i\}$, we
evaluate deviation from the grid using

\begin{equation}
  \mathcal{P}
  =
  \sum_i
  \left(
  x_i \bmod 8
  \right),
  \label{eq:spacing-penalty}
\end{equation}

while additionally admitting $4$ and $12$ pixels as valid values.
Figure~\ref{fig:spacing-example} illustrates irregular gaps of
$14$, $15$, and $7$ pixels and their corresponding conforming alternatives.

\uifig{0.95}{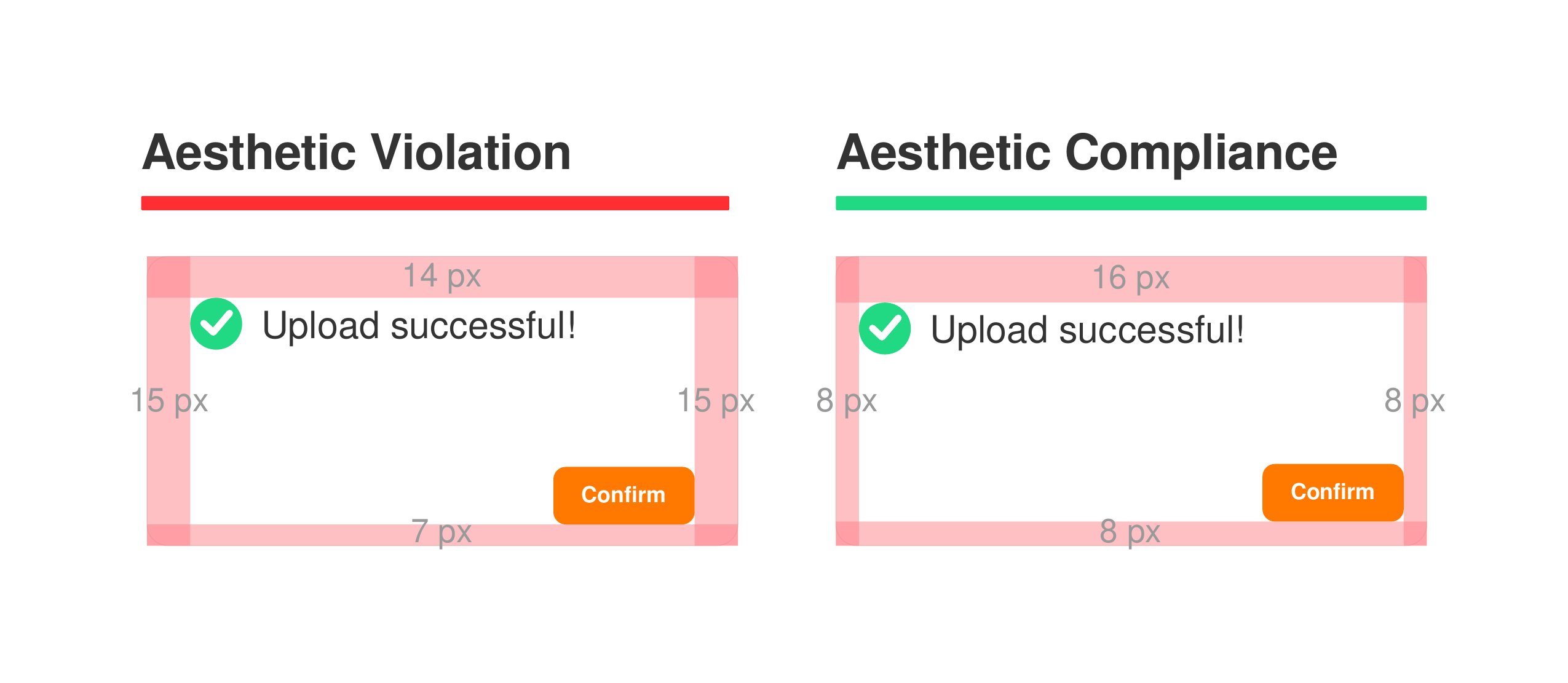}
  {\textbf{Spacing grid.}
   The violating layout contains spacing values that do not follow the
   prescribed grid, while the conforming version uses admissible spacing
   increments.}
  {fig:spacing-example}

\subsection{Layout}
\label{ssec:layout}

Layout is a fundamental part of enterprise interface design and provides the
basis for consistent interaction and visual organization. We consider three
established principles for organizing important information and actions:
Fitts' Law, Miller's Law, and the Gutenberg Diagram.

\textbf{Fitts' Law.}
Fitts' Law states that larger and closer targets can be reached more quickly and with fewer errors than smaller and more distant targets~\citep{fitts1954information}. Screen edges have a particular advantage because pointer movement cannot continue beyond the screen boundary, making controls near the edge easier to acquire~\citep{farris2001acquisition}.

At the same time, frequently used controls should not be positioned so close
to the boundary that accidental activation becomes likely. Balancing these two
considerations, we recommend that frequently used desktop controls,
especially primary actions, be positioned approximately $5\%$ to $12\%$ of the
viewport dimension from the nearest edge, as illustrated by the primary
action in Figure~\ref{fig:fitts}. Corners should be used preferentially when
appropriate.

\uifig{1.00}{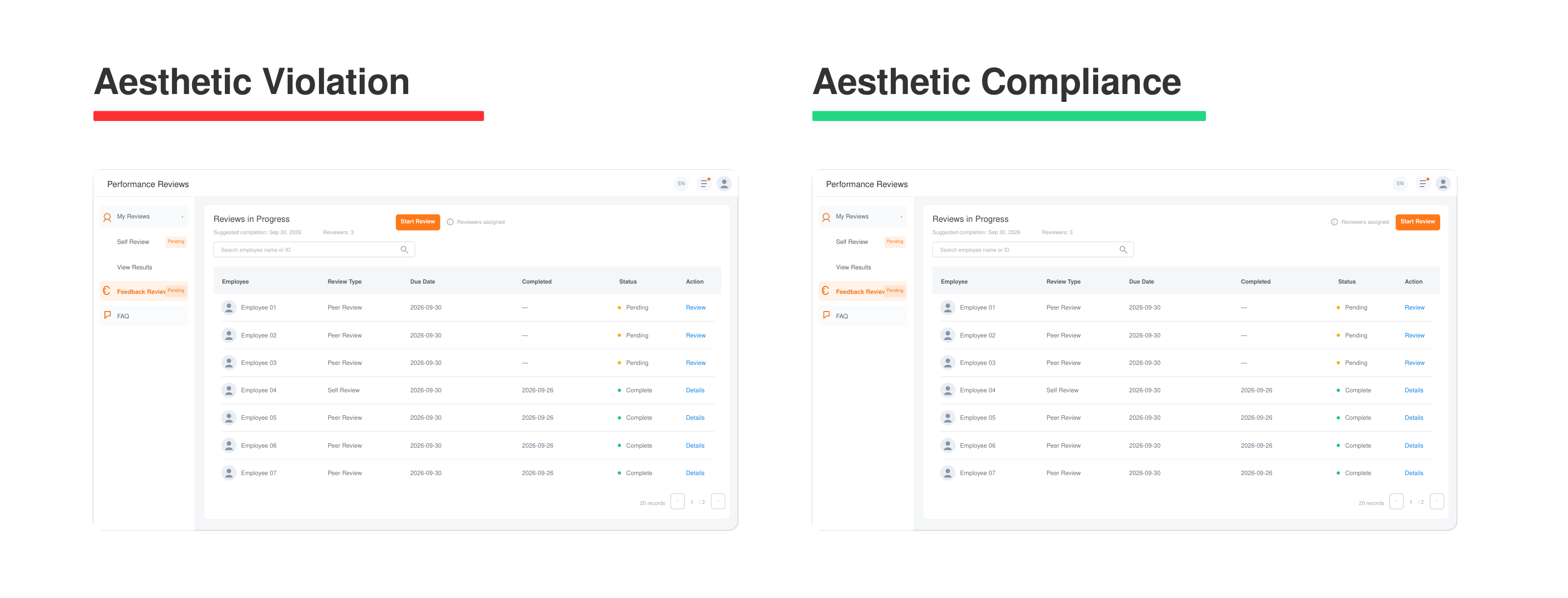}
  {\textbf{Fitts' Law.}
   A primary action placed far from the screen edge, compared with a placement
   that follows the recommended edge distance.}
  {fig:fitts}

\textbf{Miller's Law.}
Human information processing capacity is limited. The classical formulation
of Miller's Law suggests that short-term memory can simultaneously maintain
approximately five to nine items \citep{miller1956magical}. Later work
suggests that effective working-memory capacity is often closer to three to
five chunks \citep{cowan2010magical}.

For enterprise applications, we adopt the more conservative interpretation and
recommend that a group of similar content blocks contain no more than five
items. When more content is required, the blocks can be reorganized into
multiple groups or rows, as illustrated by the six-block layout in
Figure~\ref{fig:miller}.

\uifig{1.00}{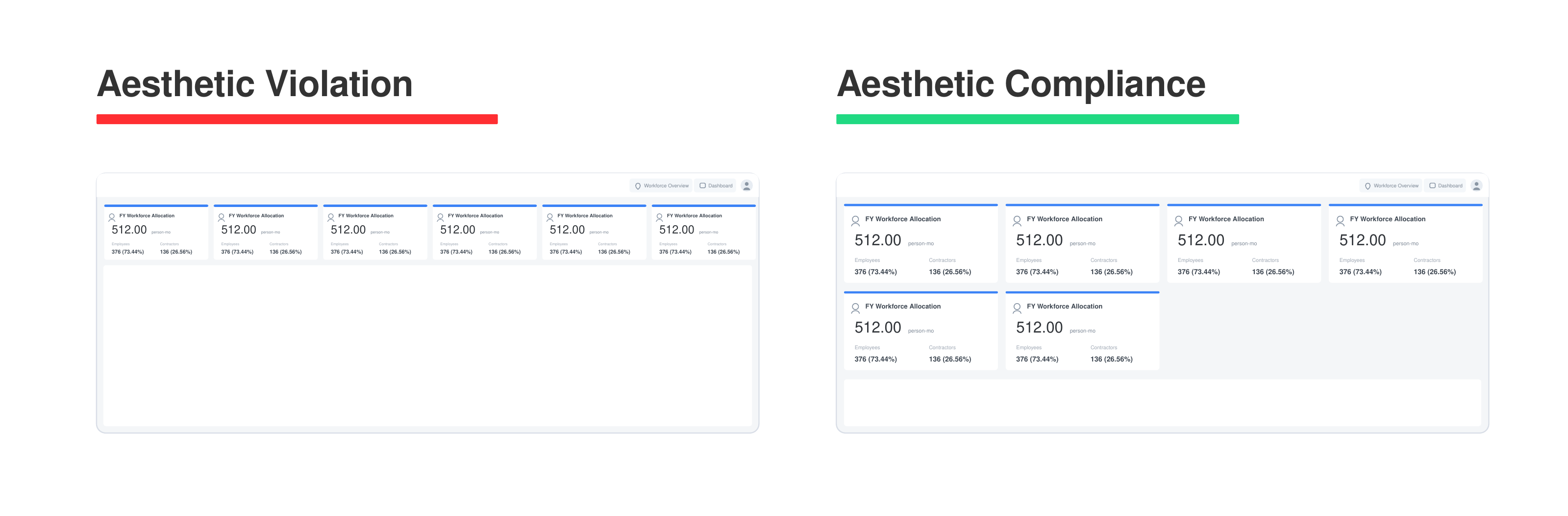}
  {\textbf{Miller's Law.}
   Six similar content blocks are presented as one group in the violating
   example, while the conforming version reorganizes them into smaller groups.}
  {fig:miller}

\textbf{Gutenberg Diagram.}
For interfaces following a left-to-right reading convention, visual attention
typically begins in the upper-left region and progresses toward the
lower-right region. The Gutenberg Diagram divides the page into four
quadrants and emphasizes the upper-left and lower-right regions as the
beginning and ending areas of the reading path.

Accordingly, we recommend placing the most important identifying or promotional
information in the upper-left region, such as logos or other high-priority
information. As illustrated by the confirmation dialog in
Figure~\ref{fig:gutenberg}, the most important operation, such as a confirmation
action, is preferentially placed in the lower-right region.

\uifig{1.00}{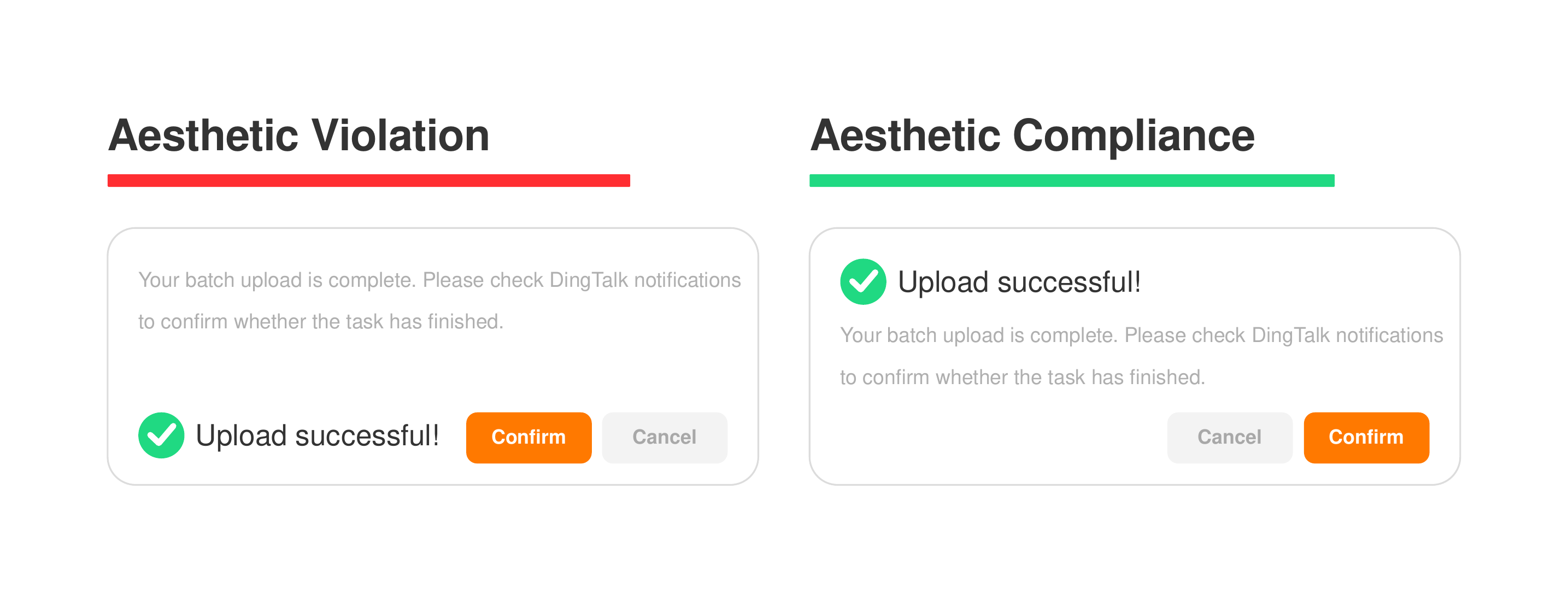}
  {\textbf{Gutenberg Diagram.}
   The violating example places important information and actions against the
   expected reading flow, while the conforming layout follows the recommended
   upper-left to lower-right organization.}
  {fig:gutenberg}

\subsection{Typography and Text Layout}
\label{ssec:typography}

Typography is one of the fundamental components of systematic interface
design. Enterprise users rely on text to understand information and complete
their work, making a coherent type system important for both reading
efficiency and task efficiency~\citep{dyson2004physical}.

\textbf{Typeface.}
For efficiency and platform compatibility, we recommend using the system
default typeface in enterprise applications. Within a single page, no more
than one CJK typeface and one Latin typeface should be used.
Figure~\ref{fig:typeface} contrasts mixed typefaces with a unified typeface
system.

This rule can be checked by extracting the typeface information from the
design source and comparing the set of typefaces used on the page.

\uifig{0.85}{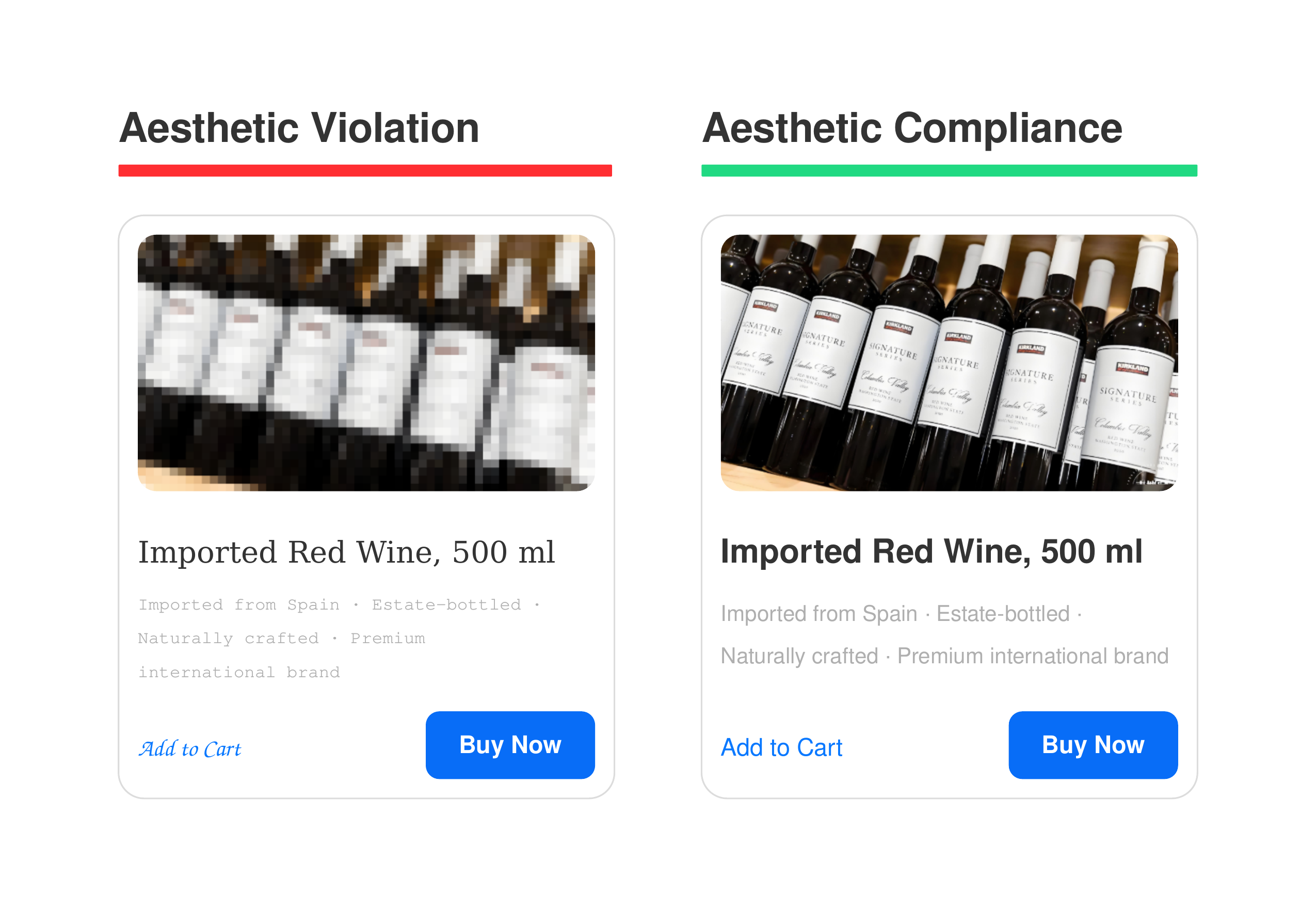}
  {\textbf{Typeface discipline.}
   The violating example mixes several unrelated typefaces within one
   interface, while the conforming version uses a unified typeface system.}
  {fig:typeface}

\textbf{Type scale and line height.}
A type scale is a regular set of font sizes used to establish textual
hierarchy. Line height defines the vertical space occupied by each line of
text. We take

\[
h_{\mathrm{line}} = 1.5s
\]

as a common accessibility-oriented reference, where $s$ denotes the font
size. However, a fixed multiplicative ratio causes the additional vertical
space to increase with font size. For large display text, this may weaken the
visual continuity between lines, particularly when several font sizes are
mixed within the same interface.

We therefore adopt an additive rule to control the spacing between lines
across font sizes, as illustrated in Figure~\ref{fig:type-scale}:

\begin{equation}
  h_{\mathrm{line}}
  =
  s + 8
  \quad \text{(pixels)}.
  \label{eq:line-height}
\end{equation}

The constant $8$ also gives line heights close to the $1.5\times$ reference
for the commonly used body-text sizes of $14$\,px and $16$\,px. To maintain a
clear reading hierarchy, no more than five distinct type sizes should be used
within one application.

The rule is checked by extracting font-size and line-height parameters from
the design source.

\uifig{0.72}{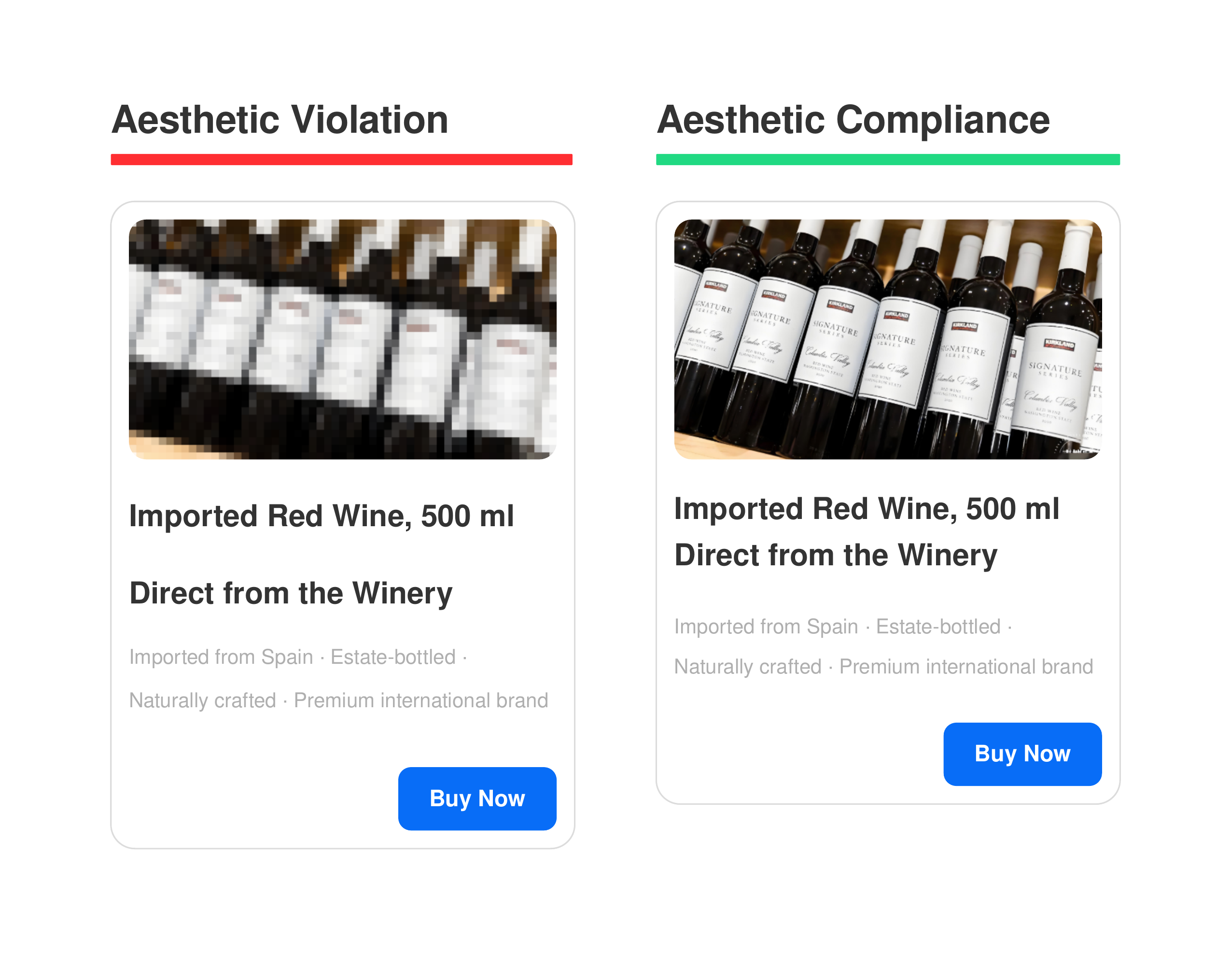}
  {\textbf{Type scale and line height.}
   The violating example uses line spacing that weakens the continuity of the
   text block, while the conforming version follows the additive line-height
   rule in Eq.~\eqref{eq:line-height}.}
  {fig:type-scale}

\textbf{Font weight.}
Font weight is an important typographic variable for expressing hierarchy and
distinguishing important content. We use $\{400,500,600\}$ as the preferred
set of weights and allow no more than three font-weight values within one
application.
The link and button labels in Figure~\ref{fig:font-weight} illustrate
consistent font weights across actions.

The implementation extracts the \texttt{font-weight} value of each text
element and checks whether the observed values follow the prescribed set and
cardinality.

\uifig{0.85}{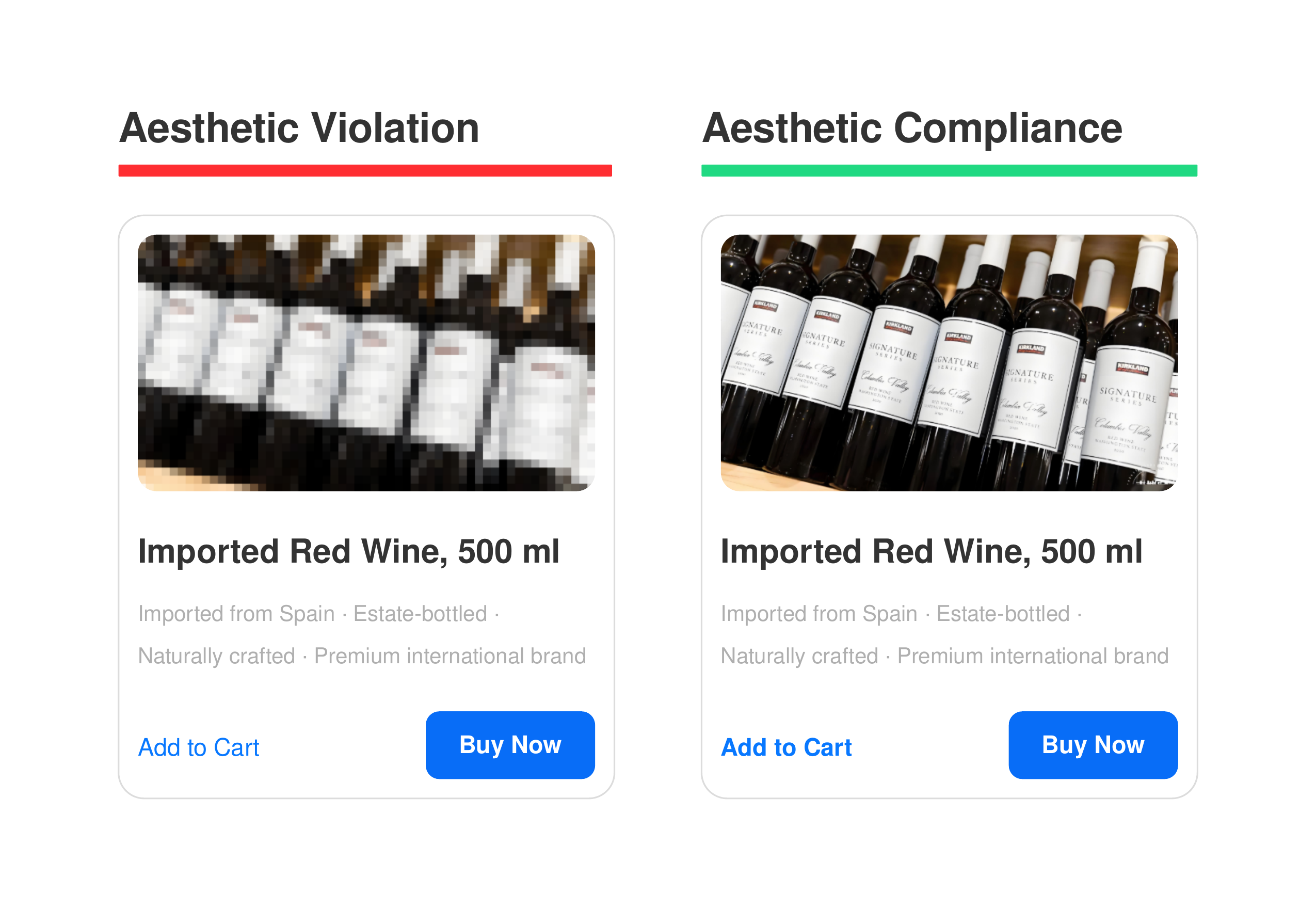}
  {\textbf{Font weight.}
   The violating example uses an inconsistent weight assignment, while the
   conforming example follows the prescribed font-weight system.}
  {fig:font-weight}

\textbf{Line length and alignment.}
Line length has been shown to affect on-screen reading and information retrieval~\citep{dyson2004physical}. To maintain reading comfort in Web applications, we recommend no more than
$85$ characters in a single line of text. This property can be checked through
OCR followed by character counting.

Text alignment and spacing can affect visual-search performance on Web pages~\citep{ling2007influence}. For tables and lists, textual content should be left-aligned, while Arabic
numerals and other numerical values should be right-aligned. This arrangement
(Figure~\ref{fig:alignment}) allows textual entries to share a stable starting
position and numerical values to share a stable ending position.

For a left-aligned column, we describe the condition

\begin{equation}
  x_{\mathrm{left}} = C,
  \label{eq:left-align}
\end{equation}

while a right-aligned column satisfies

\begin{equation}
  x_{\mathrm{right}} = W - C,
  \label{eq:right-align}
\end{equation}

where $W$ denotes the column width and $C$ denotes the corresponding inset.

\uifig{1.00}{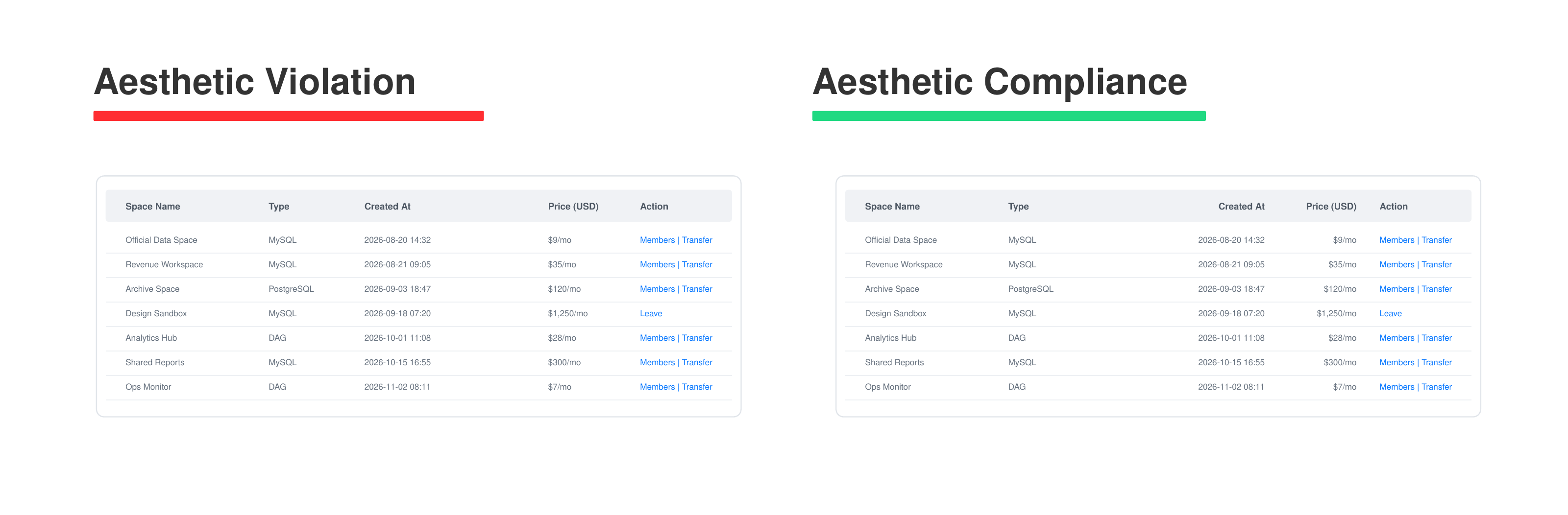}
  {\textbf{Text and numerical alignment.}
   Numerical values are left-aligned in the violating table, while the
   conforming version right-aligns numerical content and left-aligns text.}
  {fig:alignment}

\subsection{Imagery}
\label{ssec:imagery}

Images increase the dimensionality of information conveyed by an interface and
can provide visual explanations that complement textual content. Their
presentation quality is primarily affected by geometric proportion and visual
clarity.

\textbf{Aspect ratio.}
Different image proportions support different presentation scenarios.
Excluding full-width and full-screen banner images, we identify several
commonly used aspect ratios.

A $1{:}1$ image provides a simple composition with strong subject presence and
is commonly used for products, avatars, and close-up content. A $4{:}3$ image
provides a compact frame that is relatively easy to compose. A $16{:}9$ image
provides a wider field of view and is widely used for video-oriented content.

The complete admissible set specified in our guideline is

\begin{equation}
  \mathcal{R}
  =
  \left\{
  1,\,
  \frac{4}{3},\,
  \frac{16}{9},\,
  \frac{3}{4},\,
  \frac{9}{16}
  \right\}.
  \label{eq:aspect-set}
\end{equation}

For an image with width $w$ and height $h$, its aspect ratio is compared with
the nearest admissible value. We write the criterion as

\begin{equation}
  \min_{r \in \mathcal{R}}
  \left|
  \frac{w}{h} - r
  \right|
  < 0.05.
  \label{eq:aspect}
\end{equation}

Figure~\ref{fig:image-ratio} illustrates the visual distortion caused by
compressing an image vertically.

\uifig{0.85}{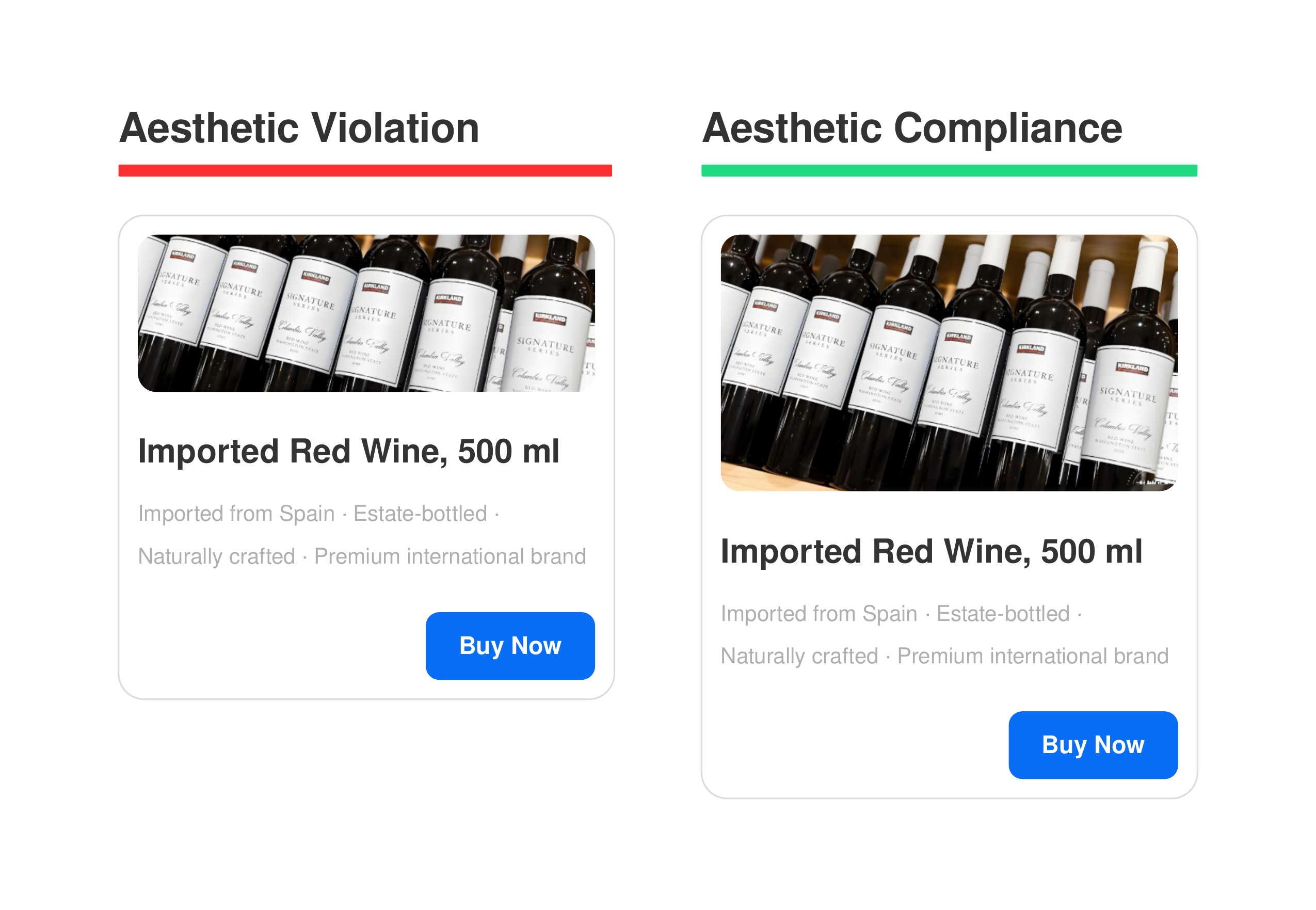}
  {\textbf{Image aspect ratio.}
   The violating example distorts the image proportion, while the conforming
   version preserves an admissible aspect ratio.}
  {fig:image-ratio}

\textbf{Image clarity.}
Recognizability is treated as a basic requirement for image presentation.
Images should remain sufficiently clear for users to identify their visual
content. The pixelated image in Figure~\ref{fig:image-clarity} illustrates how
reduced clarity obscures visual detail. We propose applying image-quality
assessment algorithms to image regions identified in the interface.

Several commonly used image-quality metrics are listed as references for this
purpose. Their definitions and properties are summarized in
Table~\ref{tab:iqa}.

\uifig{0.85}{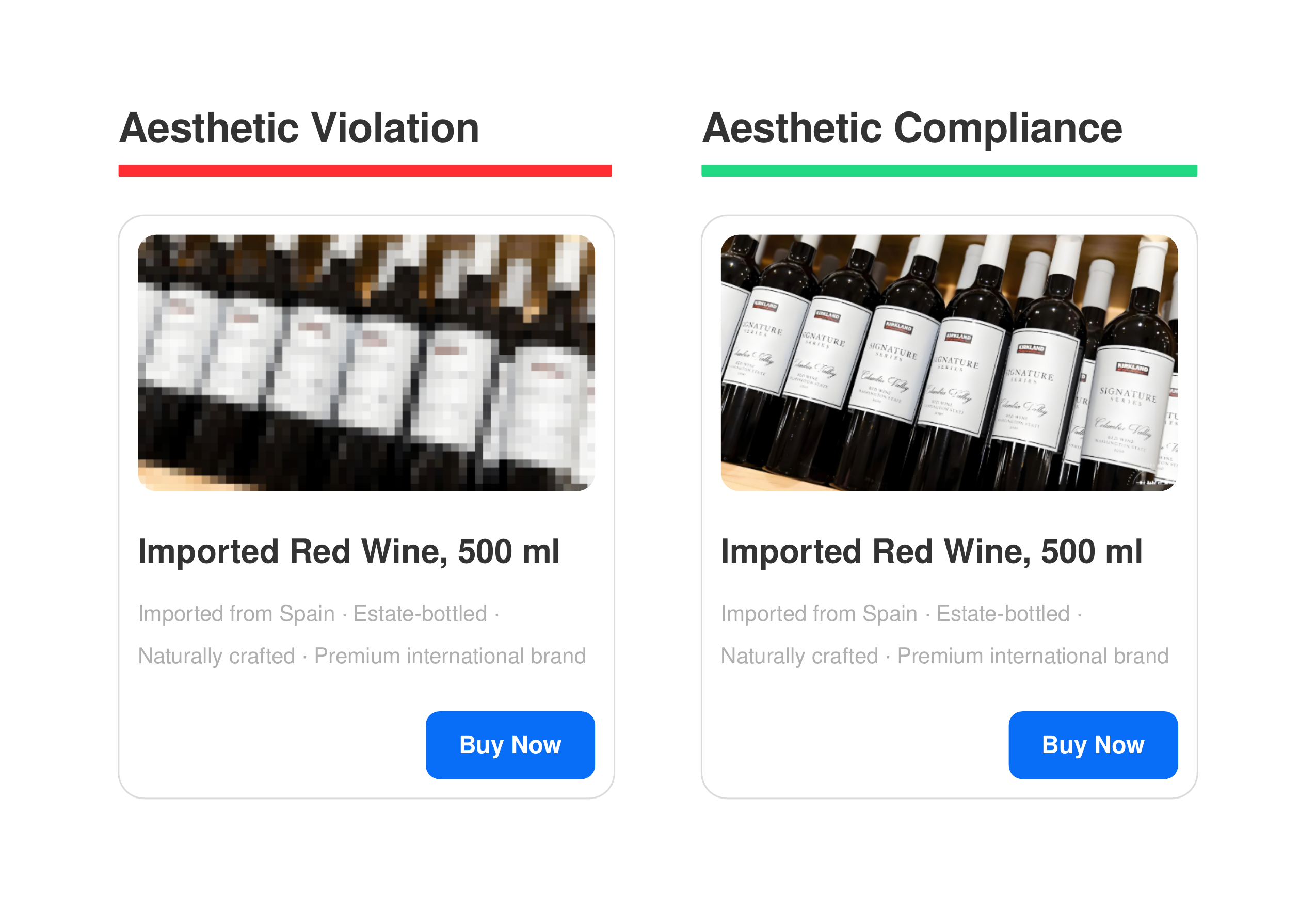}
  {\textbf{Image clarity.}
   The image in the violating example is insufficiently clear, while the
   conforming version preserves recognizable visual detail.}
  {fig:image-clarity}

\begin{table}[t]
\centering
\footnotesize
\setlength{\tabcolsep}{5pt}
\renewcommand{\arraystretch}{1.2}
\caption{
Image-quality metrics included in our professional UI design guideline as
references for evaluating image clarity.
}
\label{tab:iqa}
\begin{tabular}{@{}
  >{\raggedright\arraybackslash}p{0.16\linewidth}
  >{\raggedright\arraybackslash}p{0.78\linewidth}@{}}
\toprule
\textbf{Metric}
& \textbf{Definition and properties} \\
\midrule

PSNR
& Peak signal-to-noise ratio is a full-reference image-quality metric that
compares the maximum possible signal power with the power of corrupting noise.
It is measured in decibels, and larger values indicate less distortion.
Because it is based on pixel-level error, it does not explicitly model the
visual characteristics of human perception. \\

SSIM
& Structural similarity evaluates image similarity from luminance, contrast,
and structure~\citep{wang2004image}. Its value lies in $[-1,1]$, with larger values indicating less
distortion. In practical computation, an image can be divided into local
windows. For $N$ windows, the average structural similarity is

\[
\mathrm{MSSIM}(X,Y)
=
\frac{1}{N}
\sum_{i=1}^{N}
\mathrm{SSIM}(x_i,y_i).
\]
\\

IFC
& The information fidelity criterion evaluates image quality using natural
scene statistics and characteristics of the human visual system. It measures the mutual information between a test image and a reference image~\citep{sheikh2005information}. \\

VIF
& Visual information fidelity extends the information-based formulation of IFC and focuses on the amount of visual information lost between a test image and its reference~\citep{sheikh2006image}. \\

MSE / RMSE
& Mean squared error and root mean squared error measure pixel-level error
between corresponding images. They are simple objective measures but do not
explicitly account for characteristics of human visual perception. \\

\bottomrule
\end{tabular}
\end{table}

\subsection{Shadow}
\label{ssec:shadow}

Shadows originate from the physical relationship between objects and surfaces
at different distances. User interfaces reproduce this visual cue to
communicate the height and layer relationships among elements~\citep{creager2016toward}. Components at
different semantic layers therefore use different shadow properties.

We divide interface elements into four semantic levels.
\textbf{Level~0} represents elements resting directly on the base surface.
These include buttons, inputs, search fields, tags, tables, links, pagination
controls, steps, breadcrumbs, switches, radio buttons, checkboxes, and
progress bars. Their projection coincides with the component itself, so no
visible shadow is used.

\textbf{Level~1} represents low-level elevation and is used for navigation
elements and card hover states. \textbf{Level~2} represents components that
expand from elements on the base surface and remain associated with them,
including drop-down containers and drawers. \textbf{Level~3} represents
high-level components used for prominent prompts and operations, including
dialogs, modal windows, and toast notifications.
For example, Figure~\ref{fig:shadow} shows how a shadow distinguishes an
expanded date picker from the base surface.

The shadow parameters specified in our guideline are summarized in
Table~\ref{tab:shadow-levels}. For Levels~1 to~3, the same offset magnitude is
applied in five directions. These directions are $(d,d)$, $(0,-d)$, $(d,0)$,
$(0,d)$, and $(-d,0)$. The spread value is zero in all cases.

\begin{table}[t]
\centering
\footnotesize
\setlength{\tabcolsep}{6pt}
\renewcommand{\arraystretch}{1.25}
\caption{
Shadow hierarchy specified in our professional design guideline, following
the Fusion Design shadow scale.
}
\label{tab:shadow-levels}
\begin{tabular}{@{}c
  >{\raggedright\arraybackslash}p{0.43\linewidth}
  c c l@{}}
\toprule
\textbf{Level}
& \textbf{Typical components}
& \textbf{Offset $d$}
& \textbf{Blur}
& \textbf{color} \\
\midrule

0
& Buttons, inputs, search fields, tags, tables, links, pagination, steps,
breadcrumbs, switches, radio buttons, checkboxes, and progress bars
& $0$
& $0$
& \texttt{rgba(0,0,0,0)}
\\

1
& Navigation and card hover states
& $1$\,px
& $3$\,px
& \texttt{rgba(0,0,0,0.12)}
\\

2
& Drop-down containers and drawers
& $2$\,px
& $4$\,px
& \texttt{rgba(0,0,0,0.12)}
\\

3
& Dialogs, modal windows, and toasts
& $20$\,px
& $30$\,px
& \texttt{rgba(0,0,0,0.15)}
\\

\bottomrule
\end{tabular}
\end{table}

For implementation, UI components are first identified and their shadow
parameters are extracted. The observed parameters are then compared with the
semantic level associated with each component type.

\uifig{0.85}{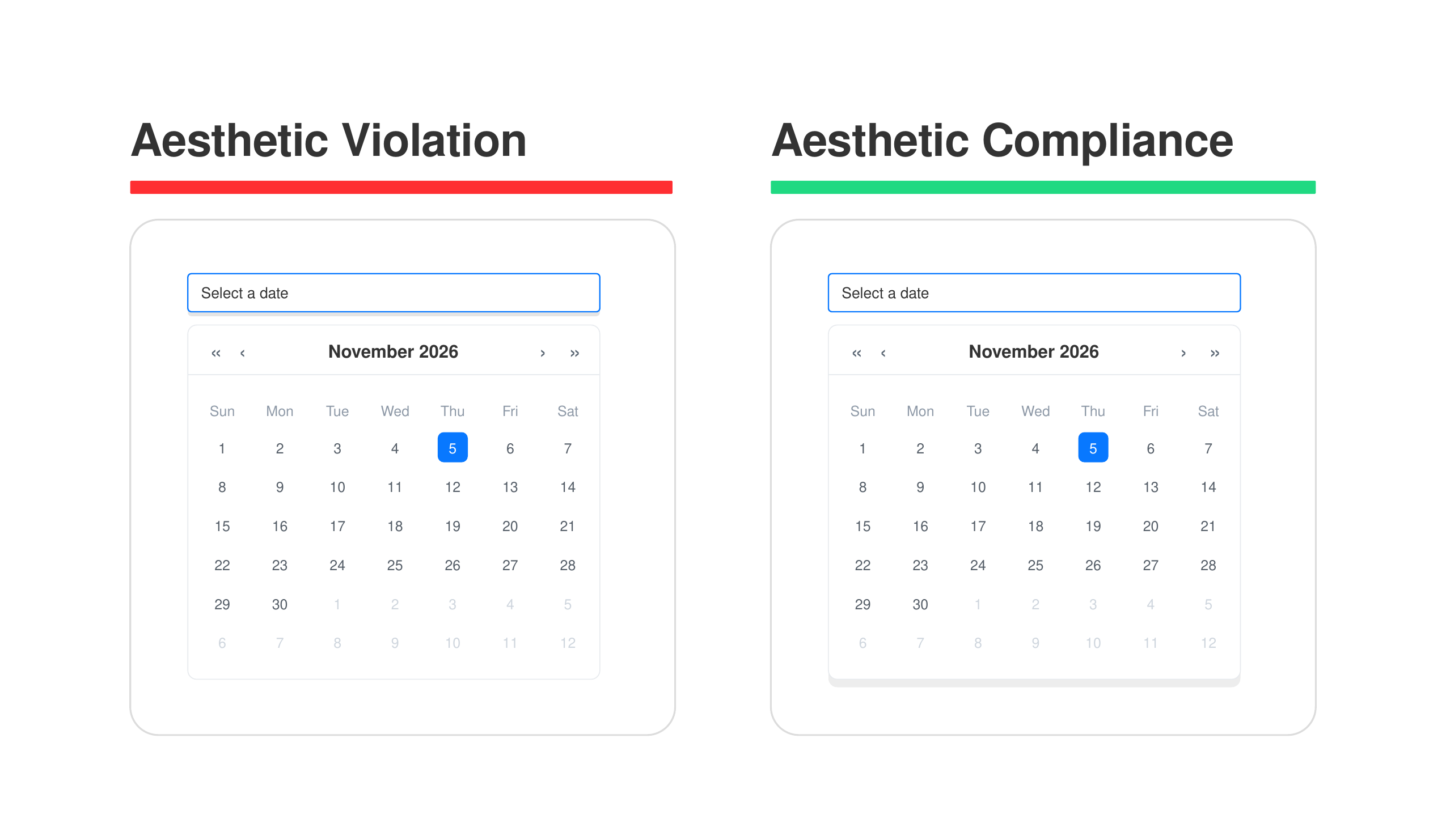}
  {\textbf{Shadow hierarchy.}
   The violating example uses a shadow treatment inconsistent with the
   component's semantic level, while the conforming example follows the
   prescribed shadow hierarchy.}
  {fig:shadow}

\subsection{Consistency}
\label{ssec:consistency}

In addition to the visual quality of individual elements, overall interface quality depends strongly on whether visual rules are applied consistently across the page~\citep{kellogg1987conceptual}. We consider consistency at two levels: container consistency
and component consistency.

\textbf{Container consistency.}
Containers carry information and separate different content regions.
Equivalent containers within the same page should use consistent corner
radii, shadows, and spacing between containers.
Figure~\ref{fig:container} illustrates how consistent corner radii and spacing
establish a regular arrangement of containers.

For implementation, container components are identified and their corner
radius, shadow, and spacing parameters are compared across equivalent
instances.

\uifig{1.00}{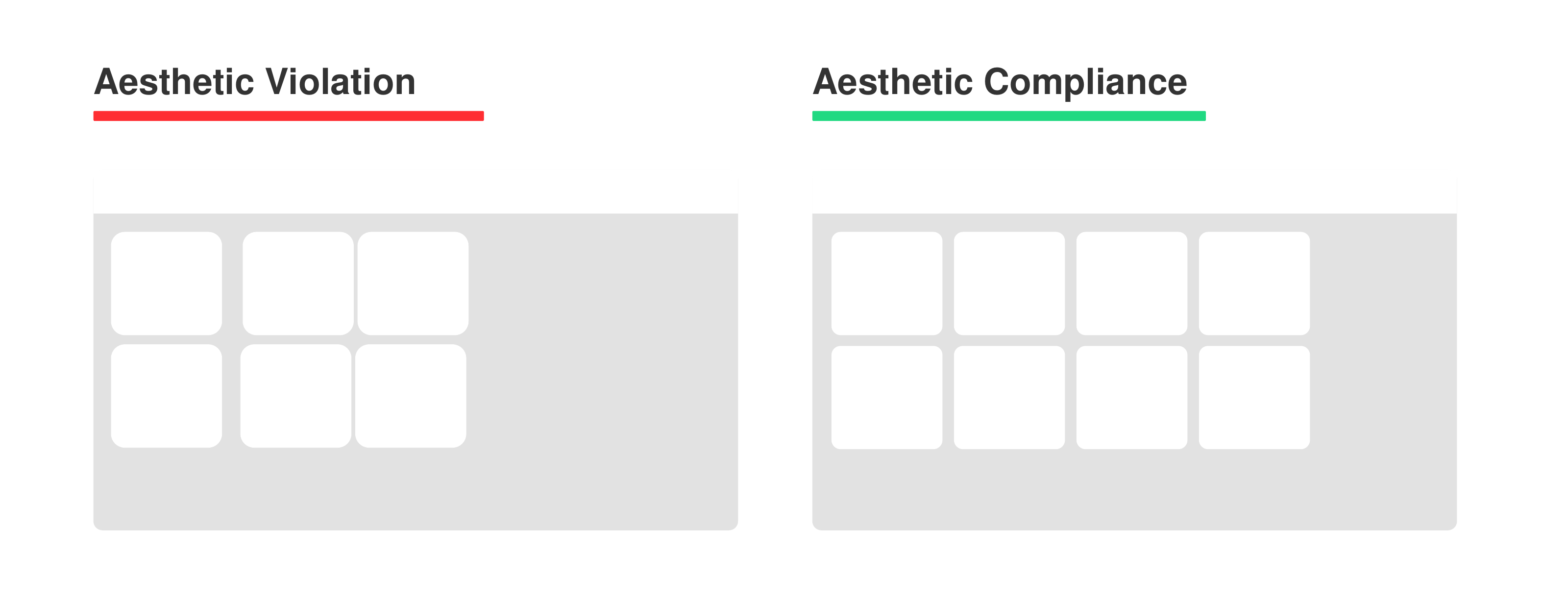}
  {\textbf{Container consistency.}
   Equivalent containers use different corner radii, shadows, or spacing in
   the violating example, while the conforming version applies a consistent
   container specification.}
  {fig:container}

\textbf{Component consistency.}
Components are among the most frequently repeated elements in an application.
We therefore require one page to use components derived from a single
specification. Instances of the same component type should maintain consistent
parameters, as illustrated by the input fields in Figure~\ref{fig:component}.

For implementation, component-recognition methods can be used to identify
component instances on the page. Parameters are then compared within each
component class to determine whether components of the same type remain
consistent.

\uifig{0.90}{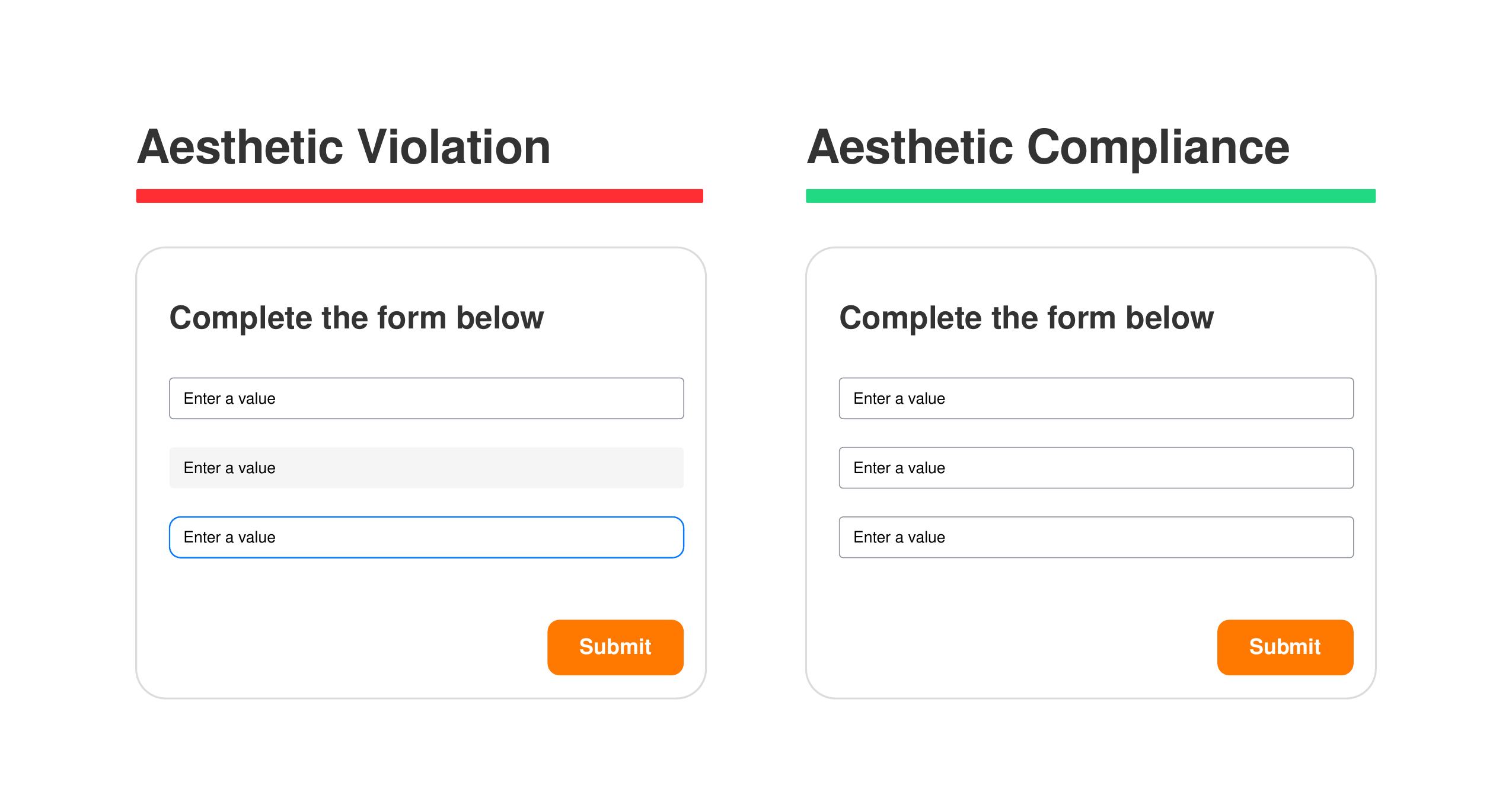}
  {\textbf{Component consistency.}
   Instances of the same component type use different specifications in the
   violating example, while the conforming version follows a shared component
   specification.}
  {fig:component}

%% file: sections/appendix_human_anno.tex
\section{Human Annotation and Evaluation Details}
\label{app:human_evaluation}

\subsection{Annotation Procedure}
\label{app:annotation_procedure}

\begin{figure}
    \centering
    \includegraphics[width=1\linewidth]{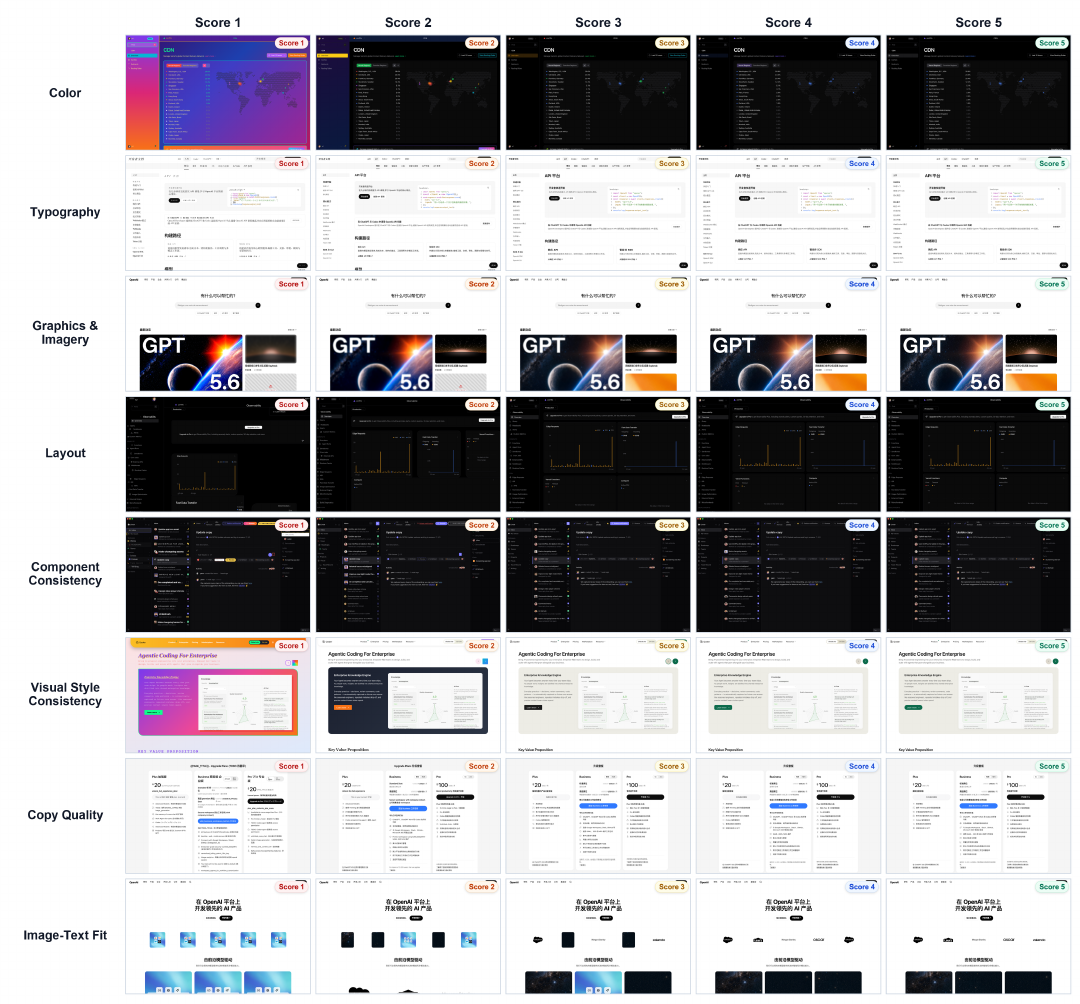}
    \caption{Visual rubric exemplars used to calibrate annotators before formal annotation. Rows correspond to the eight aesthetic dimensions, and columns correspond to rating levels from 1 to 5.}
    \label{fig:rating_calibration_examples}
\end{figure}

We initially recruited 51 professional UI designers, all with at least 3 years of industry experience in interface and product design. Among them, 45.1\% had at least 5 years of experience, including 17.8\% with at least 10 years. All had worked at leading Internet and technology companies and had substantial experience with B2B or B2C products. All annotators held at least a bachelor's degree and had received formal training in design-related disciplines, including interaction design, visual communication, and related fields. All annotators followed the same annotation guidelines throughout the annotation process. Before formal annotation, a professional UI designer prepared a set of visual rubric exemplars to operationalize the 1--5 rating scale. For each of the eight aesthetic dimensions, the exemplar set contained one representative UI for each score level, resulting in 40 examples in total. All annotators reviewed the same exemplars together with the scoring guidelines before beginning formal annotation. This calibration step was intended to align their interpretation of the five score levels and reduce differences in individual rating scales. Figure~\ref{fig:rating_calibration_examples} presents the complete exemplar set.

Each of the 1,395 reference UIs was independently rated by three designers
along eight aesthetic dimensions, yielding 33,480 dimension-level ratings
before quality control.

Given the importance of annotation quality for reliable model evaluation~\citep{pang2025vlms}, we manually inspected the rating distribution of each designer as an annotator-level quality-control step. We excluded annotators exhibiting degenerate
response patterns, defined as assigning the same extreme score (either 1 or 5)
to all ratings they submitted. Seven annotators met this criterion, and all
annotations from these annotators were removed. The final annotation set
therefore contains ratings from 44 professional designers.

Our aesthetic guidelines contain a broader set of
principle-level criteria, including rules for color, spacing, layout,
typography, imagery, shadow, and consistency. To derive the dimensions used
for human evaluation, we worked with the designers to consolidate these
fine-grained principles into perceptually coherent categories that can be
reliably judged from rendered UI screenshots. Closely related principles were
grouped into broader evaluation dimensions. For example, spacing and spatial
organization are jointly considered under layout, while component-level
regularity and visual treatments across the page are evaluated separately as
component consistency and visual style consistency. Shadow-related
principles are assessed under visual style consistency, together with
borders, corner radii, and other visual treatments. We retain color and
typography as separate dimensions and assess graphical content under
graphics and imagery. We additionally include
copy quality to assess the clarity and appropriateness of textual content,
and image--text fit to assess the semantic agreement between visual elements
and their associated text or context. This adaptation yields the shared
eight-dimensional rubric used for Aesthetic Scoring and Text-to-UI Generation
(Tasks~1 and~4).

For each UI, annotators provided ratings from 1 to 5 along eight aesthetic
dimensions: color, typography, graphics and imagery, layout, component
consistency, visual style consistency, copy quality, and Image--Text Fit.
For each evaluation, annotators were also asked to provide a brief comment
summarizing their assessment. In addition, they recorded a concise description
of the most salient design issue together with its corresponding region or
component when applicable. The resulting annotations therefore contain both
quantitative ratings and qualitative design feedback for each UI.

\textbf{Ground-truth aggregation.}
After annotator-level quality control, we compute the mean opinion score (MOS)
over all retained expert ratings for each UI and aesthetic dimension.
No additional rating-level outlier filtering is applied.
The resulting MOS is rounded to one decimal place and is used as the ground-truth score for subsequent evaluation.

\subsection{Evaluation Criteria and Ground-Truth Distribution}
\label{app:rating_criteria}

\textbf{Fine-grained criteria.}
Annotators evaluate each UI along eight fine-grained aesthetic criteria.
Each criterion is independently rated from 1 to 5 according to its
corresponding scoring rubric.
Table~\ref{tab:fine_grained_criteria} summarizes the definition of each
criterion.

\begin{table*}[t]
    \centering
    \small
    \caption{Fine-grained criteria used for human UI evaluation.}
    \label{tab:fine_grained_criteria}
    \begin{tabular}{l p{0.73\textwidth}}
        \toprule
        \textbf{Criterion} & \textbf{Definition} \\
        \midrule

        Color &
        Whether the page palette is visually harmonious, colors are used
        consistently, and sufficient contrast is maintained between text and
        background. \\

        Typography &
        Whether font choices, sizes, weights, line heights, and textual
        hierarchy are clear, consistent, and readable. \\

        Graphics \& Imagery &
        Whether images, icons, illustrations, and other visual assets are
        clear, intact, appropriately proportioned, and visually well
        presented. \\

        Layout &
        Whether page structure, information hierarchy, alignment, spacing,
        grouping, and space utilization are visually appropriate. \\

        Component Consistency &
        Whether components with equivalent functions or hierarchy follow
        consistent rules in their size, structure, appearance, and state
        representation. \\

        Visual Style Consistency &
        Whether corner radii, shadows, borders, line weights, icons, and
        other visual treatments form a coherent visual language across the
        page. \\

        Copy Quality &
        Whether headings, body text, buttons, hints, and other textual
        content are clear, natural, concise, and appropriate for the page
        context. \\

        Image--Text Fit &
        Whether images, icons, and other visual elements semantically match
        their associated text, function, and surrounding context. \\

        \bottomrule
    \end{tabular}
\end{table*}

\begin{figure}[t]
    \centering
    \includegraphics[
        width=1\textwidth
    ]{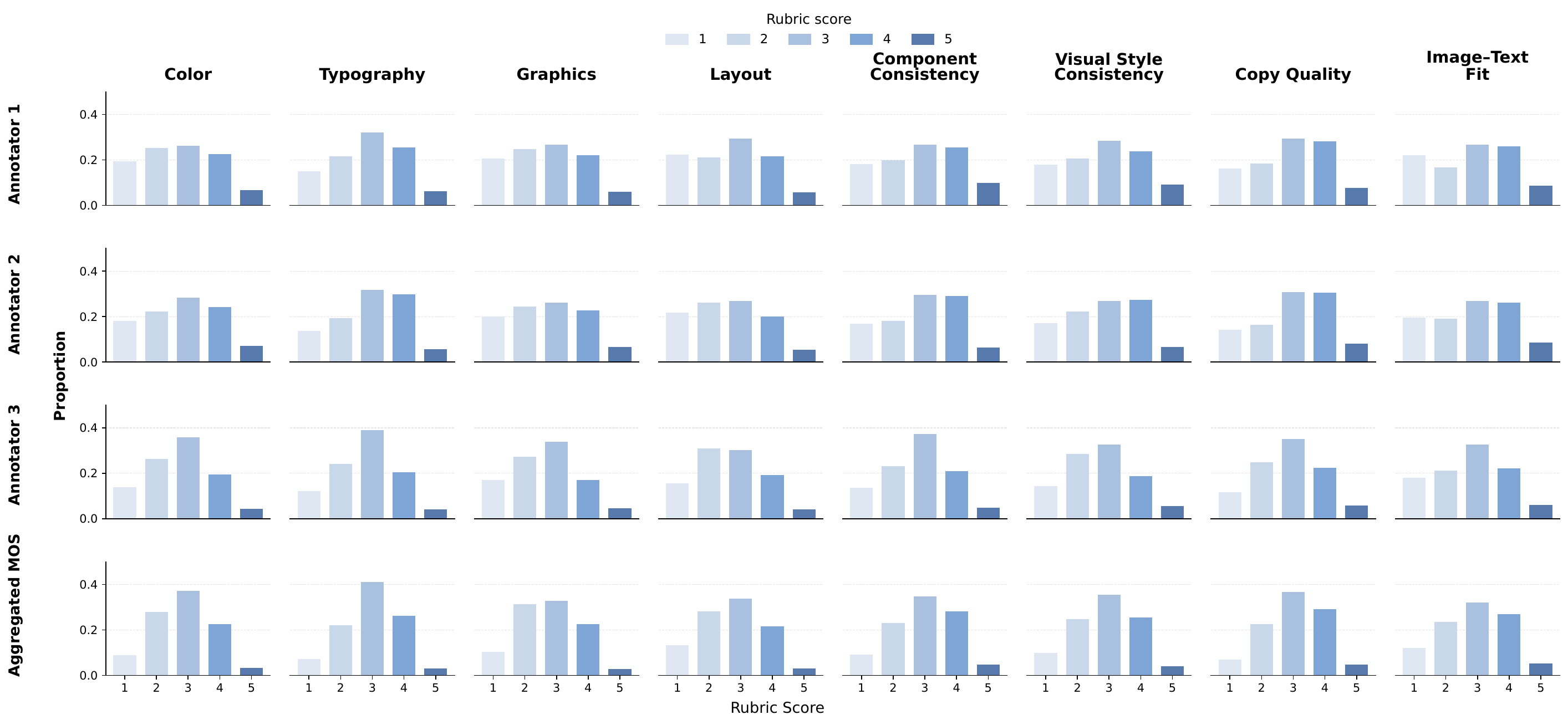}
    \caption{
Score distributions from three individual annotators and the aggregated human MOS across eight aesthetic dimensions.
}
    \label{fig:aesthetic_score_distribution}
\end{figure}

\textbf{Ground-truth distribution.}
The benchmark ground truth consists of eight dimension-level MOS
scores for each UI.
Figure~\ref{fig:aesthetic_score_distribution} compares the score
distributions of the three individual annotators with the final aggregated
MOS across all eight aesthetic dimensions.
For visualization only, the MOS values are mapped to the nearest
1--5 rubric level, while all quantitative evaluations use the one-decimal MOS scores.

\subsection{Inter-Annotator Reliability}
\label{sec:inter_annotator_reliability}

We further assess the reliability of the professional aesthetic annotations. Given the ordinal 1--5 rating scale and the rotating-rater setting, we use
ordinal Krippendorff's $\alpha$ to quantify inter-annotator agreement for each
of the eight aesthetic dimensions. As shown in
Table~\ref{tab:annotation_reliability}, $\alpha$ ranges from 0.328 to 0.431,
with a macro-average of 0.368. Layout shows the highest agreement
($\alpha=0.431$), while Color shows the lowest ($\alpha=0.328$).
Visual-style Consistency, Copy Quality, and Image--Text Fit obtain
$\alpha$ values of 0.366, 0.334, and 0.402, respectively, and do not exhibit
substantially larger disagreement than the other dimensions. Overall, the
results indicate that professional designers share some common judgment on
fine-grained UI aesthetics, while noticeable individual variation remains.

Since the benchmark uses the mean of the retained expert ratings as its
ground-truth score, we additionally examine the reliability of rating
aggregation. On the UIs retaining all three valid ratings, we use a one-way
random-effects intraclass correlation coefficient (ICC), which is appropriate
for the rotating-rater setting. The single-rating reliability, ICC(1,1),
ranges from 0.329 to 0.431 across dimensions, whereas the reliability of the
average of three independent ratings, ICC(1,3), ranges from 0.596 to 0.694,
with a macro-average of 0.637. These results show that aggregating multiple
professional ratings improves score reliability relative to relying on a
single designer, while fine-grained aesthetic assessment still retains
non-negligible inter-expert variation.

\begin{table}[t]
    \centering
    \small
    \caption{
    \textbf{Inter-annotator reliability across aesthetic dimensions.}
    We report ordinal Krippendorff's $\alpha$ for inter-annotator agreement
    and ICC(1,3) for the reliability of the average of three ratings.
    }
    \label{tab:annotation_reliability}
    \begin{tabular}{lcc}
        \toprule
        \textbf{Dimension} &
        \textbf{Krippendorff's $\alpha$} &
        \textbf{ICC(1,3)} \\
        \midrule
        Color                       & 0.328 & 0.596 \\
        Typography                  & 0.334 & 0.609 \\
        Graphics \& Imagery         & 0.346 & 0.614 \\
        Layout                      & 0.431 & 0.694 \\
        Component Consistency       & 0.399 & 0.672 \\
        Visual-style Consistency    & 0.366 & 0.635 \\
        Copy Quality                & 0.334 & 0.610 \\
        Image--Text Fit             & 0.402 & 0.667 \\
        \midrule
        \textbf{Macro Avg.}         & \textbf{0.368} & \textbf{0.637} \\
        \bottomrule
    \end{tabular}
\end{table}

\subsection{Model Evaluation Prompt}
\label{app:evaluation_prompt}

For model evaluation, we convert the human annotation criteria into a unified
evaluation prompt.
All evaluated models receive the same task instruction, scoring criteria,
and output format.
The complete prompt used for aesthetic scoring is shown in
Figure~\ref{fig:judgment_prompt_1},
Figure~\ref{fig:judgment_prompt_2}, and
Figure~\ref{fig:judgment_prompt_3}.

\begin{figure*}[t!]
\begin{tcolorbox}[promptbox={Aesthetic Scoring Prompt, Part I}]
\footnotesize
\setlist[itemize]{leftmargin=*, nosep, topsep=2pt}

\textbf{Task}

You are a professional UI design reviewer. Rate the given UI page screenshot
on each of the following UI aesthetic dimensions according to the
corresponding 1--5 scoring rubric.

\vspace{1mm}
\textbf{1. Color}

Evaluate whether the palette harmonizes with the page as a whole, colors are
used consistently, and text background contrast is clear.

\begin{itemize}
    \item \textbf{5:}
    Mature and harmonious palette with clear roles for primary, secondary,
    background, and accent colors. Color usage follows a consistent logic.
    Text background contrast is clear, with no noticeable color clashes or
    excessive saturation.

    \item \textbf{4:}
    Overall harmonious palette with a clear color system. Only a few minor
    issues are present, such as an occasionally distracting color, slightly
    weak local contrast, or slight overuse of accent colors. Overall
    appearance and readability remain strong.

    \item \textbf{3:}
    Generally acceptable, but several noticeable problems are present, such
    as locally uncoordinated colors, unclear color hierarchy, or insufficient
    text background contrast. The page remains readable and usable.

    \item \textbf{2:}
    Serious palette problems occur across multiple regions, such as
    conflicting colors, inconsistent color usage, excessive saturation, or
    poor contrast that noticeably affects readability.

    \item \textbf{1:}
    Large-scale color conflicts or abnormal contrast make substantial
    portions of the content difficult to recognize and substantially impair
    the page's presentation and comprehension.
\end{itemize}

\vspace{1mm}
\textbf{2. Typography}

Evaluate whether fonts, sizes, weights, line heights, and text hierarchy are
clear, reasonable, and readable.

\begin{itemize}
    \item \textbf{5:}
    Complete and consistent type system with clear hierarchy among headings,
    body text, captions, and other textual elements. Font sizes, weights, and
    line heights are well coordinated and comfortable to read.

    \item \textbf{4:}
    Overall clear type system with only a few minor issues in font size,
    weight, or line height. Reading and information hierarchy remain clear.

    \item \textbf{3:}
    The page remains readable, but several noticeable typography issues are
    present, such as insufficient distinction between headings and body text,
    unstable size hierarchy, inconsistent weight usage, or locally cramped
    or loose text.

    \item \textbf{2:}
    The typography system is noticeably disorganized. Multiple textual levels
    are difficult to distinguish, and inappropriate sizes, weights, or line
    heights significantly affect readability.

    \item \textbf{1:}
    Large amounts of text are difficult to read because of severely
    inappropriate font sizes, styles, or arrangements, and the textual
    hierarchy is largely ineffective.
\end{itemize}

\vspace{1mm}
\textbf{3. Graphics \& Imagery}

Evaluate whether images, icons, illustrations, and other visual assets are
clear, intact, appropriately proportioned, and visually well presented.

\begin{itemize}
    \item \textbf{5:}
    Images are sharp, intact, correctly proportioned, and naturally cropped.
    Icons and other graphics are refined and recognizable, with no noticeable
    stretching, blurring, damage, or missing assets.

    \item \textbf{4:}
    Overall asset quality is good, with only a few minor issues such as
    slightly imperfect cropping or less refined icons. The visual experience
    remains strong.

    \item \textbf{3:}
    Assets remain usable, but several perceptible problems are present, such
    as slightly blurry images, suboptimal cropping, mediocre icon quality, or
    graphics with limited recognizability.

    \item \textbf{2:}
    Multiple important visual assets are blurry, stretched, abnormally
    cropped, low quality, or missing, noticeably degrading the page's visual
    quality.

    \item \textbf{1:}
    Large numbers of core images or graphics are broken, missing,
    unrecognizable, or severely distorted, substantially disrupting the
    presentation of the page.
\end{itemize}

\end{tcolorbox}
\caption{Model evaluation prompt for aesthetic scoring, Part I.}
\label{fig:judgment_prompt_1}
\end{figure*}

\begin{figure*}[t!]
\begin{tcolorbox}[promptbox={Aesthetic Scoring Prompt, Part II}]
\small
\setlist[itemize]{leftmargin=*, nosep, topsep=2pt}

\textbf{4. Layout}

Evaluate whether page structure, information hierarchy, alignment, spacing,
grouping, and space utilization are reasonable.

\begin{itemize}
    \item \textbf{5:}
    The page has a clear overall structure and well-defined visual priorities.
    Elements are accurately aligned, spacing is stable and regular, related
    content is properly grouped, and sections are clearly separated. No
    noticeable crowding, abnormal whitespace, misalignment, occlusion, or
    overflow is present.

    \item \textbf{4:}
    Overall layout is mature and reasonable, with only a few minor spacing,
    alignment, or proportion issues. Reading flow and information structure
    remain clear.

    \item \textbf{3:}
    The main structure is generally sound, but several noticeable problems are
    present, such as inconsistent alignment, unstable spacing, locally crowded
    or loose areas, or insufficiently clear information hierarchy.

    \item \textbf{2:}
    Serious layout problems appear across multiple areas, including obvious
    misalignment, crowding, abnormal whitespace, confused hierarchy, or
    unbalanced space allocation. Information access is noticeably affected.

    \item \textbf{1:}
    The layout is severely disrupted, with widespread overlapping,
    overflowing, misplaced, or structurally disorganized elements that make
    the page difficult to comprehend.
\end{itemize}

\vspace{1mm}
\textbf{5. Component Consistency}

Evaluate whether components with the same function or hierarchy follow
consistent design rules.

\begin{itemize}
    \item \textbf{5:}
    Buttons, cards, inputs, tags, navigation elements, and other component
    types follow clear and unified rules. Components with the same function
    use stable dimensions, structures, and state representations.

    \item \textbf{4:}
    The component system is generally consistent, with only occasional minor
    deviations in size, structure, or style.

    \item \textbf{3:}
    A basic component system is recognizable, but several noticeable
    inconsistencies are present, such as same-level buttons with different
    sizes, same-type cards with different structures, or identical states
    represented in different ways.

    \item \textbf{2:}
    Many similar components lack unified rules, and the same functions or
    hierarchy levels are frequently represented using different designs,
    resulting in weak overall system consistency.

    \item \textbf{1:}
    Similar components differ substantially in structure or appearance,
    resulting in a strongly fragmented and visibly inconsistent component
    system.
\end{itemize}

\vspace{1mm}
\textbf{6. Visual Style Consistency}

Evaluate whether the visual language, including corner radii, shadows,
borders, line weights, icon styles, and decorative treatments, is consistent.

\begin{itemize}
    \item \textbf{5:}
    The page has a stable and unified visual language. Corner radii, shadows,
    borders, line weights, icons, and decorative treatments follow clear and
    consistent rules across sections.

    \item \textbf{4:}
    The overall visual style is consistent, with only minor local deviations
    such as occasional differences in corner radius, shadow, or border style.

    \item \textbf{3:}
    A generally coherent visual direction is present, but several noticeable
    inconsistencies appear across sections in corner radii, shadows, borders,
    icons, or decorative treatments.

    \item \textbf{2:}
    Visual languages differ substantially across sections. Corner radii,
    shadows, borders, strokes, and icon styles lack consistent rules,
    producing a fragmented visual appearance.

    \item \textbf{1:}
    Sections follow substantially different visual systems, resulting in a
    severely inconsistent and disorganized overall style.
\end{itemize}

\end{tcolorbox}
\caption{Model evaluation prompt for aesthetic scoring, Part II.}
\label{fig:judgment_prompt_2}
\end{figure*}

\begin{figure*}[t!]
\begin{tcolorbox}[promptbox={Aesthetic Scoring Prompt, Part III}]
\footnotesize
\setlist[itemize]{leftmargin=*, nosep, topsep=2pt}

\textbf{7. Copy Quality}

Evaluate whether the page text is clear, natural, easy to understand, and
appropriate for the page context.

\begin{itemize}
    \item \textbf{5:}
    Headings, body text, buttons, hints, and other copy are accurate, natural,
    concise, and context appropriate. The information is clearly expressed.

    \item \textbf{4:}
    Copy is generally clear and appropriate, with only minor unnatural
    phrasing, redundancy, or imprecise wording.

    \item \textbf{3:}
    Core content remains understandable, but several noticeable problems are
    present, such as stiff phrasing, vague information, repeated content,
    mediocre organization, or occasional context inappropriate expressions.

    \item \textbf{2:}
    Multiple important pieces of text are unclear, context inappropriate,
    semantically incorrect, or incomplete, noticeably affecting comprehension.

    \item \textbf{1:}
    Large amounts of core text are garbled, meaningless, substantially
    incorrect, or incomprehensible, preventing the page from effectively
    conveying its main information.
\end{itemize}

\vspace{1mm}
\textbf{8. Image--Text Fit}

Evaluate whether images, icons, and other visual elements semantically match
their associated text, function, and surrounding context.

\begin{itemize}
    \item \textbf{5:}
    Images, icons, and text are highly compatible. Visual elements accurately
    express the corresponding content or function and effectively support
    information comprehension.

    \item \textbf{4:}
    Image text relationships are generally appropriate, with only a few
    semantically imprecise visual elements that have little impact on
    comprehension.

    \item \textbf{3:}
    Most image text relationships are reasonable, but several issues are
    present, such as weakly relevant images, icons with unclear semantics, or
    visual elements that require additional context to understand.

    \item \textbf{2:}
    Multiple important images or icons clearly mismatch their associated text
    or function, creating substantial potential for misunderstanding or
    incorrect expectations.

    \item \textbf{1:}
    Image text relationships are widely incorrect, irrelevant, or misleading,
    and key visual elements substantially interfere with correct
    interpretation.
\end{itemize}

\vspace{1mm}
\textbf{Output Requirements}

Output only the following JSON object with no additional text:

\begin{quote}
\ttfamily
\{\\
\quad "color": \{ "score": <1--5>, "reason": "<one-sentence justification>" \},\\
\quad "typography": \{ "score": <1--5>, "reason": "<one-sentence justification>" \},\\
\quad "graphics": \{ "score": <1--5>, "reason": "<one-sentence justification>" \},\\
\quad "layout": \{ "score": <1--5>, "reason": "<one-sentence justification>" \},\\
\quad "component\_consistency": \{ "score": <1--5>, "reason": "<one-sentence justification>" \},\\
\quad "visual\_style\_consistency": \{ "score": <1--5>, "reason": "<one-sentence justification>" \},\\
\quad "copy\_quality": \{ "score": <1--5>, "reason": "<one-sentence justification>" \},\\
\quad "image\_text\_fit": \{ "score": <1--5>, "reason": "<one-sentence justification>" \}\\
\}
\end{quote}

\vspace{1mm}
\textbf{Evaluation Rules}

\begin{enumerate}
    \item Score each aesthetic dimension independently according to its
    corresponding rubric.
    \item Keep each reason to a single sentence in English.
    \item Output only the specified JSON object.
\end{enumerate}

\end{tcolorbox}
\caption{Model evaluation prompt for aesthetic scoring, Part III.}
\label{fig:judgment_prompt_3}
\end{figure*}

\subsection{Annotation Examples}
\label{app:annotation_examples}

\begin{figure}[t]
    \centering
    \includegraphics[
        width=0.96\textwidth
    ]{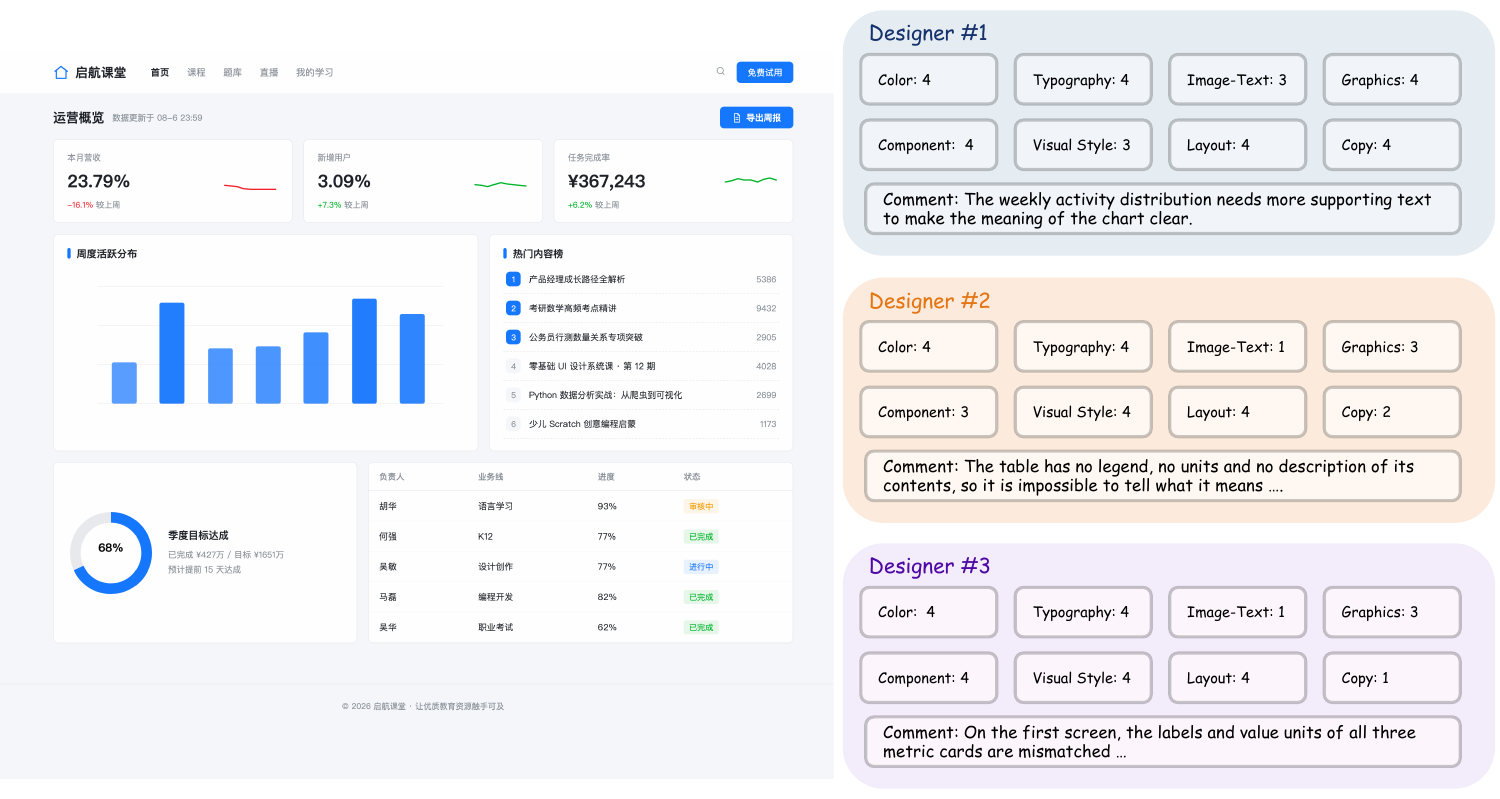}
    \caption{
    Representative human annotation example.
    }
    \label{fig:human_anno_1}
\end{figure}

\begin{figure}[t]
    \centering
    \includegraphics[
        width=0.96\textwidth
    ]{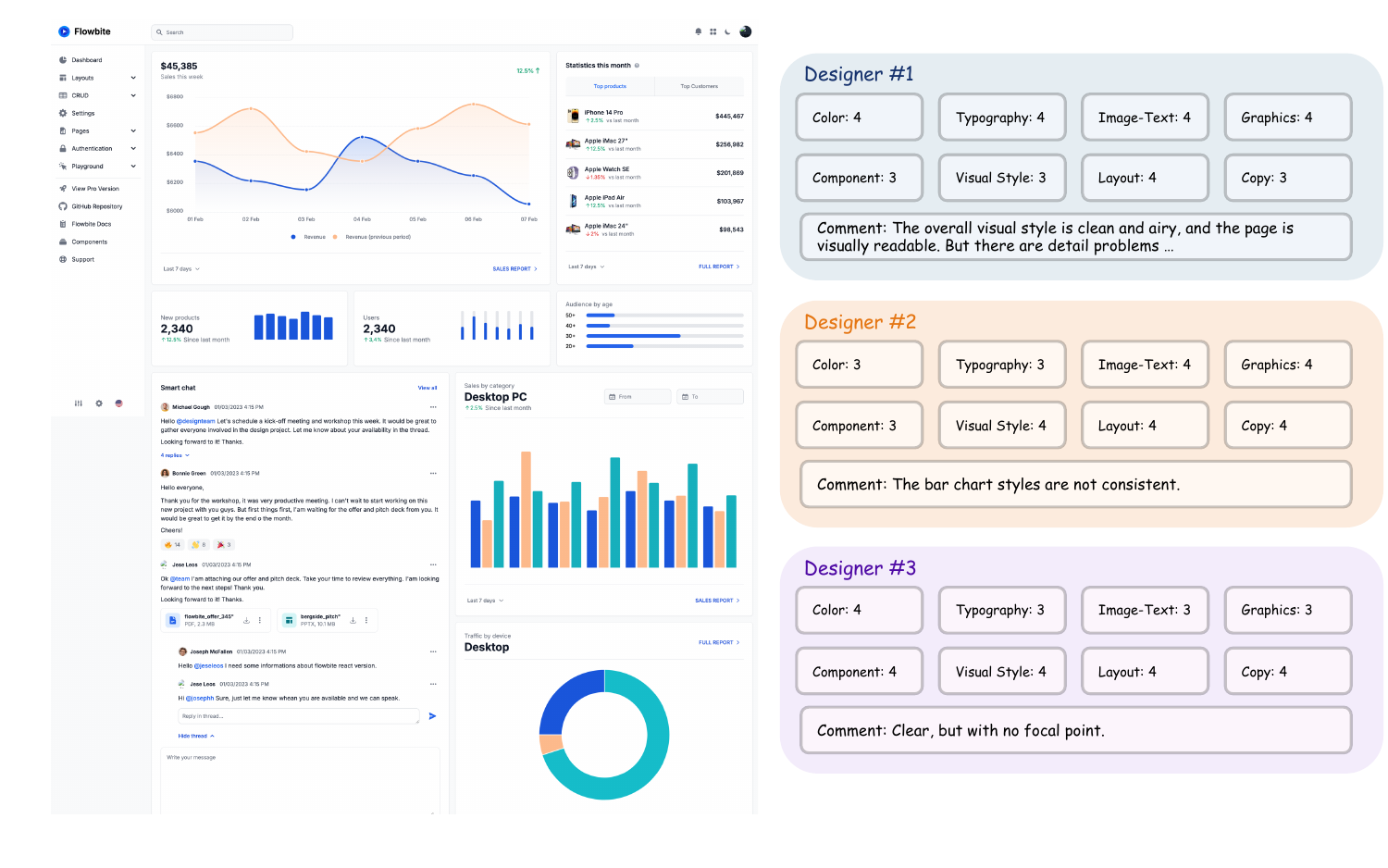}
    \caption{
    Representative human annotation example.
    }
    \label{fig:human_anno_2}
\end{figure}

\begin{figure}[t]
    \centering
    \includegraphics[
        width=0.96\textwidth
    ]{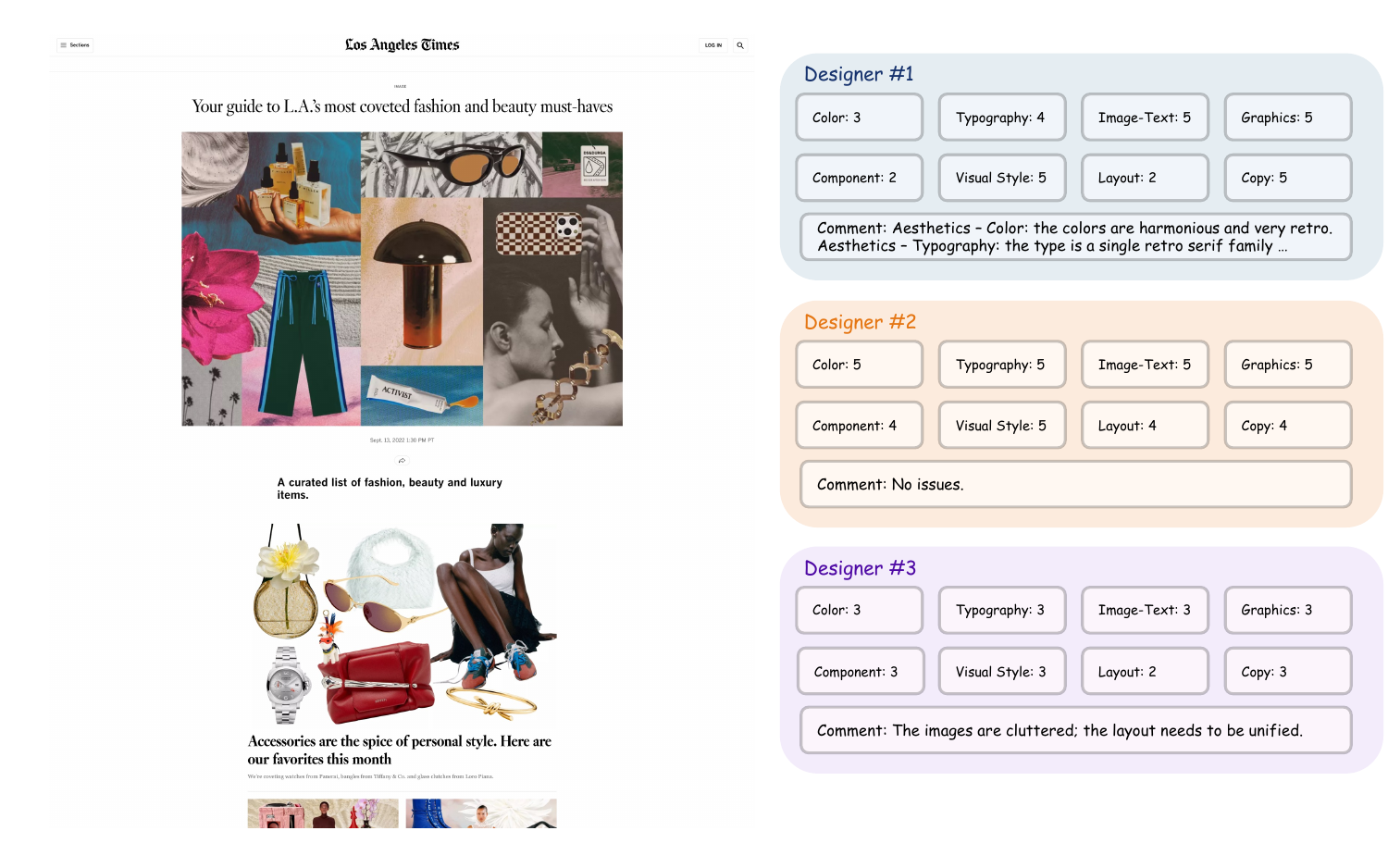}
    \caption{
    Representative human annotation example.
    }
    \label{fig:human_anno_3}
\end{figure}

We provide three representative annotation examples in
Figures~\ref{fig:human_anno_1}, \ref{fig:human_anno_2} and \ref{fig:human_anno_3} to illustrate the
human evaluation and ground-truth aggregation process.
Each example presents the original UI screenshot, the independent ratings
and qualitative comments from three professional designers.

%% file: sections/appendix_benchmark_details.tex
\section{More Details of \benchmark}
\label{sec:bench_details}
\subsection{Reference UI Collection and Quality Assurance}
\label{app:reference_ui}

\textbf{Reference UI Collection Details.}
We apply quality filtering, rendering validation, and duplicate removal to
obtain the final reference UI collection. The adapted subset comprises 300
pages from WebCode2M~\citep{gui2025webcode2m}\footnote{xcodemind/webcode2m\_purified}
and 77 pages from Design2Code~\citep{si2025design2code}\footnote{SALT-NLP/Design2Code-HARD}.
The collection contains 1,292 desktop pages, 92 mobile pages, and 11 tablet pages.

\textbf{Source Adaptation and Asset Recovery.}
The reference collection combines real-world webpages, adapted public UI
datasets, and model-generated pages. Each page is packaged with its source
code, localized assets, and a full-page screenshot. For adapted pages with
missing or placeholder images, we recover semantically appropriate visual
assets using the surrounding context and alternative text. This process
replaces 2,046 visual elements across 310 pages.

\textbf{Rendering Fidelity.}
Real-world webpages often depend on remote images, fonts, stylesheets, and
other resources whose availability can change over time. We localize these
dependencies to obtain self-contained pages that can be executed offline.
For pages with original reference screenshots, we re-render each localized
page under the corresponding viewport configuration and compare the result
with the original screenshot.

We measure rendering fidelity using pixel-level mean absolute error (MAE)
and the structural similarity index (SSIM)~\citep{wang2004image}. MAE measures
pixel discrepancies, while SSIM measures structural visual similarity. We
require $\mathrm{MAE}<10$ and rank surplus candidates by SSIM; retained pages
in this subset have a median SSIM of $0.994$. Pages with substantial visual
discrepancies are filtered or manually inspected for missing resources,
layout drift, and abnormal rendering. Before finalizing the reference pool,
professional UI designers inspect rendering quality, asset completeness,
and layout integrity. These checks establish a stable reference for
controlled degradation, reducing incidental rendering variation that could
otherwise confound attribution of the injected defects.

\subsection{Controlled Aesthetic Violations and Task Construction}
\label{app:controlled_violations}

\textbf{Controlled Violation Taxonomy.}
Tasks 2 and 3 use controlled aesthetic violation rules that are localizable,
editable in source code, and verifiable after rendering. We identify
principles from the professional UI aesthetic guidelines that can be mapped
to local visual elements or page structures and operationalize them as
executable degradation rules. Rules whose effects are visually insignificant
or difficult to attribute reliably are excluded. We run rule detectors over
all 1,395 reference UIs to assess applicability, identify candidate target
elements, and detect naturally occurring violations. A violation is
instantiated only when the required structural or semantic conditions are
present and the target region satisfies the corresponding principle. Each
degraded instance therefore has an explicit target, an associated design
principle, and a traceable visual modification.

In total, we define 13 controlled violation rules spanning five aesthetic dimensions: Typography, Layout and Reading Flow, Spacing, Visual Style Consistency, and Color. Table~\ref{tab:controlled_violation_rules} summarizes the design constraint represented by each rule, the primary property being manipulated, and the resulting aesthetic violation.
\begin{table*}[t]
    \caption{
The 13 controlled aesthetic violation rules used in our benchmark.
A rule is instantiated only when the reference UI satisfies its required
structural or semantic context.
T, L, S, V, and C denote Typography, Layout \& Reading Flow, Spacing,
Visual Style Consistency, and Color, respectively.
    }
    \label{tab:controlled_violation_rules}
    \centering
    \small

    \renewcommand{\arraystretch}{1.18}
    \setlength{\tabcolsep}{7pt}

    \begin{tabularx}{\textwidth}{
        >{\centering\arraybackslash}p{1.45cm}
        >{\raggedright\arraybackslash}p{4.25cm}
        >{\raggedright\arraybackslash}X
    }

        \arrayrulecolor{black}
        \toprule

        \rowcolor{TableHeader}
        \textbf{Dimension} &
        \textbf{Violation Rule} &
        \textbf{Description} \\

        \midrule


        \rowcolor{white}
        \dimT &
        \textbf{Peer Font Weight} &
        Changes the weight of a peer text element, breaking consistency among texts at the same visual level. \\
        \arrayrulecolor{TableRule}\specialrule{0.35pt}{0pt}{0pt}

        \rowcolor{TableAltRow}
        \dimT &
        \textbf{Peer Font Size} &
        Changes the size of a peer text element, disrupting the local typographic hierarchy. \\
        \specialrule{0.35pt}{0pt}{0pt}

        \rowcolor{white}
        \dimT &
        \textbf{Font Family Count} &
        Introduces an additional font family, increasing unnecessary variation in the page typography. \\
        \specialrule{0.35pt}{0pt}{0pt}

        \rowcolor{TableAltRow}
        \dimT &
        \textbf{Text Scale Count} &
        Introduces an additional font-size level, increasing complexity in the existing typographic scale. \\
        \specialrule{0.35pt}{0pt}{0pt}

        \rowcolor{white}
        \dimT &
        \textbf{Table Text Alignment} &
        Changes text or numeric alignment within a table, breaking its alignment consistency. \\
        \specialrule{0.35pt}{0pt}{0pt}


        \rowcolor{TableAltRow}
        \dimL &
        \textbf{Vertical Alignment} &
        Offsets a peer element from its original vertical alignment. \\
        \specialrule{0.35pt}{0pt}{0pt}

        \rowcolor{white}
        \dimL &
        \textbf{Gutenberg Information Region} &
        Alters the spatial or sequential organization of information elements under the Gutenberg reading pattern. \\
        \specialrule{0.35pt}{0pt}{0pt}

        \rowcolor{TableAltRow}
        \dimL &
        \textbf{Fitts Edge Zone} &
        Perturbs elements associated with page-edge regions, disrupting their original edge-layout relationship. \\
        \specialrule{0.35pt}{0pt}{0pt}


        \rowcolor{white}
        \dimS &
        \textbf{Peer Spacing Uniformity} &
        Perturbs peer elements to break their originally uniform spacing. \\
        \specialrule{0.35pt}{0pt}{0pt}

        \rowcolor{TableAltRow}
        \dimS &
        \textbf{Container Row Spacing} &
        Changes spacing between rows or elements within comparable containers. \\
        \specialrule{0.35pt}{0pt}{0pt}


        \rowcolor{white}
        \dimV &
        \textbf{Container Radius Consistency} &
        Changes the radius of a peer container, making it inconsistent with comparable components. \\
        \specialrule{0.35pt}{0pt}{0pt}


        \rowcolor{TableAltRow}
        \dimC &
        \textbf{Peer Accent Color} &
        Changes the accent color of a peer component, breaking color consistency among comparable components. \\
        \specialrule{0.35pt}{0pt}{0pt}

        \rowcolor{white}
        \dimC &
        \textbf{Semantic Color Stability} &
        Perturbs a color with a stable semantic role, deviating from the page's semantic color system. \\

        \arrayrulecolor{black}
        \bottomrule

    \end{tabularx}


\end{table*}

\textbf{Rule Applicability and Coverage.}
Different aesthetic rules require different page structures, semantic roles, and local design contexts, and are therefore not applicable to every reference UI. For each rule, we first perform an applicability check on the original page. A controlled violation is constructed only when the required candidate structure is present and the corresponding region of the reference UI already satisfies the target design principle.

For example, Peer Font Size and Peer Font Weight require comparable peer text elements; Container Radius Consistency requires a group of visually consistent peer containers; Table Text Alignment applies only to pages containing an appropriate tabular structure; and Semantic Color Stability and Fitts Edge Zone require more specific semantic or layout contexts. If the required element is absent, or if the reference UI already violates the target principle, the corresponding rule is not instantiated on that page.

After passing the applicability check, a degradation must additionally be locally executable, render reliably, and have an explicitly identifiable affected region. Across the candidate reference UIs, the 13 rules identify 77,277 valid candidate elements. Some rules are applicable to a broad range of pages, whereas others naturally arise only under more specific structural or semantic conditions. Consequently, the rule and dimension distributions in the formal evaluation cohort are determined by the actual applicability of each rule to the reference UIs, while preserving a well-defined design context and a verifiable clean-to-degraded correspondence for every controlled defect.

\textbf{Degradation Execution and Validation.}
Each degradation modifies a small subset of eligible elements, typically
around $20\%$ of the target group, to keep the intervention localized and
attributable. For example, Peer Font Size perturbs selected text elements at
the same hierarchy level, while Peer Accent Color changes the accent color
of a subset of comparable components. Professional UI designers iteratively
review the rule definitions and perturbation parameters. We construct both
single-violation and compositional cases, with mild and severe degradation
levels.

Each modified page is re-rendered and automatically validated before
inclusion. A candidate is retained only if the intended violation is
introduced with a perceptible magnitude, remains localized without
unintended layout changes or unrelated aesthetic violations, and yields a
detected violation set matching the injected recipe. Candidates that fail
these checks are discarded. Each retained instance contains the reference
UI, the degraded UI, and precise violation annotations identifying the
affected dimensions and regions. These instances and annotations are shared
by Aesthetic Diagnosis and Aesthetic Repair.

\textbf{Diagnosis and Repair Cohorts.}
Based on these controlled violation rules, we construct the formal Aesthetic Diagnosis and Aesthetic Repair tasks. To enable a controlled comparison between diagnosis and repair, Tasks 2 and 3 share exactly the same 660 underlying reference UI instances and the same 1,209 controlled aesthetic defects.

For Task 2, each degraded UI is paired with its corresponding clean control, resulting in 660 UI pairs for evaluating whether a model can identify and localize the injected aesthetic issues from visual inputs. Task 3 uses the same 660 degraded UIs and defect annotations to construct 660 repair instances, evaluating whether a model can correct the corresponding issues by modifying the executable page. Sharing the same underlying pages and defect ground truth allows the two capabilities to be evaluated under closely matched data conditions.

\begin{figure*}[t]
    \centering
    \includegraphics[width=\textwidth]{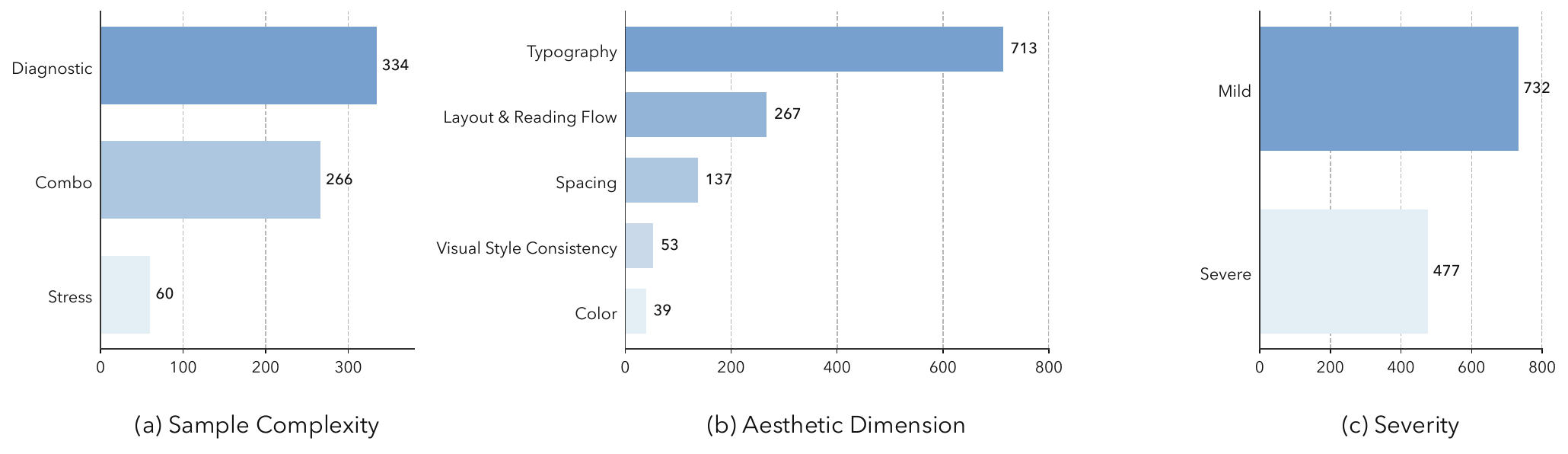}
    \caption{
    Distribution of the formal evaluation cohort shared by Tasks 2 and 3.
    (a) Complexity distribution of the 660 underlying UI instances.
    (b) Distribution of the 1,209 controlled aesthetic defects across five aesthetic dimensions.
    (c) Distribution of mild and severe defects.
    }
    \label{fig:controlled_defect_distribution}
\end{figure*}

\textbf{Defect Complexity Splits.}
We further organize the shared Diagnosis--Repair cohort into three regimes according to violation multiplicity, each targeting a distinct diagnostic challenge.
The \textit{diagnostic} split contains exactly one injected violation, enabling rule-level attribution of detection failures to a specific aesthetic rule.
The \textit{combo} split contains 2--3 co-occurring violations and tests search completeness, i.e., whether a model continues to identify the remaining defects after detecting one issue.
The \textit{stress} split contains 4--6 simultaneous violations and probes recall under defect-dense interfaces.
Together, these splits progress from isolated rule recognition to compositional diagnosis and high-density defect retrieval. As shown in Figure~\ref{fig:controlled_defect_distribution}, the 660 instances are further organized into three levels according to defect composition. The \textit{diagnostic} split contains 334 instances with a single controlled defect, providing a basic setting for isolated aesthetic issues. The \textit{combo} split contains 266 instances with multiple simultaneous defects, while the remaining 60 \textit{stress} instances contain more complex defect combinations.

The 1,209 defects span all five aesthetic dimensions: 713 in Typography, 267 in Layout and Reading Flow, 137 in Spacing, 53 in Visual Style Consistency, and 39 in Color. In terms of degradation strength, 732 defects are categorized as mild and 477 as severe. The number of degraded instances is naturally imbalanced across violation rules because not every degradation can be validly applied to every real-world UI. We apply a degradation only when the original UI satisfies the rule-specific applicability conditions, rather than forcing each rule onto a predefined number of pages. Consequently, rules that are applicable to more reference UIs yield more degraded instances, while those with more restrictive applicability conditions yield fewer.

The shared cohort establishes a direct correspondence between Tasks 2 and 3: the former evaluates whether a model can detect and localize a given aesthetic issue, whereas the latter evaluates whether it can take effective corrective action on the same issue. Because evaluating capabilities in isolation can overlook their interactions~\citep{wu2026lifeside}, we hold the underlying UI instances, defect definitions, and defect distribution fixed to analyze the alignment between principle-level judgment and UI repair.
For the cross-task Judgment--Action Association (JAA) analysis, correct
principle-level judgment requires both correct degradation detection and exact attribution of the
affected aesthetic dimensions;
localization accuracy is evaluated separately and is not included in $J_i$.

\textbf{Perceptual Comparison between Controlled Degradations and Real-World Defects.}
To examine whether our controlled degradations exhibit visual patterns similar to naturally occurring UI defects, we conduct a small-scale blind comparison study. For each of the 13 controlled violation rules, we collect five naturally occurring defect cases with clear visual evidence from public webpages, product bug reports, and official community forums, yielding 65 real-world defects in total. Each real-world case is paired with a controlled degradation from our benchmark corresponding to the same violation rule, resulting in 65 real--controlled pairs. Three non-expert participants independently evaluate all pairs. For each pair, the two UIs are presented in randomized left--right order, and participants are unaware of their sources and the degradation procedure. They are asked only to identify which UI issue was artificially introduced.

If the controlled degradations contained obvious synthetic artifacts, participants should be able to identify them substantially above the 50\% chance level. The average source-identification accuracy is 56.4\%, only modestly above chance. This result suggests that our controlled degradations do not exhibit readily identifiable synthetic artifacts and are perceptually similar to naturally occurring UI defects of the same type.

\subsection{Requirement Reconstruction}
\label{app:requirement_reconstruction}

For Text-to-UI Generation, we use GPT-5.4 to reconstruct an English
natural-language requirement from each reference UI. The requirement takes
the form of a product requirements document describing the page's purpose,
information structure, major functional modules, and overall visual
direction. Exact colors, font sizes, coordinates, and spacing values are
omitted, leaving these visual implementation choices to the evaluated model.

Each requirement is accompanied by objectively verifiable checklist items
and reviewed by professional UI designers. The requirement and checklist
specify the functional intent and content to be preserved without prescribing
the precise visual parameters of the reference page. This provides a common
specification for comparing generated UIs while allowing different visual
implementations.

At evaluation time, the model receives the reconstructed requirement,
original textual content, and image assets, without access to the reference
screenshot. Although the requirement is in English, visible text retains
the language of the original page. The model determines the layout, visual
hierarchy, typography, spacing, color, and component organization.
Asset selection and presentation are described in
Appendix~\ref{app:input_preprocessing}.

%% file: sections/appendix_experiment.tex
\section{Experimental Details}
\label{app:experimental_details}

\subsection{Model and Inference Settings}
\label{app:model_inference_settings}

We evaluate a diverse set of frontier proprietary models and open-weight multimodal models. Within each task, all models use the same task instructions, input format, and evaluation protocol.
All experiments use English prompts with temperature set to 0.
The maximum output length is set to 2,000 tokens for Tasks 1 and 2 and
8,000 tokens for Task 3.
Since Task 4 requires models to generate complete executable HTML pages,
we do not impose an explicit maximum output length for this task.
For API failures or invalid outputs, each sample is retried up to three times.

All open-weight models are evaluated on a single NVIDIA A800 GPU with 80\,GB of memory,
using a batch size of 8.
Except for hardware-specific configurations, the open-weight and proprietary models use
the same prompts, input preprocessing, and task settings.

\subsection{Input Preprocessing and Asset Organization}
\label{app:input_preprocessing}

\textbf{Visual Inputs for Tasks 1--3.}
Tasks 1--3 use full-page webpage screenshots as visual inputs.
Since some webpages are substantially taller than a typical single-image input,
we adopt a unified long-page processing strategy that preserves both global page structure
and local visual details.
The webpages are rendered with a viewport width of 1440 pixels.
Let $h$ denote the height of a full-page screenshot.
When $h \leq 1800$ pixels, the full screenshot is directly provided to the model without slicing.
When $h > 1800$ pixels, the screenshot is divided vertically into slices of
1800 pixels in height, with an overlap of 180 pixels between adjacent slices,
corresponding to a stride of 1620 pixels.
The slice starting positions are therefore
$0, 1620, 3240, \ldots$.
If the final slice produced by this fixed stride does not align with the bottom of the page,
we additionally include a bottom-aligned slice starting at $h-1800$,
ensuring that the complete page is covered.
As a result, the overlap between the final two slices may be larger than 180 pixels.

For webpages requiring multiple slices, we additionally provide a full-page overview image,
obtained by proportionally resizing the original screenshot to a height of 1500 pixels.
The overview image preserves the overall layout, information structure, and relative positions
of different page regions, while the high-resolution slices retain local details such as
typography, spacing, alignment, and component styling.
The model first receives the full-page overview, followed by all slices in top-to-bottom order.
Each slice is explicitly labeled with its position in the sequence.

The same visual preprocessing protocol is used across Tasks 1--3.
For Task 2, the clean and degraded versions are processed independently using the same
slicing strategy, and their presentation order is randomized before being given to the model.
Task 3 uses the same visual representation of the degraded UI as Task 2,
together with the corresponding HTML source code and the structured diagnosis produced
by the same model in Task 2.

\textbf{Asset Inputs for Task 4.}
Task 4 provides each model with an English webpage requirement, the visible textual content
from the original page, and a set of visual assets extracted from the corresponding Reference UI.
Since a webpage may contain a large number of duplicated, decorative, or low-information images,
we apply a fixed asset filtering procedure before generation.
The procedure includes geometric filtering, duplicate removal,
near-duplicate clustering, filtering of primarily decorative assets,
and relevance-based ranking of the remaining candidates.
At most 32 visual assets are retained for each webpage.

Each retained asset is assigned a deterministic identifier,
\texttt{V01}, \texttt{V02}, $\ldots$, \texttt{V32},
and is made available through a corresponding local relative path.
To provide direct visual access to the selected assets, we organize them into at most two
$4\times4$ contact sheets, each containing up to 16 assets.
Each entry is labeled with its asset identifier, filename, and original image dimensions.

We additionally provide a compact textual description for each selected asset,
including its identifier, relative path, width, height, and a short semantic caption.
The visual contact sheets allow the model to inspect the appearance of the candidate assets,
while the textual descriptions provide explicit correspondence between their semantic content
and local file paths.
The generated HTML can therefore reference the selected assets using the provided relative paths.

All webpage requirements in Task 4 are written in English,
while the visible text extracted from the original webpage is preserved in its source language.
Therefore, although the generation instructions are consistently given in English,
the target webpage content may span different languages.
Models do not have access to the Reference UI screenshot during generation;
the layout, visual hierarchy, typography, spacing, color, and component organization
must be determined from the provided requirement, textual content, and visual assets.

For webpages with no usable visual assets after filtering,
the model is provided with an empty asset list and is not allowed to introduce external images
or online placeholder services.
For webpages with available assets, the generated page may only reference the local visual
resources provided for that sample.

%% file: sections/appendix_more_expe.tex
\section{More Experimental Analysis}
\label{sec:more_exp}
\subsection{Validation and Calibration of GPT-5.4 as an Automatic Aesthetic Judge}
\label{sec:aesthetic_judge_calibration}

To provide an automatic aesthetic assessment for Task~4, we employ
GPT-5.4-2026-03-05 as the aesthetic judge.
This model is not included among the evaluated models in the main experiments.
Before applying it to generated UIs, we first examine its agreement with
professional human judgments using the designer-annotated data from Task~1.

Specifically, we follow the same eight-dimensional aesthetic rubric, image-input
protocol, and inference setting as in Task~1.
GPT-5.4 assigns a score from 1 to 5 for each of the eight aesthetic dimensions:
color, typography, graphics and imagery, layout, component consistency,
visual-style consistency, copy quality, and image--text fit.
We compare these ratings against the cleaned professional-designer Mean Opinion
Scores (MOS) using Spearman Rank Correlation (SRCC) and Mean Absolute Error
(MAE).

As shown in Table~\ref{tab:gpt54_calibration}, the raw GPT-5.4 ratings exhibit
moderate ranking agreement with professional designers, with a macro-average
SRCC of 0.570.
However, the raw ratings also show substantial systematic overestimation:
the macro-average MAE is 0.950 and the mean signed error is $+0.808$.
The bias is consistently positive across all eight dimensions, indicating that
GPT-5.4 tends to assign more lenient numerical ratings than professional
designers.

\textbf{Human-Anchored Calibration.}
To correct this systematic scoring bias, we perform post-hoc calibration using
the existing Task~1 human annotations.
For each aesthetic dimension $d$, we independently fit a monotonically
non-decreasing isotonic regression function
$f_d(\cdot)$ that maps the raw 1--5 GPT-5.4 rating to the corresponding
continuous professional-designer MOS.
Each mapping is constrained to the range $[1,5]$, and the eight dimensions are
calibrated independently.

All calibration functions are learned exclusively from Task~1.
We first evaluate the calibration procedure using five-fold cross-validation:
for each fold, the calibration mapping is fitted only on the training split and
evaluated on the held-out split.
After validation, we refit the eight calibration functions using all 1,395
successfully aligned Task~1 examples and freeze these mappings before applying
them to Task~4.
No Task~4 generation results, Pairwise Win Rate, or other Task~4 evaluation
signals are used to fit or select the calibration functions.

\textbf{Calibration Validation.}
Table~\ref{tab:gpt54_calibration} reports pooled out-of-fold results from the
five-fold cross-validation.
Calibration consistently reduces MAE across all eight aesthetic dimensions,
lowering the macro-average MAE from 0.950 to 0.607, corresponding to a
36.1\% reduction.
Meanwhile, the macro-average signed bias decreases from $+0.808$ to
approximately zero, indicating that the systematic leniency of the raw judge
scores is effectively removed.

The macro-average SRCC decreases moderately from 0.570 to 0.508 after
calibration.
The reduction is largely attributable to additional ties introduced by the piecewise-constant isotonic mappings.
Our calibration is intended to improve numerical agreement with the
professional-designer rating scale rather than to optimize ranking performance.

\begin{table}[t]
    \centering
    \caption{
    Five-fold cross-validation of GPT-5.4 aesthetic-score calibration against
    professional-designer MOS on Task~1.
    Bias denotes the mean signed error (\textit{prediction} $-$ \textit{MOS});
    values closer to zero indicate better calibration.
    Macro denotes the equally weighted average across the eight aesthetic
    dimensions.
    }
    \label{tab:gpt54_calibration}
    \small
    \setlength{\tabcolsep}{4.5pt}
    \renewcommand{\arraystretch}{1.08}
    \resizebox{0.7\linewidth}{!}{
    \begin{tabular}{@{}lcccccc@{}}
        \toprule
        \multirow{2}{*}{\textbf{Dimension}}
        & \multicolumn{2}{c}{\textbf{MAE} $\downarrow$}
        & \multicolumn{2}{c}{\textbf{Bias} $\rightarrow 0$}
        & \multicolumn{2}{c}{\textbf{SRCC} $\uparrow$} \\
        \cmidrule(lr){2-3}
        \cmidrule(lr){4-5}
        \cmidrule(lr){6-7}
        & \textbf{Raw} & \textbf{Cal.}
        & \textbf{Raw} & \textbf{Cal.}
        & \textbf{Raw} & \textbf{Cal.} \\
        \midrule
        Color
        & 0.95 & \textbf{0.61}
        & +0.86 & 0.00
        & 0.55 & 0.48 \\

        Typography
        & 0.70 & \textbf{0.55}
        & +0.52 & 0.00
        & 0.60 & 0.54 \\

        Graphics \& Imagery
        & 0.98 & \textbf{0.61}
        & +0.85 & 0.00
        & 0.59 & 0.52 \\

        Layout
        & 0.71 & \textbf{0.58}
        & +0.50 & 0.00
        & 0.66 & 0.60 \\

        Component Consistency
        & 1.03 & \textbf{0.58}
        & +0.95 & 0.00
        & 0.62 & 0.55 \\

        Visual Style Consistency
        & 1.00 & \textbf{0.60}
        & +0.87 & 0.00
        & 0.58 & 0.54 \\

        Copy Quality
        & 1.05 & \textbf{0.69}
        & +0.84 & 0.00
        & 0.36 & 0.30 \\

        Image--Text Fit
        & 1.17 & \textbf{0.64}
        & +1.07 & 0.00
        & 0.60 & 0.55 \\

        \midrule
        \textbf{Macro}
        & \textbf{0.95} & \textbf{0.61}
        & \textbf{+0.81} & \textbf{0.00}
        & \textbf{0.57} & \textbf{0.51} \\
        \bottomrule
    \end{tabular}
    }
\end{table}

Based on this validation, we apply the frozen dimension-specific calibration
functions to the GPT-5.4 ratings in Task~4.
For each valid generated UI, we first calibrate its eight dimension-level
ratings independently and then average the calibrated scores to obtain the
\textit{Aesthetic Judge Score}.
Invalid or failed generations receive a score of zero, and the model-level score
is computed over the same fixed evaluation denominator for all models.
The resulting metric therefore provides a human-anchored automatic assessment
of generated UI aesthetics, while Pairwise Win Rate serves as a complementary
measure of relative holistic preference.

\subsection{Additional Fine-Grained Analyses}
\label{sec:additional_fine_grained_analysis}

\begin{figure}[t]
    \centering
    \includegraphics[width=\textwidth]{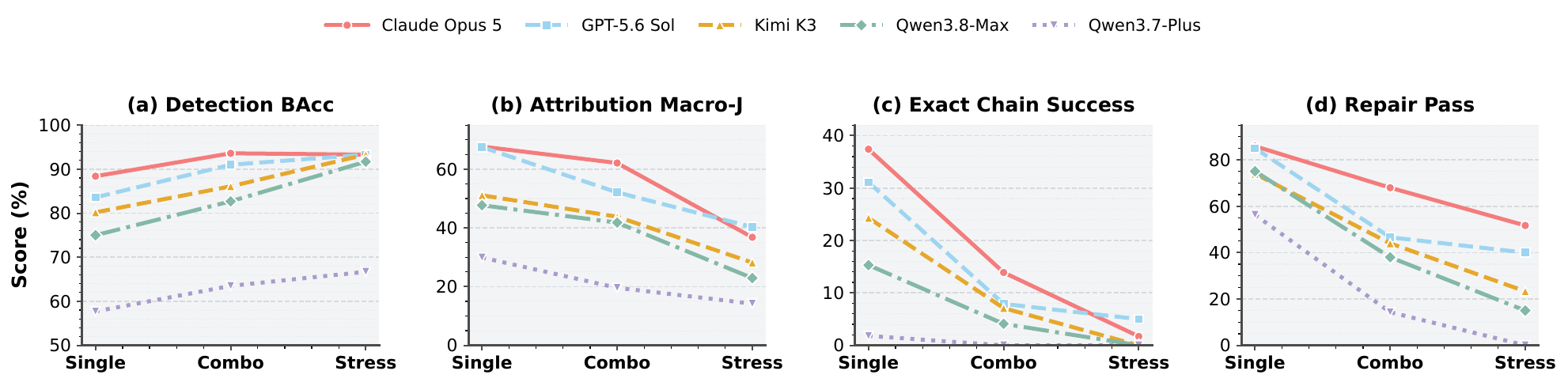}
    \caption{
    Performance under increasing defect complexity.
    We compare single-violation (\textit{Diagnostic}), compositional
    (\textit{Combo}), and high-density (\textit{Stress}) settings across
    degradation detection, violation attribution, exact-chain diagnosis, and
    aesthetic repair. Detection remains comparatively stable as more defects
    are introduced, whereas fine-grained diagnosis and repair become
    substantially more difficult.
    }
    \label{fig:complexity_gradient}
\end{figure}

\textbf{Effect of Defect Complexity.}
We first examine how model performance changes as multiple aesthetic violations
co-occur within the same interface. As shown in
Figure~\ref{fig:complexity_gradient}, increasing defect complexity has
substantially different effects on coarse detection and fine-grained
diagnosis. Degradation detection remains stable or even improves from
single-violation to compositional and stress settings, suggesting that an
interface becomes easier to recognize as problematic when more violations are
present. In contrast, attribution, exact-chain diagnosis, and repair degrade
consistently as the number of simultaneous defects increases. For example,
Claude Opus 5 maintains high detection accuracy, while its ECS decreases from
$0.374$ to $0.139$ and further to $0.017$; its Repair Pass similarly drops from
$0.859$ to $0.680$ and $0.517$. GPT-5.6 Sol exhibits the same pattern, with ECS
decreasing from $0.311$ to $0.079$ and $0.050$, and Repair Pass from $0.850$ to
$0.466$ and $0.400$. These results indicate that the principal challenge under
compositional defects is not recognizing that an interface is aesthetically
problematic, but exhaustively identifying and correcting the interacting
violations.

\begin{figure}[t]
    \centering
    \includegraphics[width=\textwidth]{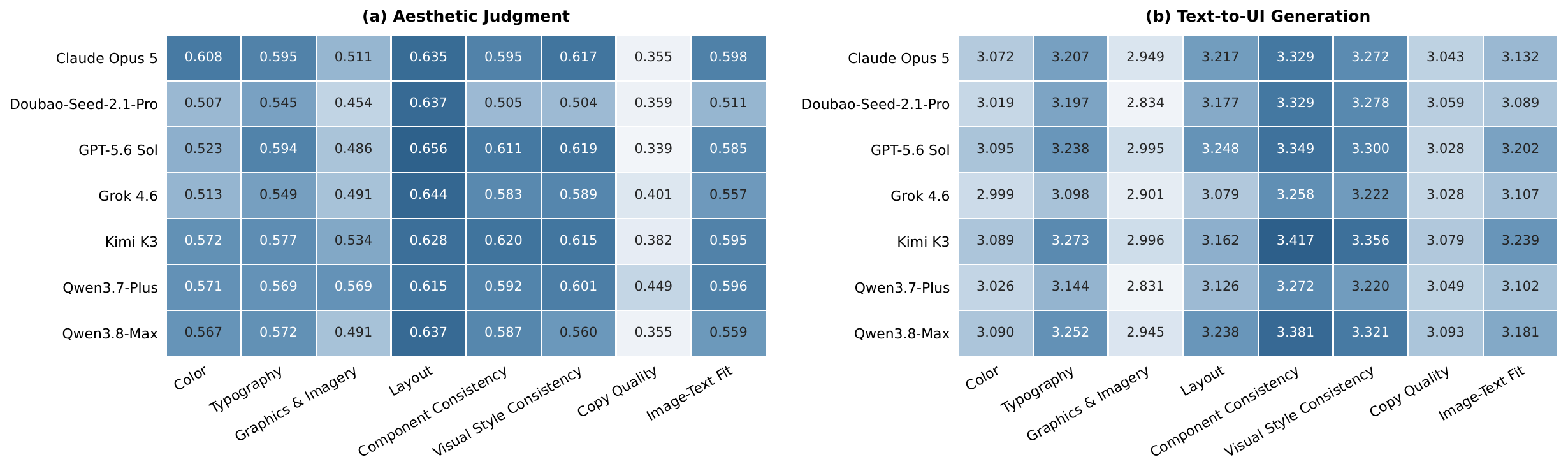}
    \caption{
    Dimension-wise aesthetic capability profiles.
    \textbf{Left:} SRCC with professional-designer MOS for Aesthetic Scoring.
    \textbf{Right:} human-calibrated aesthetic scores for successfully rendered
    Text-to-UI generations. The two panels use the same eight-dimensional
    perceptual rubric but characterize scoring and generation, respectively.
    }
    \label{fig:dimension_heatmap}
\end{figure}

\textbf{Dimension-Wise Aesthetic Capability Profiles.}
Figure~\ref{fig:dimension_heatmap} further decomposes aesthetic performance
along the eight perceptual dimensions shared by Aesthetic Scoring and
Text-to-UI Generation. For Aesthetic Scoring, the strongest frontier models
show consistently higher agreement with professional designers on
\textit{layout} (SRCC $0.615$--$0.656$), while \textit{copy quality} is
consistently more difficult (SRCC $0.339$--$0.449$). This suggests that
models capture global spatial organization more reliably than the contextual
appropriateness and quality of interface copy.

A different profile emerges for generated interfaces. Among successfully
rendered generations, \textit{component consistency} and
\textit{visual-style consistency} generally receive the highest calibrated
aesthetic scores, whereas \textit{graphics and imagery} is consistently among
the weakest dimensions. For example, GPT-5.6 Sol scores $3.349$ on component
consistency but $2.995$ on graphics and imagery, while Kimi-K3 scores $3.417$
and $2.996$, respectively. These dimension-level results show that aggregate
aesthetic scores conceal systematic differences in the types of aesthetic
knowledge that current models can judge and realize.

\noindent
\begin{minipage}{\textwidth}
\begin{wraptable}{r}{0.51\textwidth}
    \centering
    \caption{
    Decomposition of Aesthetic Repair performance.
    $F$: target fix; $C$: collateral preservation;
    $P$: content preservation; $S$: visual-scope compliance.
    }
    \label{tab:repair_decomp}
    \scriptsize
    \setlength{\tabcolsep}{3.0pt}
    \renewcommand{\arraystretch}{1.08}
    \resizebox{\linewidth}{!}{
    \begin{tabular}{@{}lccccc@{}}
        \toprule
        \textbf{Model}
        & $\mathbf{F}$
        & $\mathbf{C}$
        & $\mathbf{P}$
        & $\mathbf{S}$
        & \textbf{Repair} \\
        \midrule
        Claude Opus 5
        & 0.83 & 0.97 & 0.97 & 0.82 & \textbf{0.76} \\
        Doubao-Seed-2.1-Pro
        & 0.50 & 0.99 & 0.99 & 0.70 & 0.45 \\
        GPT-5.6 Sol
        & 0.69 & 0.95 & 0.95 & 0.75 & 0.66 \\
        Grok 4.6
        & 0.74 & 0.94 & 0.94 & 0.77 & 0.70 \\
        Kimi-K3
        & 0.59 & 0.77 & 0.77 & 0.64 & 0.57 \\
        Qwen3.7-Plus
        & 0.37 & \textbf{1.00} & \textbf{1.00} & 0.77 & 0.34 \\
        Qwen3.8-Max
        & 0.62 & 0.98 & 0.99 & 0.71 & 0.55 \\
        \bottomrule
    \end{tabular}
    }
\end{wraptable}

\textbf{What Limits Successful Repair?}
To identify the source of repair failures, Table~\ref{tab:repair_decomp}
decomposes the official Repair Pass criterion into target correction ($F$),
collateral preservation ($C$), content preservation ($P$), and visual-scope
compliance ($S$). For most frontier models, collateral and content preservation
are already high, while target correction and visual-scope compliance remain
more restrictive. Qwen3.7-Plus, for example, achieves $0.995$ collateral
preservation and perfect content preservation, but only $0.371$ target-fix
success, resulting in a Repair Pass of $0.342$. Similarly,
Doubao-Seed-2.1-Pro preserves collateral and content at $0.985$ and $0.992$,
respectively, but reaches only $0.498$ on target correction. In contrast,
Claude Opus 5 achieves substantially stronger target correction ($0.826$) and
visual-scope compliance ($0.824$). These results suggest that repair failures
are driven primarily by the difficulty of correctly resolving the intended
aesthetic defect while keeping the modification appropriately localized,
rather than by widespread corruption of unaffected content.
\end{minipage}

\subsection{Performance across Industries}
\label{sec:industry_analysis}

We further examine how model performance varies across 11 UI industries.
We first characterize the overall industry-level patterns across model groups,
then inspect individual frontier models, and finally analyze whether these
differences persist after controlling for model strength and where they emerge
along the diagnosis--repair pipeline.

\begin{figure*}[t]
    \centering
    \includegraphics[width=\textwidth]{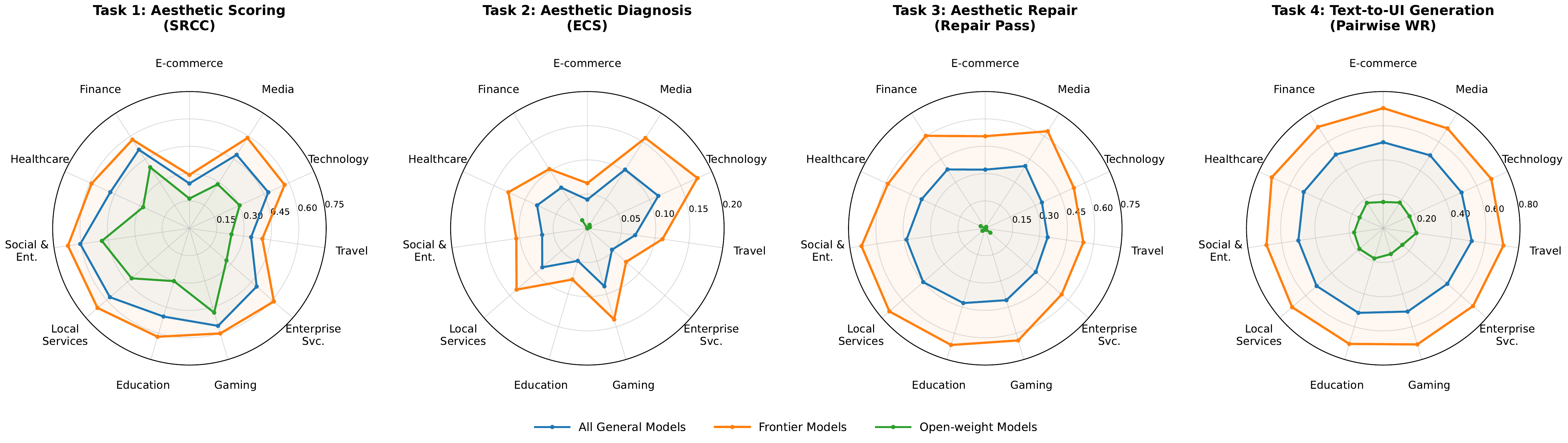}
    \caption{
    \textbf{Absolute performance across UI industries.}
    Group-average performance across 11 industries for Aesthetic Scoring
    (SRCC), Aesthetic Diagnosis (ECS), Aesthetic Repair (Repair Pass), and
    Text-to-UI Generation (Pairwise Win Rate). Each panel uses the original
    scale of its corresponding metric and should be compared within task.
    }
    \label{fig:industry_radar}
\end{figure*}

\textbf{Industry sensitivity differs substantially across tasks.}
Figure~\ref{fig:industry_radar} provides a group-level view of industry
variation. For frontier models, Aesthetic Scoring and Repair vary substantially
across industries: SRCC ranges from 0.293 on E-commerce to 0.673 on
Social \& Entertainment, while Repair Pass ranges from 0.505 on E-commerce
to 0.696 on Local Services. In contrast, Text-to-UI Generation is considerably
more stable, with frontier-model Pairwise Win Rate remaining around
0.69--0.72 across industries. Open-weight models show a large absolute gap,
particularly in Diagnosis and Repair, where performance is close to the floor.

\begin{figure*}[t]
    \centering
    \includegraphics[width=\textwidth]{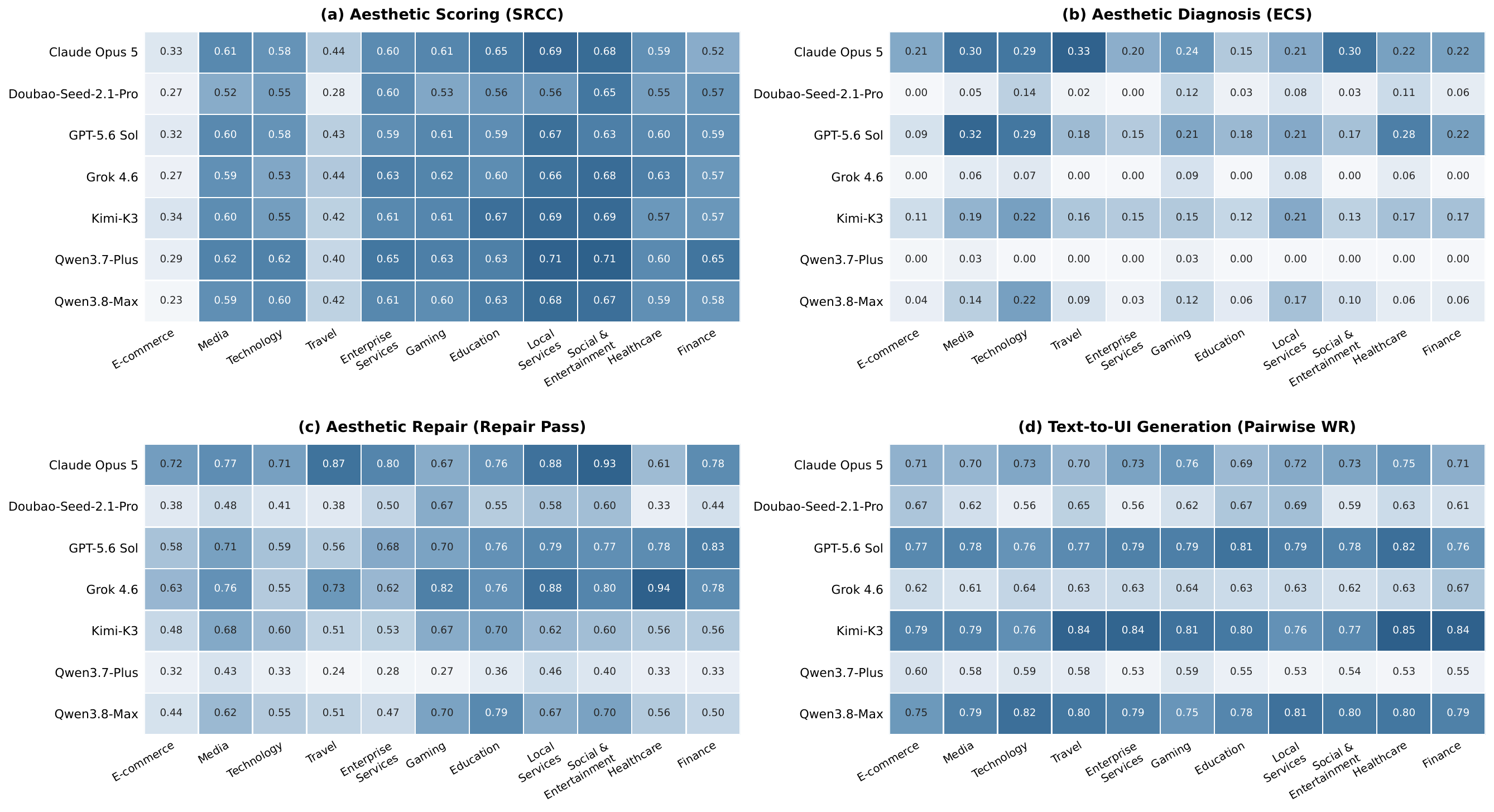}
    \caption{
    \textbf{Fine-grained performance of individual frontier models across 11 UI industries.}
    We report industry-wise results for
    (a) Aesthetic Scoring (SRCC),
    (b) Aesthetic Diagnosis (ECS),
    (c) Aesthetic Repair (Repair Pass), and
    (d) Text-to-UI Generation (Pairwise Win Rate).
    Cell values show the original metric scores, while color scales are
    independently normalized within each task for clearer comparison.
    }
    \label{fig:frontier_industry_heatmap}
\end{figure*}

\textbf{Group-level trends coexist with substantial model-specific variation.}
Figure~\ref{fig:frontier_industry_heatmap} further decomposes the frontier-model
average into individual models. Some industry patterns are shared across models:
for example, Aesthetic Scoring is generally weaker on E-commerce and stronger
on Local Services and Social \& Entertainment. However, Diagnosis and Repair
also exhibit substantial model-specific variation. In particular, ECS differs
sharply across models, with some models remaining relatively strong across
multiple industries while others approach zero on many categories. By
comparison, Text-to-UI Generation is more stable across industries within the
same model, and variation is dominated more by differences between models.

\begin{figure*}[t]
    \centering
    \includegraphics[width=\textwidth]{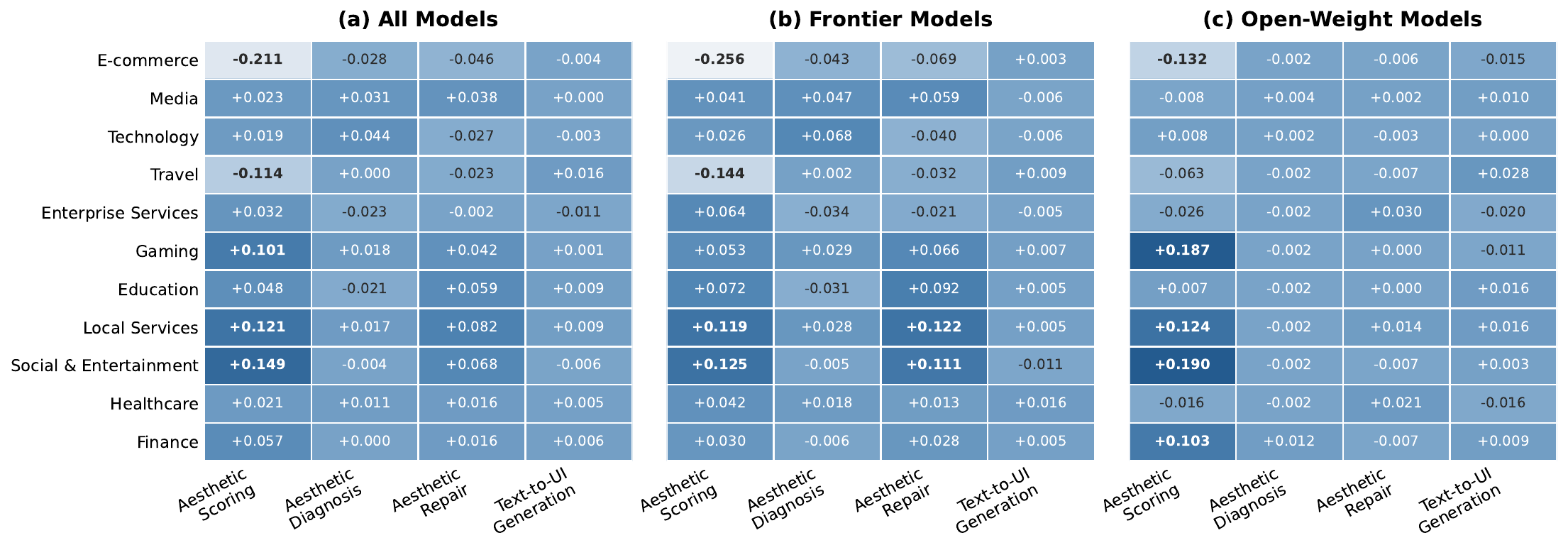}
    \caption{
    \textbf{Model-normalized industry effects across the four tasks.}
    Each cell reports the mean within-model deviation from overall performance:
    $\Delta_{g,t} =
    \frac{1}{|\mathcal{M}|}\sum_{m\in\mathcal{M}}
    (S_{m,g,t}-S_{m,\mathrm{overall},t})$.
    Positive values indicate above-overall performance and negative values
    indicate below-overall performance.
    }
    \label{fig:industry_normalized}
\end{figure*}

\textbf{Industry effects remain after controlling for model strength.}
As shown in Figure~\ref{fig:industry_normalized}, frontier models remain
consistently below their overall level on E-commerce in Scoring ($-0.256$),
Diagnosis ($-0.043$), and Repair ($-0.069$), while Local Services and
Social \& Entertainment show positive shifts in both Scoring and Repair.
Similar trends appear across all general-purpose models. By contrast,
Text-to-UI Pairwise Win Rate changes little across industries, with most
normalized deviations within about $\pm 0.02$. Open-weight models preserve some
industry structure in Scoring, but floor effects largely obscure such variation
in Diagnosis and Repair.

\begin{figure*}[t]
    \centering
    \includegraphics[width=\textwidth]{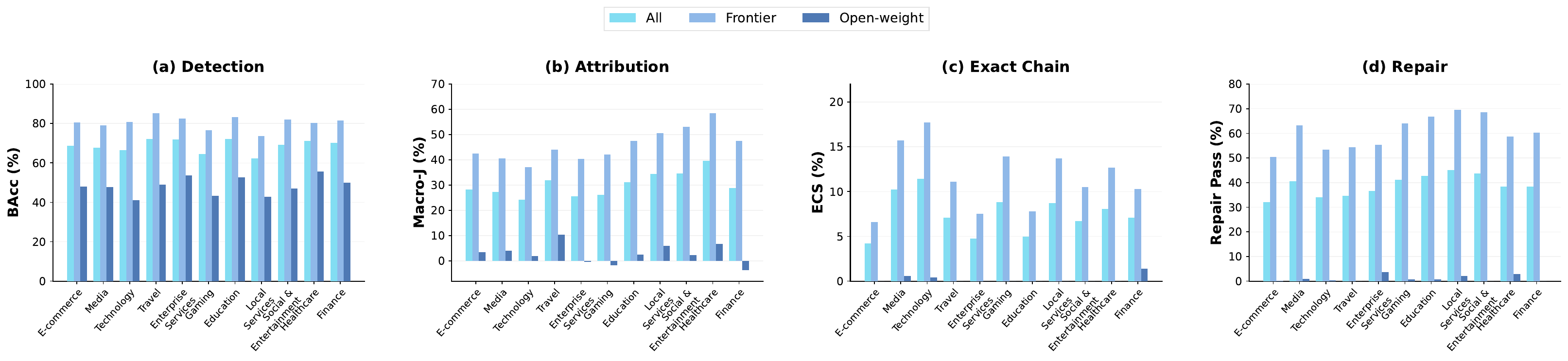}
    \caption{
    \textbf{Industry-wise performance from coarse diagnosis to corrective action.}
    We report degradation detection (BAcc), violation attribution (Macro-J),
    exact diagnosis-chain success (ECS), and final Repair Pass for the three
    model groups across industries.
    }
    \label{fig:industry_diag_repair}
\end{figure*}

\textbf{Industry differences become more pronounced at finer levels of aesthetic understanding.}
Figure~\ref{fig:industry_diag_repair} shows that frontier-model degradation
detection remains relatively stable across industries, with BAcc around
0.74--0.85. Larger variation appears in attribution and exact-chain diagnosis,
and persists in downstream repair. Open-weight models similarly retain
non-trivial coarse detection ability but drop sharply on attribution, ECS, and
Repair. This suggests that industry-specific challenges arise mainly when
models must identify and act on specific aesthetic problems, rather than when
simply recognizing that an interface is degraded.